\pdfoutput=1
\def\arxivversion{}
\documentclass[letterpaper]{article} %
\ifdefined\arxivversion
\usepackage[preprint]{aaai2027}
\newcommand{\insupp}[1]{in \cref{#1}}
\else
\usepackage[submission]{aaai2027}  %
\def\nolinks{} %
\fi
\newif\iflinks
\linkstrue
\ifdefined\nolinks\linksfalse\fi %
\ifdefined\pdfstartlink\else\linksfalse\fi
\makeatletter
\def\LP@int#1{\global\@namedef{LP@int@#1}{}}%
\iflinks
  \def\LP@dest#1{%
    \ifvmode\pdfdest name{ref.#1} xyz\relax
    \else\smash{\raise\baselineskip\hbox{\pdfdest name{ref.#1} xyz}}\fi
    \protected@write\@auxout{}{\string\LP@int{#1}}}
  \def\label#1{\@bsphack\LP@dest{#1}%
    \protected@write\@auxout{}{\string\newlabel{#1}{{\@currentlabel}{\thepage}}}\@esphack}
\fi
\makeatother
\usepackage[hyphens]{url}  %
\usepackage{graphicx} %
\usepackage{natbib}  %
\usepackage{caption} %
\usepackage{booktabs}
\usepackage{amsmath}
\usepackage{amssymb}
\usepackage{xcolor}
\usepackage{xspace}
\usepackage{xr}                      %
\usepackage[capitalize]{cleveref}    %
\usepackage{colortbl}                 %
\definecolor{LinkBlue}{HTML}{000099}
\definecolor{GoodGreen}{HTML}{1E8449}
\definecolor{BadRed}{HTML}{D62828}
\definecolor{TableOurs}{HTML}{E6F4EA}
\definecolor{TableGroup}{HTML}{F3F4F6}
\newcommand{\good}[1]{\leavevmode{\boldmath\color{GoodGreen}\textbf{#1}}}
\newcommand{\bad}[1]{\leavevmode{\boldmath\color{BadRed}\textbf{#1}}}
\makeatletter
\iflinks
  \ifdefined\nocitelinks\else
  \renewcommand\hyper@natlinkstart[1]{\leavevmode\pdfstartlink attr{/Border[0 0 0]} goto name{cite.#1}\begingroup\color{LinkBlue}}
  \renewcommand\hyper@natlinkend{\endgroup\pdfendlink}
  \renewcommand\hyper@natanchorstart[1]{\ifvmode\pdfdest name{cite.#1} xyz\relax\else\smash{\raise\baselineskip\hbox{\pdfdest name{cite.#1} xyz}}\fi}
  \renewcommand\hyper@natanchorend{}
  \fi
  \def\LP@first#1,#2\@nil{#1}
  \def\LP@link#1#2{\edef\LP@target{\LP@first#1,\@nil}%
    \@ifundefined{LP@int@\LP@target}{#2}%
      {\leavevmode\pdfstartlink attr{/Border[0 0 0]} goto name{ref.\LP@target}\begingroup\color{LinkBlue}#2\endgroup\pdfendlink}}
  \def\LP@crefwrap#1{\LP@link{#1}{\LP@cref{#1}}}
  \def\LP@Crefwrap#1{\LP@link{#1}{\LP@Cref{#1}}}
  \AtBeginDocument{%
    \expandafter\let\expandafter\LP@cref\csname cref \endcsname
    \expandafter\let\expandafter\LP@Cref\csname Cref \endcsname
    \expandafter\let\csname cref \endcsname\LP@crefwrap
    \expandafter\let\csname Cref \endcsname\LP@Crefwrap}
\fi
\makeatother

\providecommand{\nolinkurl}[1]{\url{#1}}
\providecommand{\insupp}[1]{in the supplementary} %
\newif\ifwithsupp
\withsuppfalse
\ifdefined\submissionpdf\withsuppfalse\fi
\ifdefined\arxivversion\withsupptrue\fi
\ifwithsupp\else\IfFileExists{supp.aux}{\externaldocument{supp}}{%
\makeatletter
\newlabel{sec:supp_impl}{{A}{1}}
\newlabel{sec:supp_impl@cref}{{[section][1][]A}{[1][1][]1}}
\newlabel{sec:supp_features}{{A.1}{1}}
\newlabel{sec:supp_features@cref}{{[subsection][1][1]A.1}{[1][1][]1}}
\newlabel{sec:supp_subspace}{{A.2}{1}}
\newlabel{sec:supp_subspace@cref}{{[subsection][2][1]A.2}{[1][1][]1}}
\newlabel{eq:shrink}{{S1}{1}}
\newlabel{eq:shrink@cref}{{[equation][1][]S1}{[1][1][]1}}
\newlabel{sec:supp_proof}{{A.2}{2}}
\newlabel{sec:supp_proof@cref}{{[subsection][2][1]A.2}{[1][2][]2}}
\newlabel{sec:supp_tree_metric}{{A.3}{2}}
\newlabel{sec:supp_tree_metric@cref}{{[subsection][3][1]A.3}{[1][2][]2}}
\newlabel{sec:supp_splits}{{A.4}{2}}
\newlabel{sec:supp_splits@cref}{{[subsection][4][1]A.4}{[1][2][]2}}
\newlabel{sec:supp_baselines}{{A.5}{3}}
\newlabel{sec:supp_baselines@cref}{{[subsection][5][1]A.5}{[1][3][]3}}
\newlabel{sec:supp_data}{{B}{3}}
\newlabel{sec:supp_data@cref}{{[section][2][]B}{[1][3][]3}}
\newlabel{sec:supp_worlds}{{B.1}{3}}
\newlabel{sec:supp_worlds@cref}{{[subsection][1][2]B.1}{[1][3][]3}}
\newlabel{sec:supp_cf}{{B.2}{3}}
\newlabel{sec:supp_cf@cref}{{[subsection][2][2]B.2}{[1][3][]3}}
\newlabel{tab:supp_world_generators}{{S1}{4}}
\newlabel{tab:supp_world_generators@cref}{{[table][1][]S1}{[1][3][]4}}
\newlabel{sec:supp_asvspoof}{{B.3}{4}}
\newlabel{sec:supp_asvspoof@cref}{{[subsection][3][2]B.3}{[1][4][]4}}
\newlabel{sec:supp_tree_set}{{B.4}{4}}
\newlabel{sec:supp_tree_set@cref}{{[subsection][4][2]B.4}{[1][4][]4}}
\newlabel{sec:fingerprints}{{C}{4}}
\newlabel{sec:fingerprints@cref}{{[section][3][]C}{[1][4][]4}}
\newlabel{tab:supp_tree_models}{{S2}{5}}
\newlabel{tab:supp_tree_models@cref}{{[table][2][]S2}{[1][4][]5}}
\newlabel{sec:supp_nulls}{{C}{5}}
\newlabel{sec:supp_nulls@cref}{{[section][3][]C}{[1][5][]5}}
\newlabel{sec:supp_brightness}{{C}{5}}
\newlabel{sec:supp_brightness@cref}{{[section][3][]C}{[1][5][]5}}
\newlabel{fig:fingerprints}{{S1}{6}}
\newlabel{fig:fingerprints@cref}{{[figure][1][]S1}{[1][5][]6}}
\newlabel{sec:supp_featspace}{{C}{6}}
\newlabel{sec:supp_featspace@cref}{{[section][3][]C}{[1][6][]6}}
\newlabel{sec:supp_results}{{D}{6}}
\newlabel{sec:supp_results@cref}{{[section][4][]D}{[1][6][]6}}
\newlabel{fig:discovery}{{S2}{7}}
\newlabel{fig:discovery@cref}{{[figure][2][]S2}{[1][6][]7}}
\newlabel{sec:supp_native}{{D}{7}}
\newlabel{sec:supp_native@cref}{{[section][4][]D}{[1][7][]7}}
\newlabel{tab:supp_unseen_scores}{{S3}{9}}
\newlabel{tab:supp_unseen_scores@cref}{{[table][3][]S3}{[1][6][]9}}
\newlabel{tab:supp_unseen_models}{{S4}{10}}
\newlabel{tab:supp_unseen_models@cref}{{[table][4][]S4}{[1][6][]10}}
\newlabel{tab:supp_unseen_generators}{{S5}{11}}
\newlabel{tab:supp_unseen_generators@cref}{{[table][5][]S5}{[1][6][]11}}
\newlabel{tab:discovery}{{S6}{12}}
\newlabel{tab:discovery@cref}{{[table][6][]S6}{[1][6][]12}}
\newlabel{tab:calibration}{{S7}{12}}
\newlabel{tab:calibration@cref}{{[table][7][]S7}{[1][6][]12}}
\newlabel{tab:supp_native}{{S8}{13}}
\newlabel{tab:supp_native@cref}{{[table][8][]S8}{[1][6][]13}}
\newlabel{tab:supp_fsd}{{S9}{13}}
\newlabel{tab:supp_fsd@cref}{{[table][9][]S9}{[1][7][]13}}
\newlabel{tab:tree}{{S10}{14}}
\newlabel{tab:tree@cref}{{[table][10][]S10}{[1][7][]14}}
\newlabel{tab:lineage}{{S11}{15}}
\newlabel{tab:lineage@cref}{{[table][11][]S11}{[1][7][]15}}
\newlabel{tab:supp_pipelines}{{S12}{16}}
\newlabel{tab:supp_pipelines@cref}{{[table][12][]S12}{[1][7][]16}}
\newlabel{tab:supp_encoders}{{S13}{17}}
\newlabel{tab:supp_encoders@cref}{{[table][13][]S13}{[1][7][]17}}
\newlabel{tab:supp_equal_treatment}{{S14}{17}}
\newlabel{tab:supp_equal_treatment@cref}{{[table][14][]S14}{[1][8][]17}}
\newlabel{tab:ladder}{{S15}{18}}
\newlabel{tab:ladder@cref}{{[table][15][]S15}{[1][8][]18}}
\newlabel{tab:supp_per_unseen}{{S16}{19}}
\newlabel{tab:supp_per_unseen@cref}{{[table][16][]S16}{[1][8][]19}}
\newlabel{tab:supp_pool}{{S17}{19}}
\newlabel{tab:supp_pool@cref}{{[table][17][]S17}{[1][8][]19}}
\newlabel{tab:supp_learned_head}{{S18}{20}}
\newlabel{tab:supp_learned_head@cref}{{[table][18][]S18}{[1][8][]20}}
\newlabel{sec:supp_fresh}{{E}{21}}
\newlabel{sec:supp_fresh@cref}{{[section][5][]E}{[1][21][]21}}
\newlabel{sec:supp_confirm}{{E}{21}}
\newlabel{sec:supp_confirm@cref}{{[section][5][]E}{[1][21][]21}}
\newlabel{tab:supp_fresh_scores}{{S19}{23}}
\newlabel{tab:supp_fresh_scores@cref}{{[table][19][]S19}{[1][21][]23}}
\newlabel{tab:supp_fresh_models}{{S20}{24}}
\newlabel{tab:supp_fresh_models@cref}{{[table][20][]S20}{[1][21][]24}}
\newlabel{tab:supp_fresh_calibration}{{S21}{25}}
\newlabel{tab:supp_fresh_calibration@cref}{{[table][21][]S21}{[1][21][]25}}
\newlabel{tab:supp_relatives}{{S22}{25}}
\newlabel{tab:supp_relatives@cref}{{[table][22][]S22}{[1][21][]25}}
\newlabel{tab:supp_openset}{{S23}{26}}
\newlabel{tab:supp_openset@cref}{{[table][23][]S23}{[1][21][]26}}
\newlabel{tab:supp_poolchannel}{{S24}{27}}
\newlabel{tab:supp_poolchannel@cref}{{[table][24][]S24}{[1][21][]27}}
\newlabel{tab:supp_mixed}{{S25}{27}}
\newlabel{tab:supp_mixed@cref}{{[table][25][]S25}{[1][21][]27}}
\newlabel{sec:supp_speech}{{F}{28}}
\newlabel{sec:supp_speech@cref}{{[section][6][]F}{[1][28][]28}}
\newlabel{sec:supp_moves}{{G}{28}}
\newlabel{sec:supp_moves@cref}{{[section][7][]G}{[1][28][]28}}
\newlabel{sec:supp_finetune}{{G.1}{28}}
\newlabel{sec:supp_finetune@cref}{{[subsection][1][7]G.1}{[1][28][]28}}
\newlabel{tab:speech}{{S26}{29}}
\newlabel{tab:speech@cref}{{[table][26][]S26}{[1][28][]29}}
\newlabel{tab:supp_codec_calibration}{{S27}{29}}
\newlabel{tab:supp_codec_calibration@cref}{{[table][27][]S27}{[1][28][]29}}
\newlabel{sec:supp_style}{{G.2}{29}}
\newlabel{sec:supp_style@cref}{{[subsection][2][7]G.2}{[1][29][]29}}
\newlabel{sec:supp_gallery}{{G.3}{29}}
\newlabel{sec:supp_gallery@cref}{{[subsection][3][7]G.3}{[1][29][]29}}
\newlabel{tab:supp_finetune}{{S28}{30}}
\newlabel{tab:supp_finetune@cref}{{[table][28][]S28}{[1][28][]30}}
\newlabel{sec:supp_tree}{{H}{30}}
\newlabel{sec:supp_tree@cref}{{[section][8][]H}{[1][30][]30}}
\newlabel{tab:supp_style}{{S29}{31}}
\newlabel{tab:supp_style@cref}{{[table][29][]S29}{[1][29][]31}}
\newlabel{tab:supp_gallery}{{S30}{31}}
\newlabel{tab:supp_gallery@cref}{{[table][30][]S30}{[1][29][]31}}
\newlabel{sec:supp_modern}{{I}{32}}
\newlabel{sec:supp_modern@cref}{{[section][9][]I}{[1][32][]32}}
\newlabel{sec:exp_modern}{{I}{32}}
\newlabel{sec:exp_modern@cref}{{[section][9][]I}{[1][32][]32}}
\newlabel{tab:modern}{{S31}{32}}
\newlabel{tab:modern@cref}{{[table][31][]S31}{[1][32][]32}}
\newlabel{tab:supp_verification}{{S32}{32}}
\newlabel{tab:supp_verification@cref}{{[table][32][]S32}{[1][32][]32}}
\newlabel{tab:supp_modern_all}{{S33}{33}}
\newlabel{tab:supp_modern_all@cref}{{[table][33][]S33}{[1][32][]33}}
\newlabel{tab:supp_modern_families}{{S34}{34}}
\newlabel{tab:supp_modern_families@cref}{{[table][34][]S34}{[1][32][]34}}
\newlabel{tab:supp_modern_models}{{S35}{35}}
\newlabel{tab:supp_modern_models@cref}{{[table][35][]S35}{[1][32][]35}}
\newlabel{sec:supp_controls}{{J}{36}}
\newlabel{sec:supp_controls@cref}{{[section][10][]J}{[1][36][]36}}
\newlabel{tab:controls}{{S36}{37}}
\newlabel{tab:controls@cref}{{[table][36][]S36}{[1][36][]37}}
\newlabel{tab:supp_controls_rt}{{S37}{38}}
\newlabel{tab:supp_controls_rt@cref}{{[table][37][]S37}{[1][36][]38}}
\newlabel{tab:supp_controls_swap}{{S38}{39}}
\newlabel{tab:supp_controls_swap@cref}{{[table][38][]S38}{[1][36][]39}}
\newlabel{tab:supp_resolution}{{S39}{40}}
\newlabel{tab:supp_resolution@cref}{{[table][39][]S39}{[1][36][]40}}
\newlabel{sec:supp_enroll}{{K}{41}}
\newlabel{sec:supp_enroll@cref}{{[section][11][]K}{[1][41][]41}}
\newlabel{sec:supp_related}{{L}{41}}
\newlabel{sec:supp_related@cref}{{[section][12][]L}{[1][41][]41}}
\makeatother
}\fi

\title{Fingerprints Run in the Family: Open-World Lineage Attribution of Generative Models}
\ifdefined\arxivversion
\author{Tianle Chen\textsuperscript{\rm 1}, Yuwen Tan\textsuperscript{\rm 1}, Wenqi Wang\textsuperscript{\rm 1}, Lingxiao Li\textsuperscript{\rm 1}, Jike Zhong\textsuperscript{\rm 2}, Jean Huang\textsuperscript{\rm 3}, Pushkar~Singh\textsuperscript{\rm 4}\thanks{Pushkar Singh contributed in an advisory role only. This work did not use any Google resources.}}
\affiliations{
\textsuperscript{\rm 1}Boston University\\
\textsuperscript{\rm 2}University of Southern California\\
\textsuperscript{\rm 3}University of Toronto\\
\textsuperscript{\rm 4}Google\\
\{tianle, yuwentan, wqwang, lxli\}@bu.edu, pushkarsingh@google.com
}
\else
\author{Anonymous submission}
\affiliations{}
\fi

\begin{document}
\maketitle
\begin{abstract}
Many new image generators are fine-tuned from a small number of base models. While most image forensics methods focus on whether an image is real or generated, they do not tell us which generator produced it or how it relates to other models.
Here we study this problem in an open-world setting: given a set of outputs, can we identify the registered generator or model family that produced them, or detect that they came from an unseen generator?
We introduce LineagePrint, which projects frozen forensic features onto a subspace that separates registered generators and averages them into one fingerprint per set.
We use outputs processed in the same way as the test set to center the fingerprint and calibrate the threshold for flagging unseen generators.
On sets of 8 images, LineagePrint separates 29 unseen models from 12 registered generators with an AUROC of 0.89 across ten processing pipelines. The strongest out-of-distribution baseline reaches 0.82 with the same set averaging and centering.
LineagePrint also outperforms the baselines on a second image dataset and on speech across codecs, and the false-alarm rate stays close to the target.
Light descendants and weight merges are often attributed to their base model, while heavy style fine-tunes rarely are.
Using the distance to a base model's fingerprint, we verify family membership for 25 documented relatives with an AUROC of 0.92.
Models with the same architecture can have different fingerprints: Playground v2, trained from scratch on the SDXL architecture, is as far from SDXL as from SD1 and SD2.
Changing only the decoder switches the attribution from SD1 to SD2.
By contrast, SDXL, FLUX.1, and SD3.5 are still attributed to their original families after a round trip through another family's autoencoder.
\end{abstract}
\begin{figure*}[t]
  \centering
  \includegraphics[width=\textwidth]{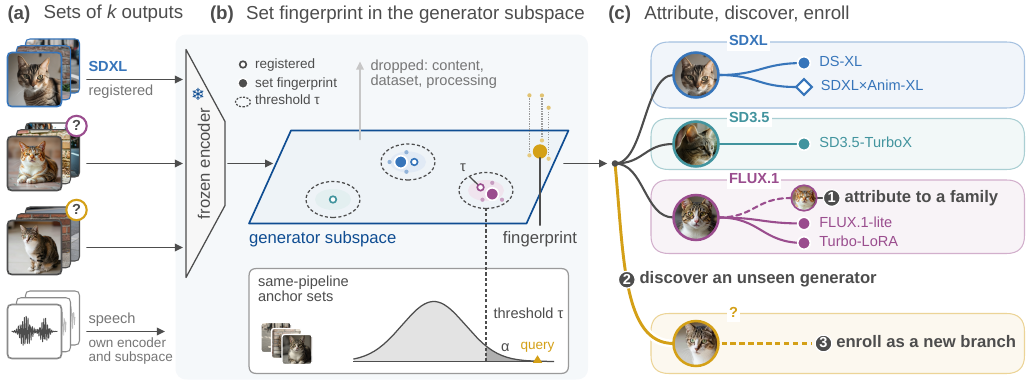}
  \caption{\textbf{LineagePrint.}
  (a) Sets of $k$ outputs from a registered generator (SDXL) and from generators to identify (?) enter a frozen encoder.
  (b) The generator subspace keeps the directions that separate registered generators and drops content, dataset and processing.
  A set's fingerprint is the mean of its $k$ projected outputs.
  Anchor sets from the same pipeline fix the threshold $\tau$, and the gold query exceeds it.
  (c) A fingerprint inside a ring is (1) attributed to that family (here FLUX.1).
  One outside every ring is (2) discovered as an unseen generator (\cref{sec:exp_discovery}) and (3) enrolled as a new branch (\insupp{sec:supp_enroll}).
  Each family band holds light descendants that are attributed to their base (\cref{sec:exp_tree}).
  DS: DreamShaper, Anim: Animagine, Turbo-LoRA: FLUX.1-dev with the Turbo-Alpha LoRA fused.}
  \label{fig:pipeline}
\end{figure*}
 \section{Introduction}
\label{sec:intro}

New image generators appear every week, and many are fine-tunes of a few base models.
Community Forensics~\cite{park2025community} collects images from 4,803 generators, and many of them are based on Stable Diffusion~\cite{rombach2022ldm}.
Most work in image forensics asks whether an image is real or generated. Detectors generally work best on generators represented in their training data~\cite{corvi2023diffusion,ojha2023universal,park2025community}.

For platforms, licensors, and investigators responding to synthetic media, it also matters where the outputs came from.
\emph{Which generator produced this set of outputs?}
\emph{Is it a registered generator or one we have not seen before?}
\emph{Which model family does it belong to?}
A family includes a base model and descendants produced by fine-tuning, distillation, weight merging, or version updates.
These questions are \emph{open-world} because the list of registered generators is never complete.
Prior attribution methods score one image at a time~\cite{yu2019attributing,song2024manifpt,yang2023pose,nguyen2025forensic,girish2021discovery}. Lineage methods need model weights, prompts, or a fingerprint for each reference model~\cite{horwitz2025unsupervised,ha2026identifying,sinitsa2024deep}.
By contrast, we attribute a set of released outputs without accessing the generator's weights or querying it.

Averaging features from a few outputs of the same generator gives a stable representation, which we call its \emph{fingerprint}.
Two properties make this fingerprint useful.
First, fingerprints of different generators \emph{differ along a few directions}. These directions stay nearly the same when the captions or rendering setup change, so we can use a subspace fit on one dataset to analyze another (\cref{sec:method_subspace}).
Second, fingerprints are often \emph{inherited}, consistent with the fine-tuning relationships that DIF~\cite{sinitsa2024deep} finds from images. Our method usually attributes light descendants and weight merges to their base model, but rarely does so for heavy style fine-tunes (\cref{sec:exp_tree}).

Based on these observations, we propose LineagePrint (\cref{fig:pipeline}).
We apply linear discriminant analysis (LDA) to frozen encoder features from the registered generators to fit a \emph{generator subspace}. The fingerprint of a set is the mean of its projected features.
The subspace keeps only the directions that separate generators, so their fingerprints remain distinguishable when the dataset or processing changes.
To detect unseen generators, LineagePrint (LDA) scores a set by its distance to the nearest registered fingerprint. LineagePrint (PLDA) uses a probabilistic LDA likelihood ratio instead.
An unlabeled \emph{pool} of generated outputs centers the fingerprint. We use \emph{anchor sets} from registered generators to calibrate a split-conformal threshold with a target false-alarm rate of $\alpha$. Both use the same processing as the test sets.
Set averaging, the LDA and PLDA back end, and conformal calibration come from camera fingerprints~\cite{lukas2006digital}, speaker verification~\cite{dehak2011frontend,garciaromero2011length} and conformal testing~\cite{bates2023outliers}. LineagePrint brings them to open-world generator discovery.

We test LineagePrint in three settings.
For \emph{discovery}, we score 29 unseen models from 11 families and 12 unseen generators on a second dataset~\cite{lin2014coco,young2014flickr30k} against the registered generators across ten processing pipelines. We compare with out-of-distribution scores~\cite{hendrycks2017msp,liu2020energy,lee2018mahalanobis,sun2022knn} using the same set averaging and centering.
For \emph{lineage}, we use an 18-model family tree with known relationships and descendants from two newer families. We test when a descendant can still be attributed to its base model.
To see what changes the fingerprint, we vary the decoder, sampling resolution, or autoencoder one at a time.

\noindent In summary, our contributions are:
\begin{itemize}
  \item \textbf{Open-world lineage attribution from sets.} We ask whether a set of outputs came from a registered generator, a model family, or an unseen generator. We introduce LineagePrint, a closed-form set fingerprint with a calibrated false-alarm rate.
  \item \textbf{Discovery of unseen generators (\cref{sec:exp_discovery}).} From sets of 8 images, LineagePrint separates 29 unseen models from the registered generators with \good{AUROC 0.89} across ten processing pipelines. The strongest out-of-distribution baseline reaches 0.82 with the same set averaging and centering. LineagePrint keeps the false-alarm rate near $\alpha$.
  \item \textbf{Fingerprints run in families, up to a boundary (\cref{sec:exp_tree}).} LineagePrint attributes light descendants and merges to their base model in 551 of 576 sets, compared with just 1 of 360 sets from heavy style fine-tunes. The distance to the base model's fingerprint verifies family membership with \good{AUROC 0.92}. \emph{A shared architecture does not make a shared fingerprint}: Playground v2, trained from scratch on the SDXL architecture, lies outside the SDXL family.
  \item \textbf{What carries the family signal (\cref{sec:exp_controls}).} Swapping only the decoder changes the attribution from SD1.5 to SD2.1 in all 24 sets. By contrast, SDXL, FLUX.1, and SD3.5 keep their original family attribution when their images are passed through another family's autoencoder. In controlled LoRA fine-tuning, forensic features follow the base model while semantic features follow the style.
\end{itemize}
\section{Related Work}
\label{sec:related}

\noindent\textbf{AI-generated image detection and its generalization.}
Some detectors train on images from one generative adversarial network (GAN)~\cite{wang2020cnn} or on frozen CLIP~\cite{radford2021clip} features~\cite{ojha2023universal}.
Thresholds can shift on unseen generators, but a small target set can help recalibrate them~\cite{yang2026calibrated}.
These detectors ask whether an image is real or generated. We ask which generator or family produced a set of outputs.

\noindent\textbf{Model attribution and fingerprints.}
Camera forensics averages sensor noise across images to obtain a device fingerprint~\cite{lukas2006digital}. \citet{marra2019gans} use the same idea for GANs.
\citet{yu2019attributing} learn fingerprints that separate GANs differing only in their training seed, and ManiFPT~\cite{song2024manifpt} defines fingerprints by the artifacts in a model's outputs. Other methods attribute single images to text-to-image generators~\cite{sha2023defake}, or to an architecture or family~\cite{yang2022dnadet}.
Open-set attribution rejects images of unseen generators~\cite{yang2023pose}. \citet{girish2021discovery} and others~\cite{nguyen2025forensic} cluster the rejected images into new sources.
These methods score one test image at a time. We instead compute a fingerprint for the whole test set in a subspace that separates registered generators. We use it to flag unseen generators at a target false-alarm rate.

\noindent\textbf{Model lineage and provenance.}
From weights, \citet{horwitz2025unsupervised} recover fine-tuning relationships in a Stable Diffusion family tree, including both parents of a merge.
Neural Lineage and Neural Phylogeny find the parent and the direction of a fine-tune~\cite{yu2024neural,yu2025phylogeny}.
From outputs, DIF~\cite{sinitsa2024deep} extracts a fingerprint of one model from a few of its images and relates fine-tuned models to it. The closest work, \citet{ha2026identifying}, tests the output features of a prompted suspect model against a pool of base models.
These methods need the weights, prompts sent to the suspect model, or a fingerprint extracted for each reference model.
We use one subspace to attribute community fine-tunes and descendants of FLUX.1 and SD3.5 from released outputs. We also place weight merges in a family tree and test how decoder swaps change the family fingerprint.
Concurrent work recovers family structure from an image attributor~\cite{livne2026scalable}.

\noindent\textbf{Out-of-distribution scores, set-level tests and conformal calibration.}
Out-of-distribution (OOD) scores include maximum softmax probability (MSP)~\cite{hendrycks2017msp}, energy~\cite{liu2020energy}, Mahalanobis distance~\cite{lee2018mahalanobis} and nearest-neighbor (kNN) distance~\cite{sun2022knn}.
Set-level tests such as maximum mean discrepancy (MMD)~\cite{gretton2012kernel} ask whether a sample differs from a reference, and MMD on embeddings separates generative processes~\cite{mukherjee2026beyond}.
Conformal prediction turns a score into a threshold with a finite-sample guarantee~\cite{vovk2005conformal}. Conformal outlier tests bound the false-alarm rate when calibration and test inliers are exchangeable~\cite{bates2023outliers}.
We draw calibration sets from registered generators and process the outputs through the same pipeline as the test sets.
\section{LineagePrint}
\label{sec:method}

\noindent\textbf{Problem setup.}
We observe a set $S=\{x_1,\dots,x_k\}$ of $k$ outputs from one generator after processing, such as encoding with a codec, resizing, or uploading to a platform.
The registered generators $g=1,\dots,M$ form the gallery. A generator outside the gallery is unseen.
Registered generator $g$ provides $n_g$ fitting outputs $X_g$, $n=\sum_g n_g$, and a frozen encoder $\phi$ maps each output $x$ to $\phi(x)\in\mathbb{R}^d$.
Attribution assigns $S$ to a registered generator or family, and open-world discovery decides whether $S$ comes from an unseen generator.
We use two groups of outputs processed through the same pipeline as $S$. An unlabeled pool of generated outputs centers the fingerprint (\cref{sec:method_fingerprint}), and anchor sets of registered generators calibrate the threshold (\cref{sec:method_calibration}).

\subsection{Generator subspace}
\label{sec:method_subspace}

Generators differ mainly along a few feature directions.
We measure this with the participation ratio $(\sum_j s_j^2)^2/\sum_j s_j^4$ of the singular values $s_j$ of the centered generator means, which estimates the effective number of directions.
In the forensic features of \cref{sec:exp_setup}, the mean features of 12 generators have a participation ratio of 2.30 of a possible 11. The first two directions span nearly the same plane in two datasets with different captions and rendering setups. The affinity between the planes is 0.90, compared with 0.52 for random splits. We measure affinity as the mean squared cosine of the principal angles between the planes.
Mean images show the same structure in raw pixels, with global brightness as the first axis (\insupp{sec:fingerprints}).
To keep the directions that separate registered generators, we apply reduced-rank linear discriminant analysis (LDA)~\cite{hastie2009elements} on the fitting features.
Let $m_g$ be the mean feature of generator $g$, $m$ the mean of all fitting features, and $\pi_g=n_g/n$.
The within-generator scatter is $\Sigma_W=\sum_g\pi_g\hat\Sigma_g$, where $\hat\Sigma_g$ is the covariance of generator $g$ shrunk toward its diagonal with the Ledoit-Wolf intensity~\cite{ledoit2004shrinkage} (\insupp{sec:supp_subspace}).
The between-generator scatter is $S_B=\hat\Sigma_T-\Sigma_W$, where $\hat\Sigma_T$ is the shrunk covariance of all fitting features. Without shrinkage this is $\sum_g\pi_g(m_g-m)(m_g-m)^\top$.
The generator subspace $W=[w_1,\dots,w_{M-1}]\in\mathbb{R}^{d\times(M-1)}$ consists of the top $M-1$ solutions of
\begin{equation}
  S_B\,w=\lambda\,\Sigma_W\,w,
  \label{eq:lda}
\end{equation}
ordered by $\lambda$, the ratio of between-generator to within-generator variance along $w$.

Whitening the full feature space, as the Mahalanobis score~\cite{lee2018mahalanobis} does, enlarges a shift of dataset or processing along directions in which each generator barely varies. The generator subspace drops such directions unless they separate generator means (\cref{sec:exp_discovery}).

\subsection{Set fingerprint}
\label{sec:method_fingerprint}

The fingerprint of a set is the mean of its projected features.
Let $m_0=W^\top m$ be the projected mean of all fitting outputs.
Registered generator $g$ has the gallery fingerprint
\begin{equation}
  c_g=\frac{1}{n_g}\sum_{x\in X_g}W^\top\phi(x)-m_0 .
  \label{eq:gallery}
\end{equation}
A change in dataset or processing can shift the test features together.
We remove this common shift with the mean of the pool $P$, which is disjoint from the test sets. This pool centering is a first-moment version of the feature alignment used in domain adaptation~\cite{sun2016coral}.
The pool can be small and does not need outputs from unseen generators. In the main tests it contains only outputs from registered generators, and using 50 images changes the discovery AUROC by 0.01 (\insupp{tab:supp_poolchannel}).
With $m_P$ the mean of $W^\top\phi(x)$ over the pool, the set fingerprint is
\begin{equation}
  f(S)=\frac{1}{k}\sum_{x\in S}W^\top\phi(x)-m_P .
  \label{eq:fingerprint}
\end{equation}
Averaging over $k$ outputs reduces variation from content and sampling while retaining the generator offset.

\subsection{Attribution, discovery and family tree}
\label{sec:method_readouts}

\noindent\textbf{Attribution and discovery.}
LineagePrint (LDA) assigns a set to the generator with the nearest gallery fingerprint. Its discovery score is the squared distance to that fingerprint:
\begin{equation}
  \hat g(S)=\arg\min_{g} d_g,\; s(S)=\min_{g} d_g,\; d_g=\lVert f(S)-c_g\rVert^2 .
  \label{eq:attr}
\end{equation}
For family attribution, we fit the subspace with family labels. The gallery then holds one fingerprint per family, and \cref{eq:attr} assigns the family.
Our second variant, LineagePrint (PLDA), replaces the distance with a two-covariance probabilistic LDA (PLDA) model~\cite{ioffe2006plda}.
It scores $s(S)=-\max_g\ell_g(S)$, where $\ell_g(S)$ is the log-likelihood ratio between ``$S$ comes from $g$'' and ``$S$ comes from an unseen generator'', and attributes $S$ to $\arg\max_g\ell_g(S)$.
In our experiments, we use the distance for attribution and for comparing fingerprints in family trees, merges, and verification. We use PLDA as the default discovery score because its results vary less across pools and set sizes (\insupp{tab:supp_poolchannel}).

\noindent\textbf{Family tree.}
We compute one fingerprint $f_u$ for each model $u$ from all its outputs. We compare two models using the fingerprint distance $D(u,v)=\lVert f_u-f_v\rVert$.
A descendant is attributed to its nearest registered base model with \cref{eq:attr}.
For a weight merge $q$ of parents $u$ and $v$, we project its fingerprint onto the segment between the parent fingerprints:
\begin{equation}
  t(q)=\frac{\langle f_q-f_u,\,f_v-f_u\rangle}{\lVert f_v-f_u\rVert^2},
  \label{eq:merge}
\end{equation}
so $t=0$ is parent $u$ and $t=1$ is parent $v$. We compare $t$ with the mixing weight $w$ that the merge gives parent $v$.
The family-tree subspace keeps the top 64 directions of the generator means of the Community Forensics small release, fit without the generators that match a Family-Tree model, and whitens them (\insupp{sec:supp_tree_metric}).

\subsection{Calibrated discovery}
\label{sec:method_calibration}

We flag a set as unseen if its discovery score exceeds a threshold $\tau$.
A pipeline can move the scores of registered generators, so we calibrate $\tau$ on the anchor sets.
For each anchor set $\tilde S_i$, we draw a registered generator uniformly at random and then sample $k$ of its outputs. These outputs do not overlap with the pool or the test sets.
Given the anchor scores $s(\tilde S_1),\dots,s(\tilde S_N)$ and a target false-alarm rate $\alpha$, split-conformal calibration~\cite{vovk2005conformal} sets
\begin{equation}
  \tau=s_{(j)},\qquad j=\lceil (N+1)(1-\alpha)\rceil ,
  \label{eq:tau}
\end{equation}
where $s_{(j)}$ is the $j$-th smallest score ($\alpha=0.05$, $N$ from 20 to 216 by test).

\noindent\textbf{Proposition 1 (False-alarm control, proof \insupp{sec:supp_proof}).}
\textit{Let $S$ be a set from a registered generator. If the scores $s(\tilde S_1),\dots,s(\tilde S_N),s(S)$ are exchangeable, then $P\big(s(S)>\tau\big)\leq\alpha$.}

Anchor and test sets drawn the same way from the registered generators and the same pipeline are exchangeable. For each pipeline with anchor sets, the bound holds over the registered generators and the sampled sets, not separately for each generator.
\section{Experiments}
\label{sec:experiments}

We test three claims.
LineagePrint discovers unseen generators across datasets and processing, at a chosen false-alarm rate (\cref{sec:exp_discovery}).
Fingerprints run in families, up to a boundary (\cref{sec:exp_tree}).
The decoder and the sampling resolution carry part of the family signal (\cref{sec:exp_controls}).

\subsection{Setup}
\label{sec:exp_setup}

\noindent\textbf{Data.}
We build two datasets of generated images that differ in captions and rendering setup. In the COCO dataset, 12 text-to-image generators render captions of COCO~\cite{lin2014coco} photographs. In the Flickr dataset, 12 generators render captions of Flickr30k~\cite{young2014flickr30k} photographs. Ten generators are shared across the two datasets.
Our Family-Tree set contains 13 public models and 5 weight merges with known relationships. We also use 15 newer open models from the FLUX.1, SD3.5, Z-Image, and SANA families to render the same 400 COCO captions.
CF-Small holds 1,951 community generators from Community Forensics~\cite{park2025community}.

\noindent\textbf{Features.}
We use the frozen penultimate features of the Community Forensics (CF) ViT-S/16 detector~\cite{park2025community}, a real-versus-generated classifier trained without generator labels.
Images are resized to 256 pixels before encoding. The 15 newer models and AuraFlow are not among its training generators. The detector saw outputs from the CF-Small generators and 12 of the 13 public Family-Tree models, labeled only as ``generated''.

\noindent\textbf{Baselines.}
Four standard out-of-distribution (OOD) scores use the same features and fitting images: MSP~\cite{hendrycks2017msp} and Energy~\cite{liu2020energy} of a logistic-regression classifier over the registered generators, kNN distance~\cite{sun2022knn} and Mahalanobis distance~\cite{lee2018mahalanobis}.
Per-image baselines average the score over the set.
Set-mean baselines score the mean feature of the set after the pool mean is subtracted. These baselines use the same averaging and centering as LineagePrint. This lets us isolate the effect of the generator subspace and its score.

\noindent\textbf{Metrics.}
Discovery AUROC measures how well we separate sets from unseen and registered generators (chance: 0.50).
At a threshold set for $\alpha=0.05$, we report the false-alarm rate (FPR) and the detection rate (TPR), the fraction of unseen-generator sets flagged.

\subsection{Discovery across datasets and processing}
\label{sec:exp_discovery}

\begin{table}[t]
\centering
\caption{\textbf{Discovery of unseen generators.} CF features, sets of $k=8$, pool of registered generators only, chance 0.50. \emph{29 models}: unseen models from 11 families against 12 registered generators on 400 fresh captions, 36 sets each. \emph{12 generators}: fit in the COCO dataset, test in the Flickr dataset, 12 families, 120 sets each. New / All: mean over the six pipelines never used in fitting / all ten. Best in bold, second best underlined. Intervals \insupp{tab:supp_fresh_scores,tab:supp_unseen_scores}.}
\label{tab:unseen}
\fontsize{8}{9.5}\selectfont
\setlength{\tabcolsep}{2.2pt}
\setlength{\aboverulesep}{0pt}\setlength{\belowrulesep}{0pt}\renewcommand{\arraystretch}{1.14}
\begin{tabular}{@{}lcccccc@{}}
\toprule
& \multicolumn{6}{c}{Discovery AUROC $\uparrow$} \\
\cmidrule(l){2-7}
& \multicolumn{3}{c}{29 models} & \multicolumn{3}{c}{12 generators} \\
\cmidrule(lr){2-4}\cmidrule(l){5-7}
Method & Clean & New & All & Clean & New & All \\
\midrule
\rowcolor{TableGroup}\multicolumn{7}{@{}l}{\textit{Per-image scores, averaged over the set}} \\
MSP~\cite{hendrycks2017msp} & 0.77 & 0.61 & 0.66 & 0.84 & 0.68 & 0.72 \\
Energy~\cite{liu2020energy} & 0.80 & 0.65 & 0.69 & 0.91 & 0.73 & 0.77 \\
Energy, centered & 0.73 & 0.70 & 0.71 & 0.82 & 0.76 & 0.77 \\
Mahalanobis~\cite{lee2018mahalanobis} & 0.57 & 0.61 & 0.62 & 0.62 & 0.60 & 0.61 \\
kNN~\cite{sun2022knn} & 0.57 & 0.54 & 0.55 & 0.58 & 0.58 & 0.58 \\
\midrule
\rowcolor{TableGroup}\multicolumn{7}{@{}l}{\textit{Set-mean baselines: the same scores on the centered set mean}} \\
MSP & \underline{0.82} & 0.82 & 0.82 & \underline{0.94} & \underline{0.89} & \underline{0.90} \\
Energy & 0.81 & 0.81 & 0.81 & 0.93 & \underline{0.89} & \underline{0.90} \\
Mahalanobis & 0.77 & 0.75 & 0.76 & 0.75 & 0.67 & 0.69 \\
Nearest centroid & 0.77 & 0.75 & 0.76 & 0.68 & 0.66 & 0.66 \\
\midrule
\rowcolor{TableOurs}LineagePrint (LDA) & 0.81 & \underline{0.86} & \underline{0.86} & 0.87 & 0.88 & 0.88 \\
\rowcolor{TableOurs}LineagePrint (PLDA) & \textbf{0.90} & \textbf{0.88} & \textbf{0.89} & \textbf{0.97} & \textbf{0.93} & \textbf{0.94} \\
\bottomrule
\end{tabular}
\end{table}
 
\begin{figure}[t]
  \centering
  \includegraphics[width=\columnwidth]{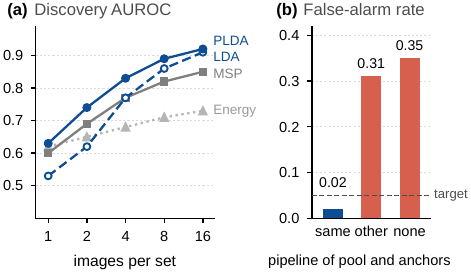}
  \caption{\textbf{Discovery against set size and anchor source.}
  The 29 unseen models, sets of $k=8$ in (b), mean over ten pipelines.
  (a) Discovery AUROC rises with the number of images per set for every score, and LineagePrint (PLDA) stays ahead of the best set-mean baseline (MSP) and the best per-image score (Energy). The dashed line is LineagePrint (LDA).
  (b) The false-alarm rate of LineagePrint (PLDA) stays at the 0.05 target (dashed) only when the pool and the anchor sets come from the same pipeline as the query, and not when they come from the nearest other pipeline or from unprocessed images (none).}
  \label{fig:discovery_main}
\end{figure}

We first test whether LineagePrint discovers generators it has never seen when the captions, the dataset and the processing change.
Both tests use sets of $k=8$ images and ten processing pipelines. Clean images, JPEG 70 and blur are fitting conditions, JPEG 80 is close to them, and six pipelines are never used in fitting.
In the first test, we fit on the COCO dataset and score 29 unseen models from 11 families against the 12 registered generators using 400 new captions.
In the second test, we fit on COCO and test on Flickr. The 12 unseen generators include AuraFlow, SD3.5, and the ten shared generators, which we hold out of the fit one at a time.

\begin{figure}[t]
  \centering
  \includegraphics[width=\columnwidth]{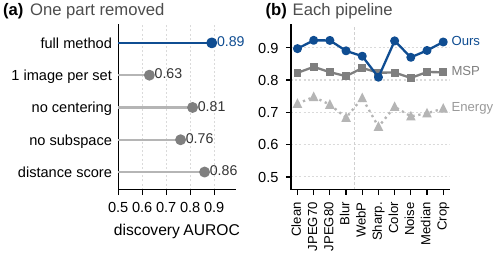}
  \caption{\textbf{What each part adds, and each pipeline.}
  The 29 unseen models, sets of $k=8$.
  (a) Discovery AUROC over ten pipelines with one part of the method removed.
  (b) Discovery AUROC on each pipeline for LineagePrint (PLDA, Ours), the best set-mean baseline (MSP) and the best per-image score (Energy). The six pipelines right of the dotted line are never used in fitting.}
  \label{fig:ablation}
\end{figure}

\noindent\textbf{Unseen generators are discovered.}
LineagePrint (PLDA) outperforms every out-of-distribution baseline in both tests, on clean images and across the six new pipelines (\cref{tab:unseen}).
On the 29 unseen models it reaches a discovery AUROC of \good{0.89}.
Set averaging and centering raise the baseline AUROC from 0.69 to 0.82. LineagePrint improves it by another 0.07.
It is ahead of every other score on nine of the ten pipelines and 0.01 behind MSP on the set mean under sharpening (\cref{fig:ablation}b).
The lead grows with the number of images per set (\cref{fig:discovery_main}a) and holds on the second dataset, where LineagePrint reaches 0.93 on the six new pipelines against 0.89.
A second run on 400 new captions, with the comparison fixed in advance, confirms the lead (0.86 against 0.76, \insupp{sec:supp_confirm}).
The closed-form LineagePrint method performs about as well as a prototype classifier trained on the same features.
On native-resolution crops of 25 of these models LineagePrint (LDA) reaches 0.94, against at most 0.79 for the set-averaged noise residual~\cite{marra2019gans}, DIF~\cite{sinitsa2024deep} and Forensic Self-Descriptions~\cite{nguyen2025forensic}.

\noindent\textbf{What each part adds.}
Removing one part of LineagePrint at a time shows that all four matter (\cref{fig:ablation}a).
With a single image instead of a set of eight the AUROC drops from 0.89 to 0.63, without the pool centering to 0.81, with raw features in place of the generator subspace to 0.76, and with the distance in place of the PLDA score to 0.86.
The subspace makes a larger difference across datasets. On the second dataset, AUROC rises from 0.66 with raw features to 0.88 with the subspace.

\noindent\textbf{Beyond these two tests.}
The gain does not rely on the encoder having seen the generators.
On the Community Forensics evaluation set (CF-Eval)~\cite{park2025community}, which holds 21 generators outside the CF detector's training set, 15 of them registered, LineagePrint (LDA) reaches a discovery AUROC of 0.97 and attributes 0.99 of the registered sets to the right generator.
On ASVspoof 2021 LA~\cite{liu2023asvspoof2021}, with anti-spoofing features~\cite{tak2022wav2vec} and sets of 16 utterances, LineagePrint (PLDA) reaches 0.83 over five codecs, a telephone line and uncompressed audio, compared with 0.78 for the best baseline (\insupp{sec:supp_speech}).

\noindent\textbf{False alarms stay within the target rate.}
\label{sec:exp_calibration}
With anchor sets from the test pipeline the FPR is \good{0.02} for the 0.05 target and stays between 0.01 and 0.05 on each of the ten pipelines.
At this threshold LineagePrint flags 0.66 of the unseen-model sets, compared with 0.52 for the best set-mean baseline.
With a pool and anchors from clean images the FPR rises to \bad{0.35} (\cref{fig:discovery_main}b), so both must share the processing of the query.

\noindent\textbf{Unseen descendants.}
Of the 29 models, 17 derive from a registered generator or are its sibling.
Seven of them, such as FLUX.1-lite-8B and SD1.4, are attributed to that relative in most of their sets, and the other ten are mostly flagged as new.

\noindent\textbf{From discovery to enrollment.}
Sets that are flagged as unseen can be enrolled without any label.
We cluster the flagged sets and add each cluster to the gallery as a new fingerprint.
On 19 generators outside the gallery, the clusters have a purity of 0.77. Later sets are assigned to the cluster for their own generator with an accuracy of 0.76.

\subsection{Fingerprints run in families}
\label{sec:exp_tree}

\begin{figure*}[t]
  \centering
  \includegraphics[width=\textwidth]{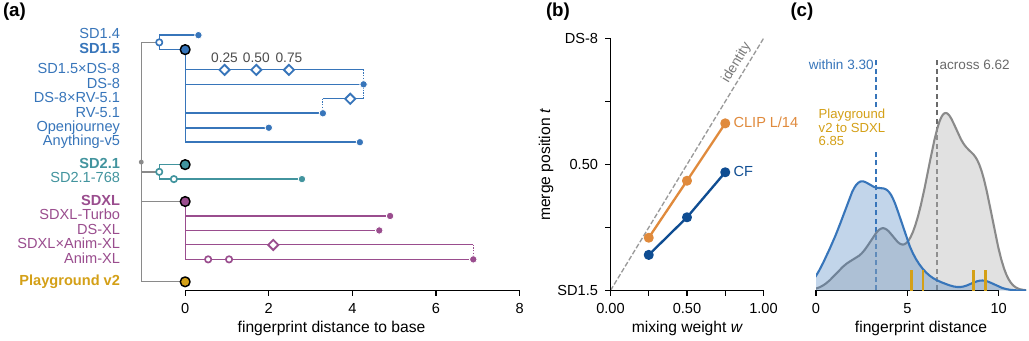}
  \caption{\textbf{Fingerprints of a model family tree.}
  (a) Documented tree of the 18 Family-Tree models, each at its measured fingerprint distance to its family's base (Playground v2 gold, bases ringed, hollow circles unsampled checkpoints placed schematically, merges as diamonds on a bar between their parents).
  (b) Position $t$ of each SD1.5$\times$DreamShaper-8 (DS-8) merge vs its mixing weight $w$ (CF and CLIP ViT-L/14 features, identity dashed).
  (c) Pair distances within a family, across families, and from Playground v2 to SDXL (ticks).
  DS: DreamShaper, RV: RealisticVision, Anim-XL: Animagine-XL-3.1. CF unless stated, $k=16$.}
  \label{fig:tree}
\end{figure*}

Next, we test whether LineagePrint can trace descendants to their base models.

\noindent\textbf{A controlled family tree.}
The Family-Tree set holds the SD1, SD2 and SDXL families and Playground v2~\cite{playground2023v2}, with SD1.5, SD2.1, SDXL and Playground v2 registered (sets of $k=16$).
The mean fingerprint distance is 3.30 within a family and 6.62 across families (\cref{fig:tree}a,c). The distance also correlates with the number of edges between two models in the documented tree (Spearman correlation 0.46 over 153 pairs).
We take the kind of each descendant from its model card: a heavy style fine-tune is trained toward a distinct visual style, such as anime or photorealism.
The eight light descendants, a sibling version, a 768-pixel relative, a distilled model, a light fine-tune and four merges with a base parent, are attributed to the base of their family in \good{551 of 576} sets.
Merges of SD1.5 and DreamShaper-8 lie between their parents in the order of their mixing weight, at positions 0.14, 0.29 and 0.47 for weights 0.25, 0.50 and 0.75, nearer the parent whose autoencoder they keep (\cref{fig:tree}b).

\noindent\textbf{Where inheritance stops.}
LineagePrint attributes the five heavy style fine-tunes in the tree to their base models in only \bad{1 of 360} sets.
The contrast is clearest in the SDXL family. LineagePrint attributes every set from the light fine-tune DreamShaper-XL to SDXL, but no sets from the heavy style fine-tune Animagine-XL-3.1 (\cref{tab:families}c).
Playground v2 has the architecture and the autoencoder of SDXL, was trained from scratch, and lies as far from the SDXL family (6.85) as from SD1 (7.02) and SD2 (6.46, \cref{fig:tree}c).
A shared architecture and autoencoder thus do not make a shared fingerprint, whereas in GANs architecture traces survive retraining~\cite{yang2022dnadet}.

\begin{table*}[t]
\centering
\caption{\textbf{Descendants in three families.} CF features, sets of $k=16$ images, 72 sets per model, 17 registered bases. To base: share of a model's sets whose nearest registered fingerprint is the base of its family. $d_\mathrm{base}$ and $d_\mathrm{other}$: fingerprint distance to that base and to the nearest other base. Green: attributed to the base in at least 0.80 of the sets, the rule fixed before the run. Red: attributed to another base. $^\dagger$Chosen before the run. Relations come from the model cards.}
\label{tab:families}
\small
\setlength{\tabcolsep}{2pt}
\renewcommand{\arraystretch}{1.15}
\begin{tabular*}{\textwidth}{@{\extracolsep{\fill}}llccc@{\hspace{10pt}}llccc@{}}
\toprule
\multicolumn{5}{@{}l}{\textbf{(a) FLUX.1 family}, base FLUX.1-dev} & \multicolumn{5}{l@{}}{\textbf{(b) SD3.5 family}, base SD3.5-large} \\
\cmidrule(r{8pt}){1-5}\cmidrule(l{8pt}){6-10}
Descendant & Relation & To base $\uparrow$ & $d_\mathrm{base}$ $\downarrow$ & $d_\mathrm{other}$ & Descendant & Relation & To base $\uparrow$ & $d_\mathrm{base}$ $\downarrow$ & $d_\mathrm{other}$ \\
\midrule
FLUX.1-lite-8B$^\dagger$ & distilled & \good{0.96} & 1.47 & 3.38 & SD3.5-large-TurboX$^\dagger$ & few-step & \good{0.99} & 2.63 & 4.09 \\
FLUX.1-dev + Turbo-Alpha$^\dagger$ & fused LoRA & \good{0.90} & 2.01 & 3.23 & SD3.5-medium & sibling & \good{0.96} & 2.62 & 4.25 \\
FLUX.1-schnell & sibling & \good{0.96} & 1.76 & 3.48 & SD3.5-medium-turbo & sibling, distilled & 0.43 & 4.94 & 4.82 \\
FLUX.1-dev $\times$ Krea-dev & merge & \good{0.99} & 1.91 & 3.97 & SD3.5-large-turbo & distilled & \bad{0.01} & 5.19 & 4.02 \\
\cmidrule(l{8pt}){6-10}
FLUX.1-Krea-dev & variant & \bad{0.00} & 4.73 & 3.07 & \multicolumn{5}{l@{}}{\textbf{(c) SDXL family}, base SDXL} \\
AWPortrait-FL & fine-tune & \bad{0.04} & 4.11 & 3.28 & DreamShaper-XL & light fine-tune & \good{1.00} & 4.64 & 6.54 \\
Flex.1-alpha & from schnell & \bad{0.00} & 5.27 & 2.78 & Animagine-XL-3.1 & heavy fine-tune & \bad{0.00} & 6.89 & 4.56 \\
\bottomrule
\end{tabular*}
\end{table*}
 
\noindent\textbf{Newer families.}
\label{sec:exp_scale}
FLUX.1~\cite{bfl2024flux} and SD3.5~\cite{esser2024sd3}, which the CF detector never saw, show the same pattern with 17 bases registered (\cref{tab:families}).
Mean within-family distances are 3.61 for FLUX.1 and 4.60 for SD3.5, compared with distances of 7.93 and 7.42 to models from other families, respectively.
Three few-step or distilled descendants chosen before the run are attributed to their base in \good{205 of 216} sets, and FLUX.1-schnell, SD3.5-medium and a merge of FLUX.1-dev with FLUX.1-Krea-dev in at least 0.96 of theirs.
By contrast, three other relatives of FLUX.1-dev and the distilled SD3.5-large-turbo are attributed to another base, and SD3.5-medium-turbo reaches its base in 0.43 of its sets.

\noindent\textbf{Family verification.}
We also score each set from the 25 relatives against every registered base.
The fingerprint distance verifies the documented family with AUROC \good{0.92}, compared with 0.82 for the same distance on CLIP ViT-L/14 features, and this includes the relatives whose nearest base is another one.
At a larger scale, among 495 community fine-tunes with a documented SD1/SD2 or SDXL base, unseen fine-tunes are attributed to their family with top-1 accuracy 0.87, on par with linear classifiers. Within one dataset the raw set mean and standard distances on it separate the two families as well (\insupp{tab:lineage}).

\subsection{What carries the family signal}
\label{sec:exp_controls}

Finally, we examine which part of a generator carries the family fingerprint.
To this end, we change one component of a generator at a time and attribute its 24 sets of $k=16$ images again (\insupp{tab:controls}).

\noindent\textbf{Decoder.}
The decoder affects whether sets are attributed to SD1 or SD2.
When we decode SD1.5 latents with the SD2.1 decoder, all 24 sets are attributed to SD2.1. In the reverse swap, only 1 of 24 sets retains its attribution.
Openjourney, an SD1.5 fine-tune that ships another decoder, returns to SD1.5 in all 24 sets once decoded with the SD1.5 decoder, and DreamShaper-8 in 11 of 24.

\noindent\textbf{Sampling resolution.}
At 512 pixels, none of the sets from SDXL, DreamShaper-XL, or Playground v2 keep their original attribution. Their fingerprints move toward SD1.5 and SD2.1.
Registering a base at the query resolution repairs this: with FLUX.1-dev and SD3.5-large also registered at 512 and 768 pixels, four of their relatives sampled at these resolutions reach the base of their family in 182 of 192 sets (\insupp{tab:supp_resolution}).

\noindent\textbf{Autoencoder.}
After a round trip through another family's autoencoder, 19 to 24 of 24 sets retain their family attribution for SDXL, FLUX.1, and SD3.5. This suggests that the autoencoder alone does not account for their fingerprints.
After a round trip through each other's autoencoder, SD1.5 and SD2.1 retain their attribution in only 3 and 1 of 24 sets, respectively.

\noindent\textbf{Base model against style.}
In controlled LoRA~\cite{hu2022lora} fine-tuning of SD1.5 and SDXL on photographs and anime, the base model explains 0.61 of the variance of forensic features and the style 0.24. With CLIP features the shares reverse (0.24 and 0.57).
With CLIP features, the anime fine-tunes Anything-v5 (of SD1.5) and Animagine-XL-3.1 (of SDXL) are closer to each other than to their own base models.
Thus, forensic features follow the base model, whereas semantic features follow the style.

\subsection{What this means for provenance}
\label{sec:exp_implications}

A registry should hold one fingerprint for each base model and sampling resolution, and record the decoder a model ships with, because the sampling resolution and decoder can both shift the fingerprint.
A set from an unknown source should be centered and calibrated with outputs of the registered generators that went through the same upload pipeline.
A flagged set that lies close to a registered base is a candidate descendant of that base, and one that lies far from every base is enrolled as a new branch (\insupp{sec:supp_enroll}).
\section{Conclusion}
\label{sec:conclusion}

LineagePrint identifies generators from sets of outputs using fingerprints computed in a subspace that separates registered generators.
On 8-image sets from 29 unseen models across 11 families, LineagePrint reaches an AUROC of 0.89 over ten pipelines. The strongest out-of-distribution baseline reaches 0.82 on the centered set mean. LineagePrint also detects unseen speech generators across codecs, and the false-alarm rate stays close to the target.
LineagePrint usually attributes light descendants to their base models, but rarely does so for heavy style fine-tunes. Merges project between their parents, and the distance to a base model verifies family membership with an AUROC of 0.92.
Flagged sets can be enrolled as new branches (\insupp{sec:supp_enroll}) to expand the gallery for provenance, licensing, and platform integrity.

\noindent\textbf{Limitations.}
The discovery tests cover three caption sets with one run each, and the speech evidence covers one corpus.
The lineage experiments cover latent text-to-image families with relations from model cards.
A base fingerprint is specific to the sampling resolution.
The pool and the anchor sets must be generated outputs from the test pipeline, and a pool of real photographs gives discovery AUROC 0.37 (\insupp{tab:supp_pool}).
The false-alarm bound is marginal over the registered generators, and a skewed mix of generators calls for class-conditional calibration~\cite{ding2023classconditional}.
We assume each query set comes from one generator. In sets of eight, one image from another generator lowers AUROC by 0.01, and two raise the false-alarm rate of LineagePrint (PLDA) to about 0.26 (\insupp{tab:supp_mixed}).
The lead in discovery holds with forensic features and shrinks or reverses with CLIP and DINOv2 features (\insupp{tab:supp_encoders}).
We compare against one published open-set attribution system, Forensic Self-Descriptions.

\ifwithsupp
\clearpage
\ifdefined\arxivversion\appendix\fi %
\raggedbottom %
\setcounter{section}{0}\setcounter{figure}{0}\setcounter{table}{0}\setcounter{equation}{0}
\renewcommand{\thesection}{\Alph{section}}
\renewcommand{\thefigure}{S\arabic{figure}}
\renewcommand{\thetable}{S\arabic{table}}
\renewcommand{\theequation}{S\arabic{equation}}
\ifdefined\theHsection\renewcommand{\theHsection}{supp.\Alph{section}}\fi
\ifdefined\theHfigure\renewcommand{\theHfigure}{supp.\arabic{figure}}\fi
\ifdefined\theHtable\renewcommand{\theHtable}{supp.\arabic{table}}\fi
\ifdefined\theHequation\renewcommand{\theHequation}{supp.\arabic{equation}}\fi
\ifdefined\arxivversion\section*{Appendix}\else\section*{Supplementary Material}\fi
\noindent This supplement gives the implementation details (\cref{sec:supp_impl}), the datasets and generators (\cref{sec:supp_data}), the discovery tests in full, with every score and its interval, the pixel fingerprints at 256 pixels and on native crops, every baseline under equal treatment and the pool study (\cref{sec:supp_results}), the confirmation on fresh captions (\cref{sec:supp_fresh}), the speech experiment (\cref{sec:supp_speech}), three studies of what moves a fingerprint (\cref{sec:supp_moves}), the heavy fine-tunes of the Stable Diffusion family tree (\cref{sec:supp_tree}), every model of the FLUX.1 and SD3.5 experiment (\cref{sec:supp_modern}), every condition of the autoencoder and resolution controls (\cref{sec:supp_controls}), the discovery and enrollment of unseen generators (\cref{sec:supp_enroll}) and the related work in full (\cref{sec:supp_related}).

\section{Implementation details}
\label{sec:supp_impl}

Every step of LineagePrint is closed form on cached features.
Every random choice (splits, generator subsets, set draws) uses a seed derived from SHA-256 hashes of identifiers, and the same seeds are shared across encoders and methods.

\subsection{Encoders and features}
\label{sec:supp_features}

\noindent\textbf{Images.}
Every image, generated or real, is converted to RGB and resized to $256\times256$ pixels with bicubic interpolation.
Processing pipelines act on this image, and each encoder then applies its own preprocessing.
CF features are the 384-dimensional penultimate activations of the released Community Forensics ViT-S/16 detector~\cite{park2025community}, after resizing the short side to 440 pixels, a center crop of 384 pixels and ImageNet normalization.
CLIP ViT-B/32 and ViT-L/14~\cite{radford2021clip} give 512- and 768-dimensional image embeddings, DINOv2-L~\cite{oquab2024dinov2} its 1,024-dimensional class token, and SigLIP so400m at 384 pixels~\cite{zhai2023siglip,alabdulmohsin2023shape} a 1,152-dimensional image embedding.
Features are extracted once, and all methods read the same cached features.

\noindent\textbf{What the encoders saw.}
The CF encoder (\nolinkurl{OwensLab/commfor-model-384}) is trained end to end as a binary classifier of real against generated images with a single sigmoid output, so its training used no generator labels~\cite{park2025community}.
It was trained on the 4,782 generators of the Community Forensics training set, 4,763 latent diffusion models from Hugging Face and 19 hand-picked models, and the 21 CF-Eval generators are not among them~\cite{park2025community}.
Neither unseen generator of the cross-dataset test, AuraFlow and SD3.5, is among them, nor are 7 of the 10 registered generators (PixArt-$\Sigma$, SANA, Hunyuan-DiT, Lumina-Image 2.0, Kolors, CogView4 and FLUX.1-dev).
SD1.5 and SDXL are among them, and so is a third-party upload named \texttt{playground-v25}.
None of the 15 newer models of \cref{tab:supp_modern_models} is among them, and the training list holds no generator with a FLUX or SD3 pipeline.
CF-Small is drawn from the training images of this detector, so all 1,951 CF-Small generators are among its training generators, and so are 12 of the 13 public Family-Tree models (RealisticVision-5.1 appears only as third-party copies).
In the lineage-at-scale and Family-Tree experiments, unseen therefore means absent from the LineagePrint gallery and fit, while the encoder saw these generators under the single label ``generated''.
CLIP ViT-B/32, CLIP ViT-L/14, DINOv2-L and SigLIP were pretrained on web images without real/generated or generator labels.
For speech, the AASIST network was trained to separate bona fide from spoofed speech on ASVspoof 2019 LA (systems A01 to A06), none of which is among the 11 unseen 2021 systems, and wav2vec 2.0 base was trained on LibriSpeech without spoofed speech.

\noindent\textbf{Speech.}
Each utterance contributes its first 64,600 samples (about 4 seconds at 16\,kHz), repeated when it is shorter.
The feature is the 160-dimensional penultimate layer of the frozen anti-spoofing network of \citet{tak2022wav2vec}, an XLS-R~\cite{babu2022xlsr} wav2vec 2.0~\cite{baevski2020wav2vec2} front end with an AASIST back end~\cite{jung2022aasist}, trained on ASVspoof 2019 LA (checkpoint \nolinkurl{SpeechAntiSpoofingBenchmarks/W2V2-AASIST}).
We standardize the features and reduce them to at most 128 dimensions with principal component analysis (PCA), both fit on the ASVspoof 2019 LA training utterances only.

\subsection{Generator subspace and scores}
\label{sec:supp_subspace}

\noindent\textbf{Shrinkage LDA.}
To keep the covariances well conditioned, we shrink each one toward its diagonal,
\begin{equation}
  \hat\Sigma_g=(1-\rho_g)\,C_g+\rho_g\,\operatorname{diag}(C_g),
  \label{eq:shrink}
\end{equation}
where $C_g$ is the sample covariance of the features of generator $g$.
Without shrinkage, the between-generator scatter of \cref{eq:lda} is $S_B=\sum_g\pi_g(m_g-m)(m_g-m)^\top$.
The intensity $\rho_g\in[0,1]$ is the Ledoit-Wolf estimate~\cite{ledoit2004shrinkage} on the standardized residuals of generator $g$, computed in blocks of 128 features.
With shrinkage, $S_B=\hat\Sigma_T-\Sigma_W$ is this scatter shrunk toward its diagonal with the intensity $\rho_T$ of $\hat\Sigma_T$, plus $\sum_g\pi_g(\rho_g-\rho_T)\,(C_g-\operatorname{diag}(C_g))$.
Each covariance is stored as a low-rank term plus a diagonal, so no $d\times d$ matrix is formed.
We solve \cref{eq:lda} with a Lanczos eigensolver and conjugate-gradient inverses, and add a ridge of $10^{-10}$ times the mean diagonal for numerical stability.
The subspace keeps $\min(M-1,d)$ directions, and image features enter the fit at full dimension.

\noindent\textbf{PLDA.}
LineagePrint (PLDA) whitens the LDA projection by the within-generator covariance of the projected fitting outputs and scales each projected output to a fixed length~\cite{garciaromero2011length}.
From these embeddings it estimates a between-generator covariance $\Phi_b$ and a within-generator covariance $\Phi_w$, each with a ridge of $10^{-6}$.
For a set of $k$ outputs, $\ell_g(S)$ is the log-likelihood ratio of the set mean under ``the set comes from generator $g$'' against ``the set comes from an unseen generator drawn from $\Phi_b$'', with the within-generator noise of a mean of $k$ outputs.
We compute all ratios after a simultaneous diagonalization of $\Phi_b$ and $\Phi_w$.

\noindent\textbf{Calibration.}
The threshold of \cref{eq:tau} uses the strict rule $s(S)>\tau$.
For images, each anchor set first draws a registered generator uniformly at random among those with at least $2k$ anchor outputs and then draws $k$ of its outputs.

\noindent\textbf{Proof of Proposition 1.}
\label{sec:supp_proof}
The probability in Proposition~1 is over the draw of the anchor and test sets, and so over the registered generator a set comes from, as in conformal outlier testing~\cite{bates2023outliers}. The guarantee is therefore marginal over the registered generators. Control for each generator separately would need class-conditional calibration~\cite{ding2023classconditional}.
By exchangeability $s(S)$ is equally likely to hold each of the $N+1$ ranks among the $N+1$ scores (ties only lower the chance of exceeding $\tau$), and $s(S)>\tau$ needs a rank above $j$, which has probability at most $(N+1-j)/(N+1)\leq\alpha$.\hfill$\square$

\subsection{Family-tree distance}
\label{sec:supp_tree_metric}

The family-tree subspace is fit on the images of the generators named in the CF-Small shards (up to 96 images each, \cref{sec:supp_cf}).
Before the fit we remove every generator whose name, ignoring case and punctuation, contains the name of a Family-Tree model, such as \texttt{runwayml/stable-diffusion-v1-5} or \texttt{frankjoshua/sdxl-turbo}.
This rule removes 64 of the 2,253 names, and the fit uses the remaining 2,189.
These images are disjoint from the Family-Tree images.
We standardize the features, center the generator means and keep the top 64 right singular vectors of the matrix of generator means.
In this 64-dimensional space we estimate the within-generator covariance, shrink it toward its diagonal with weight 0.15, add a ridge of $10^{-5}$ times its mean variance and whiten by its inverse square root.
The fingerprint distance of \cref{sec:exp_tree} is the Euclidean distance in this whitened space, in which the pool mean of \cref{eq:fingerprint} cancels.

We also use a label-free variant that needs no generator labels.
It keeps the top 64 principal directions of the standardized features of the same 2,189 names and scales each to unit variance.
In the label-free distance the Family-Tree set has a mean distance of 2.78 within a family and 6.06 across families, and its distances agree with the known tree with Spearman correlation 0.48 (CF, $k=16$).
The controlled fine-tuning experiment of \cref{sec:supp_finetune} reports distances in this label-free variant.

\subsection{Sets, splits, centering and anchors}
\label{sec:supp_splits}

A set holds $k$ distinct outputs of one generator.
We never draw a set from a pool of fewer than $2k$ outputs.
Different sets of the same generator can share outputs, except where stated.

\noindent\textbf{The two larger discovery tests.}
Both tests, and the repeat of the first on fresh captions (\cref{sec:supp_fresh}), use the augmented fit, disjoint test sets and a pool drawn from registered generators only.
The 25-model test fits on the COCO-dataset images of the 12 COCO-dataset generators, without the images whose caption is one of the 400 Family-Tree captions or whose COCO image is shared with one of them, 507 to 510 images per generator.
It tests on the 400 captions, where each of the 25 unseen models and each registered generator gives 42 disjoint sets of 8.
The 25 models are the Family-Tree and newer models of \cref{tab:supp_unseen_models} that are neither a checkpoint of a registered generator nor a weight merge of the Family-Tree set.
One of the 25 is our weight merge of FLUX.1-dev and FLUX.1-Krea-dev.
The 12-generator test fits in the COCO dataset on 600 images per generator and tests in the Flickr dataset.
Each of the ten shared generators is held out of the fit in turn and scored against the other nine, and AuraFlow and SD3.5 are never in the fit and are scored against all ten, with 120 disjoint sets per generator.
Before the run we fixed one comparison per test, LineagePrint (LDA) against Energy on the centered set mean, on all ten pipelines for the 25 models and on the six new pipelines for the 12 generators. LineagePrint leads when the 95\% bootstrap interval of the paired difference excludes zero.
The bootstrap resamples unseen generators for the 12 generators, and families and then generators within a family for the 25 models.

\noindent\textbf{Across datasets, two unseen generators.}
Splits are by caption, so every generator and every processed copy of an image fall into the same role.
In the COCO dataset, half of the captions provide the fitting images.
In the Flickr dataset, each split assigns every output of the 12 test generators to the centering pool $P$ (20\%), the anchor outputs (40\%) or the test outputs (40\%).
The pool holds 195, 180 and 213 outputs in the three splits, for every encoder.
The pool study (\cref{tab:supp_pool}) replaces it by the pool outputs of the 10 registered generators only, by 100, 30 or 10 of them, by those of one registered generator, or by 203 to 228 real Flickr30k photographs, each passed through the test pipeline.
Each generator with at least $2k$ test outputs contributes 200 test sets per pipeline and set size.
At $k=16$ this leaves out three to five generators per split.
Each pipeline has 50 anchor sets.

\noindent\textbf{CF-Eval.}
Images are deduplicated by pixel hash across the whole release.
Each split fits on 40\% of the images of every registered generator and tests on a disjoint 40\%.
Each of the 3 splits draws 10 random orders of the 21 generators, and the first $M$ generators of an order are registered, so the galleries are nested in $M$.
Each generator contributes 200 test sets.

\noindent\textbf{Speech.}
Splits are by speaker.
The 2019 LA training utterances fit the subspace.
The target speakers are split into 25\% for centering, 37.5\% for anchors and 37.5\% for test sets, with the same roles in every channel.
We replace the pool mean $m_P$ of \cref{eq:fingerprint} by a labeled shift: the mean difference, per registered generator, between its target-channel utterances from the centering speakers and its fitting utterances, averaged over the two registered generators.
Every baseline receives the same shift.
Within a channel, sets use disjoint utterances and distinct speakers where possible.

\noindent\textbf{Lineage at scale.}
Each generator keeps 32 evaluation images.
A registered generator also contributes 8 separate fitting images.
Each draw registers 32 generators of one family and queries 83 unseen generators of each family with 20 sets per generator and set size.
Registered and query generators are disjoint in every draw.

\noindent\textbf{Family tree.}
Each of 3 splits divides the 400 captions into two halves by hash.
Within a half, the captions are cut into disjoint sets of $k$ images, so each model has 72 sets at $k=16$ (3 splits, 2 halves, 12 sets).
The fingerprint of a model pools all its sets.

\subsection{Baselines and learned heads}
\label{sec:supp_baselines}

\noindent\textbf{Baselines.}
All baselines use the same features, fitting images and test sets as LineagePrint.
MSP~\cite{hendrycks2017msp} and Energy~\cite{liu2020energy} read a multinomial logistic-regression classifier (regularization constant $C=1$) on standardized features.
Their image scores are the negative maximum softmax probability and the negative log-sum-exp of the logits, averaged over the set.
kNN~\cite{sun2022knn} is the Euclidean distance from an image to its 10th nearest registered fitting image, on unit-normalized standardized features, averaged over the set.
Mahalanobis~\cite{lee2018mahalanobis} uses the class means in the original feature space and the Ledoit-Wolf covariance of the within-generator residuals.
Each score has two aggregations: the per-image score averaged over the set, and the score of the set-mean feature.
Each is computed with and without centering, which subtracts the difference between the target-pool mean and the fitting mean, as in LineagePrint.
Standard per-image scores are the uncentered per-image averages.
Set-mean baselines are the set-mean variants, and the strongest of all variants is chosen per column on the test results, which favors the baselines.
The two set-level two-sample scores compare the set with the fitting features of each registered generator and keep the smallest statistic: an RBF-kernel MMD~\cite{gretton2012kernel} with the median-heuristic bandwidth of the fitting features and the unbiased U-statistic, and the energy distance.
For speech, kNN uses the distance of the normalized set mean to its 10th nearest fitting utterance, and every baseline receives the labeled shift of \cref{sec:supp_splits}.
In the lineage experiments the classifiers are multinomial logistic regressions on the same features, one on generator labels, whose family prediction sums its generator posteriors, and one on family labels. Their confidence (MSP or Energy, per image or on the set mean) is the discovery score.

\noindent\textbf{Learned projection heads.}
A head is a linear map to 32 dimensions followed by unit normalization.
We train it on the COCO-dataset images of the 12 generators under the 12 image versions of the augmented fit, for 8 epochs with AdamW (learning rate $2\times10^{-3}$, weight decay $10^{-4}$, batch 512).
The SupCon head is trained with the supervised contrastive (SupCon) loss~\cite{khosla2020supcon}, with generator labels and temperature 0.12.
The invariance variant adds 0.5 times the cosine distance between an image and a processed copy of it.
The ArcFace head uses an additive angular margin of 0.20 and scale 24~\cite{deng2019arcface}.
A set is scored by the distance of its mean embedding to the nearest registered generator mean.

\section{Datasets}
\label{sec:supp_data}

\subsection{COCO and Flickr datasets}
\label{sec:supp_worlds}

\begin{table*}[t]
\centering
\caption{\textbf{Generators of the two datasets.} Checkpoint, sampling steps, classifier-free guidance and output resolution of each generator. The two commercial generators are queried through their API, and their outputs are center-cropped to a square and resized to 1,024 pixels. Datasets: C is the COCO dataset and F the Flickr dataset.}
\label{tab:supp_world_generators}
\fontsize{8}{9.5}\selectfont
\setlength{\tabcolsep}{3pt}
\begin{tabular}{@{}llcccc@{}}
\toprule
Generator & Checkpoint & Steps & Guidance & Resolution & Datasets \\
\midrule
SD1.5 & \texttt{runwayml/stable-diffusion-v1-5} & 30 & 7.5 & 512 & C, F \\
SDXL & \texttt{stabilityai/stable-diffusion-xl-base-1.0} & 30 & 7.0 & 1,024 & C, F \\
PixArt-$\Sigma$ & \texttt{PixArt-alpha/PixArt-Sigma-XL-2-1024-MS} & 20 & 4.5 & 1,024 & C, F \\
SANA & \texttt{Efficient-Large-Model/Sana\_1600M\_1024px\_diffusers} & 20 & 5.0 & 1,024 & C, F \\
Hunyuan-DiT & \texttt{Tencent-Hunyuan/HunyuanDiT-Diffusers} & 20 & 5.0 & 1,024 & C, F \\
Lumina-Image 2.0 & \texttt{Alpha-VLLM/Lumina-Image-2.0} & 30 & 4.0 & 1,024 & C, F \\
Playground v2.5 & \texttt{playgroundai/playground-v2.5-1024px-aesthetic} & 30 & 3.0 & 1,024 & C, F \\
Kolors & \texttt{Kwai-Kolors/Kolors-diffusers} & 30 & 5.0 & 1,024 & C, F \\
CogView4 & \texttt{THUDM/CogView4-6B} & 30 & 3.5 & 1,024 & C, F \\
FLUX.1-dev & \texttt{black-forest-labs/FLUX.1-dev} & 20 & 3.5 & 1,024 & C, F \\
\midrule
Gemini 2.5 Flash Image~\cite{google2025geminiimage} & \texttt{gemini-2.5-flash-image} (API) & - & - & 1,024 & C \\
Gemini 3.1 Flash Image~\cite{google2025geminiimage} & \texttt{gemini-3.1-flash-image} (API) & - & - & 1,024 & C \\
AuraFlow & \texttt{fal/AuraFlow-v0.3} & 25 & 3.5 & 1,024 & F \\
SD3.5 & \texttt{stabilityai/stable-diffusion-3.5-large} & 28 & 3.5 & 1,024 & F \\
\bottomrule
\end{tabular}
\end{table*}

A dataset is one combination of captions, their real images and one rendering setup (\cref{sec:fingerprints}).
The COCO dataset renders captions of COCO~\cite{lin2014coco} validation images, and the Flickr dataset renders one caption for each of 1,200 Flickr30k~\cite{young2014flickr30k} photographs.
\Cref{tab:supp_world_generators} lists the 14 generators.
The ten shared generators are SD1.5~\cite{rombach2022ldm}, SDXL~\cite{podell2024sdxl}, PixArt-$\Sigma$~\cite{chen2024pixartsigma}, SANA~\cite{xie2025sana}, Hunyuan-DiT~\cite{li2024hunyuandit}, Lumina-Image 2.0~\cite{qin2025lumina2}, Playground v2.5~\cite{li2024playground}, Kolors~\cite{kolors2024}, CogView4~\cite{zheng2024cogview3,zai2025cogview4} and FLUX.1-dev~\cite{bfl2024flux}.
Playground v2.5 in the two datasets is a different model from Playground v2 in the Family-Tree set (\cref{sec:supp_tree_set}).
Ten render both datasets, two commercial generators render only the COCO dataset, and AuraFlow~\cite{fal2024auraflow} and SD3.5~\cite{esser2024sd3} render only the Flickr dataset.
The discovery test with AuraFlow and SD3.5 unseen (\cref{tab:discovery}) uses 1,000 generated images per dataset, 83 or 84 per generator.

The augmented fit uses 12 versions of every COCO-dataset fitting image: the clean image, JPEG at quality 90, 70, 50 and 30, Gaussian blur with $\sigma$ 0.5, 1.0 and 1.5, bicubic down- and up-sampling by 0.5 and 0.25, JPEG 50 followed by blur 1.0, and down-sampling by 0.5 followed by JPEG 70.
The caption of \cref{tab:supp_pipelines} defines the 10 test pipelines.

\subsection{CF-Eval and CF-Small}
\label{sec:supp_cf}

\noindent\textbf{CF-Eval.}
The Community Forensics evaluation set~\cite{park2025community} holds 21 generators: two GANs (DF-GAN, GALIP), Hourglass diffusion, Kandinsky 2.2, Stable Cascade, DeciDiffusion v2, FLUX.1 dev and schnell, three latent consistency LoRAs (on SD1.5, SDXL and SSD-1B), a community Midjourney-style model, and the commercial DALL$\cdot$E 2 and 3, Firefly Image 2 and 3, Ideogram v1 and v2, Imagen 3 and Midjourney v5.2 and v6.1.
After pixel deduplication each generator has 266 to 709 test images in the first split.

\noindent\textbf{CF-Small.}
CF-Small is built from 30 generated-image shards and 2 real-image shards of the Community Forensics small release (Hugging Face dataset \nolinkurl{OwensLab/CommunityForensics-Small}), which hold 2,253 generator names, community generators from Hugging Face.
After merging names across files and deduplicating pixels, 1,951 generators have at least 32 images, and these form CF-Small.
All of them carry the latent diffusion architecture label of Community Forensics.

\noindent\textbf{Base families from model cards.}
We assign a base family to each CF-Small generator from its Hugging Face model card and configuration files, in this order of priority.
(1) A declared base model, also through a declared parent, names the family (41 generators).
(2) Otherwise, the Diffusers pipeline class names it: \texttt{StableDiffusionXLPipeline} gives SDXL, and \texttt{StableDiffusionPipeline} gives SD1/SD2 (279 generators).
(3) Otherwise, explicit base-model tokens in the repository name decide (175 generators).
Conflicting declarations would give no label, and none occur.
This gives 380 SD1/SD2 generators and 115 SDXL generators.
The pipeline class does not separate SD1 from SD2, so the two form one family here.
The other 1,456 generators have no base family and stay out of the lineage experiments, while the gallery-size study of \cref{sec:supp_gallery} uses all 1,951.

\subsection{ASVspoof protocol}
\label{sec:supp_asvspoof}

ASVspoof 2019 LA~\cite{wang2020asvspoof2019} holds 19 speech generators, A01 to A19.
The training part holds A01 to A06, and we fit on them (9,000 utterances per split).
ASVspoof 2021 LA~\cite{liu2023asvspoof2021} reuses the 2019 evaluation generators and sends them through six transmission conditions, five codecs (A-law, $\mu$-law, G.722, GSM, Opus) and a public switched telephone network (PSTN) line, plus the uncompressed channel.
In 2021, A16 and A19 are the same systems as A04 and A06, so they are the registered generators, and the 11 generators A07 to A15, A17 and A18 are unseen.
We sample up to 1,500 utterances per generator in each 2019 part and up to 300 per generator and channel in the 2021 evaluation part, 71,400 utterances in all, including 6,600 bona fide utterances, which are not from a generator and stay out of every metric.

\subsection{Family-Tree set}
\label{sec:supp_tree_set}

\begin{table*}[t]
\centering
\caption{\textbf{Family-Tree set.} Thirteen public models and five weight merges. A merge $A\times B$ ($w$) sets the denoising network and text-encoder weights to $(1-w)A+wB$ and takes the autoencoder, tokenizer and scheduler of $A$. DS-8 is DreamShaper-8, RV-5.1 is RealisticVision-5.1 and Anim-XL is Animagine-XL-3.1. Playground v2 shares the SDXL architecture and has its own weights. Heavy style fine-tunes are trained toward a distinct visual style (anime, photorealism, Midjourney look) according to their model cards. DreamShaper-XL is a light general-purpose fine-tune. Sampler: steps / guidance / resolution in pixels. Every model renders the same 400 COCO captions with one seed per caption, shared by all models.}
\label{tab:supp_tree_models}
\fontsize{8}{9.5}\selectfont
\setlength{\tabcolsep}{4pt}
\begin{tabular}{@{}lllc@{}}
\toprule
Model & Checkpoint or parents & Relation & Sampler \\
\midrule
\multicolumn{4}{@{}l}{\textit{SD1 family}} \\
SD1.4 & \texttt{CompVis/stable-diffusion-v1-4} & sibling of SD1.5 (earlier release) & 50 / 7.5 / 512 \\
SD1.5 & \texttt{stable-diffusion-v1-5/stable-diffusion-v1-5} & registered base & 50 / 7.5 / 512 \\
DreamShaper-8 & \texttt{Lykon/dreamshaper-8} & heavy style fine-tune & 50 / 7.5 / 512 \\
RealisticVision-5.1 & \texttt{SG161222/Realistic\_Vision\_V5.1\_noVAE} & heavy style fine-tune & 50 / 7.5 / 512 \\
Openjourney & \texttt{prompthero/openjourney} & heavy style fine-tune & 50 / 7.5 / 512 \\
Anything-v5 & \texttt{stablediffusionapi/anything-v5} & heavy style fine-tune & 50 / 7.5 / 512 \\
SD1.5$\times$DS-8 (0.25, 0.50, 0.75) & SD1.5 and DreamShaper-8 & weight merges & 50 / 7.5 / 512 \\
DS-8$\times$RV-5.1 (0.50) & DreamShaper-8 and RealisticVision-5.1 & weight merge & 50 / 7.5 / 512 \\
\midrule
\multicolumn{4}{@{}l}{\textit{SD2 family}} \\
SD2.1 & \texttt{sd2-community/stable-diffusion-2-1-base} & registered base & 50 / 7.5 / 512 \\
SD2.1-768 & \texttt{sd2-community/stable-diffusion-2-1} & \begin{tabular}[t]{@{}l@{}}768-pixel relative of SD2.1\\(via SD2.0-768-v)\end{tabular} & 50 / 7.5 / 768 \\
\midrule
\multicolumn{4}{@{}l}{\textit{SDXL family}} \\
SDXL & \texttt{stabilityai/stable-diffusion-xl-base-1.0} & registered base & 30 / 7.0 / 1,024 \\
SDXL-Turbo & \texttt{stabilityai/sdxl-turbo} & distillation & 4 / 0.0 / 512 \\
DreamShaper-XL & \texttt{Lykon/dreamshaper-xl-1-0} & light fine-tune & 30 / 7.0 / 1,024 \\
Animagine-XL-3.1 & \texttt{cagliostrolab/animagine-xl-3.1} & heavy style fine-tune & 30 / 7.0 / 1,024 \\
SDXL$\times$Anim-XL (0.50) & SDXL and Animagine-XL-3.1 & weight merge & 30 / 7.0 / 1,024 \\
\midrule
\multicolumn{4}{@{}l}{\textit{Own family}} \\
Playground v2 & \texttt{playgroundai/playground-v2-1024px-aesthetic} & registered base & 30 / 3.0 / 1,024 \\
\bottomrule
\end{tabular}
\end{table*}

The Family-Tree set (\cref{tab:supp_tree_models}) holds 13 public models and 5 weight merges with known relations.
Its captions are the first 400 captions of the COCO-dataset caption list.
The seed of each caption is derived from a SHA-256 hash of its index and COCO annotation id, and every model uses the same seed for the same caption.
Each model uses half-precision weights where the checkpoint provides them.

The known tree follows the documented lineage of the model cards.
It joins the four families at a virtual root, which stands for no model, with four children: SD1.2, SD2.0-base, SDXL and Playground v2.
SD1.4 and SD1.5 are both children of SD1.2, and the attribution test counts SD1.4 as SD1.5's earlier version.
SD2.1 is a child of SD2.0-base, and SD2.1-768 is a child of SD2.0-768-v, itself a child of SD2.0-base.
Animagine-XL-3.1 descends from SDXL through Animagine-XL 2.0 and 3.0.
We did not sample SD1.2, SD2.0-base, SD2.0-768-v, Animagine-XL 2.0 and 3.0, so these five checkpoints enter the tree as unlabeled nodes.
The four SD1 fine-tunes descend from SD1.5, and SDXL-Turbo and DreamShaper-XL from SDXL.
Each merge descends from both of its parents.
The tree distance of two models is the number of edges on the shortest path between them, including edges through unlabeled nodes.
Playground v2 forms a family of its own, which gives 56 model pairs within a family and 97 across families.
On this tree the fingerprint distance agrees with the tree distance with Spearman correlation 0.46 over the 153 model pairs (CF, $k=16$).

\section{Generator mean images}
\label{sec:fingerprints}

Even in raw pixels, the mean images of generators differ along a few compact, interpretable and portable directions.

\noindent\textbf{Two datasets.}
We build two datasets of generated images that differ in captions, real images and rendering setup.
In the COCO dataset 12 text-to-image generators render captions of COCO~\cite{lin2014coco} photographs, and in the Flickr dataset 12 render captions of Flickr30k~\cite{young2014flickr30k} photographs. Every generated image thus has a matched real image.
Ten generators appear in both datasets. Gemini 2.5 and 3.1 Flash Image~\cite{google2025geminiimage} appear only in the COCO dataset, and AuraFlow~\cite{fal2024auraflow} and SD3.5~\cite{esser2024sd3} only in the Flickr dataset (\cref{tab:supp_world_generators}).

\noindent\textbf{Generator mean images.}
Following \citet{marra2019gans}, who average the noise residuals of a generative adversarial network (GAN), we average the pixels of each generator over the captions that every generator rendered.
For generator $g$ and caption $i$, let $x_{g,i}$ be the rendered image ($256\times256$ RGB pixels). The mean image over $n$ captions is $\mu_g = \frac{1}{n}\sum_{i} x_{g,i}$, computed across 2,711 COCO and 1,200 Flickr30k captions.
Subtracting the average over generators, $\bar{\mu}$, cancels the content common to all of them, so $\mu_g - \bar{\mu}$ keeps how generator $g$ differs from the others.
A singular value decomposition of the centered mean images gives axes $j = 1, 2, \dots$ with singular values $s_1 \geq s_2 \geq \dots \geq 0$, and axis $j$ carries $s_j^2$ of the energy $\sum_j s_j^2$.

\noindent\textbf{Compact.}
We measure compactness with the participation ratio $\mathrm{PR} = (\sum_j s_j^2)^2 / \sum_j s_j^4$, the effective number of axes the differences occupy (1 for a single axis, $G-1$ for an even spread over $G$ generators).
In the COCO dataset the 12 generators give PR 1.51, and the first two axes carry 0.91 of the energy.
In the Flickr dataset the 10 shared generators give PR 1.33, with 0.94 of the energy on the first two axes.
In both datasets none of 10,000 draws from any of four null models reaches a PR this low (\cref{fig:fingerprints}c, \cref{sec:supp_nulls}).

\noindent\textbf{Interpretable.}
The first axis is global exposure, the overall brightness of an image. The generators line up along this axis from darker to brighter images (\cref{fig:fingerprints}a).
This axis matches the documented pull of Stable Diffusion toward medium brightness~\cite{lin2024common}.

\noindent\textbf{Portable.}
The 10 shared generators keep their first-axis order across the two datasets (Spearman correlation 0.93, \cref{fig:fingerprints}b).

\begin{figure*}[t]
  \centering
  \includegraphics[width=\textwidth]{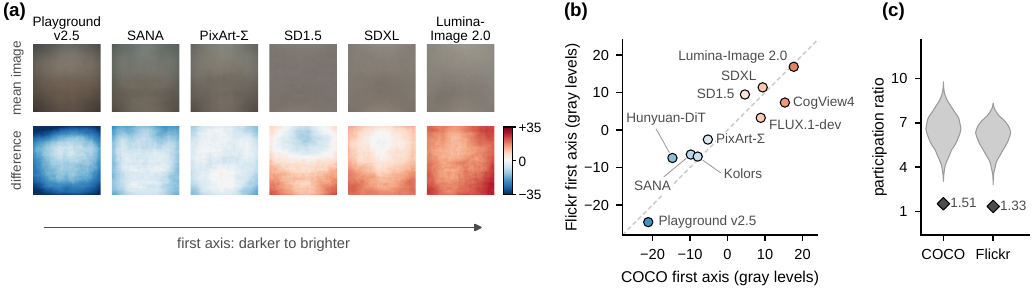}
  \caption{\textbf{Generator mean images in two datasets.}
  (a) COCO-dataset mean images of six generators along the first axis, and their brightness difference from the average generator (blue darker, red brighter).
  (b) First-axis coordinates of the 10 shared generators in the two datasets (equal coordinates dashed).
  (c) Participation ratio of the centered generator mean images (diamonds) against real images split at random into one group per generator (violins, 10,000 draws).}
  \label{fig:fingerprints}
\end{figure*}

\noindent\textbf{Null models for generator mean images.}
\label{sec:supp_nulls}
The four null models of \cref{sec:fingerprints} are random directions in pixel space, independent pixels with the observed per-pixel variances, the matched real images split at random into one group per generator, and equally separated generators.
The null for plane affinity draws random planes of smooth, low-frequency patterns with the same brightness component.

\noindent\textbf{First axis and plane alignment.}
\label{sec:supp_brightness}
The squared cosine of the first axis of \cref{sec:fingerprints} with a uniform image is 0.93 (COCO) and 0.95 (Flickr), and the average brightness of each color channel carries 0.79 and 0.85 of the energy.
Affinity is the mean squared cosine of the principal angles between two subspaces. The planes of the first two axes of the 10 shared generators in the two datasets have affinity 0.87, against a 95th percentile of 0.57 for random smooth planes with the same brightness component.

\noindent\textbf{Fingerprints without brightness.}
We subtract from each generator mean image of \cref{sec:fingerprints} its average in each color channel and repeat the analysis.
PR rises to 2.58 (COCO) and 2.61 (Flickr).
It stays below at least 9,999 of the 10,000 draws of every null model in both datasets.
The brightness-free planes align across datasets with affinity 0.87, against a null 95th percentile of at most 0.29.
The first-axis order also repeats, with Spearman correlation 0.90.
Generator differences are therefore more than a brightness shift.

\noindent\textbf{Generator means in feature space.}
\label{sec:supp_featspace}
We repeat the analysis on the frozen CF features (384 dimensions) of the clean images, with one mean feature per generator (507 to 510 images per generator in the COCO dataset, 1,200 in the Flickr dataset).
The 12 generator means of the COCO dataset have PR 2.30 of a possible 11, with 0.78 of the energy on the first two axes. Random splits of the same images into 12 groups give a mean PR of 7.24.
The 10 shared generators have PR 2.60 in the COCO dataset and 2.46 in the Flickr dataset.
The planes of their first two axes align across the two datasets with affinity 0.90, against a 95th percentile of 0.52 for planes from random splits of the Flickr images into 10 groups (1,000 draws).
The first-axis order of the 10 generators repeats with Spearman correlation 0.93, and the second-axis order with 0.90.
With features standardized per dimension the values are PR 2.56, affinity 0.90 and Spearman 0.90.

\section{Full results}
\label{sec:supp_results}
\begin{table*}[t]
\centering
\caption{\textbf{Every score on the Family-Tree captions and across datasets.} AUROC per unseen generator, averaged over the 25 unseen models on the Family-Tree captions (left, \cref{sec:supp_splits}) and the 12 unseen generators of the cross-dataset test (right, the right columns of \cref{tab:unseen}), one run. Second line: 95\% bootstrap interval of LineagePrint (LDA) minus that row, over families and then generators (left) and over generators (right). Per-image scores are averaged over the set, and set-mean scores score the mean feature of the set. Centered: pool mean minus fitting mean subtracted. Nearest centroid: standardized raw features.}
\label{tab:supp_unseen_scores}
\footnotesize
\setlength{\tabcolsep}{4pt}
\begin{tabular}{@{}l ccc@{\hspace{10pt}}ccc@{}}
\toprule
& \multicolumn{3}{c}{25 unseen models} & \multicolumn{3}{c}{12 unseen generators} \\
\cmidrule(lr){2-4}\cmidrule(l){5-7}
Method & Clean & New & All & Clean & New & All \\
\midrule
LineagePrint (LDA) & 0.82 & 0.86 & 0.86 & 0.87 & 0.88 & 0.88 \\
\midrule
\multicolumn{7}{@{}l}{\textit{Per-image scores averaged over the set}} \\
MSP & 0.72 & 0.59 & 0.63 & 0.84 & 0.68 & 0.72 \\
 & {\scriptsize [$-$0.02, +0.23]} & {\scriptsize [+0.15, +0.38]} & {\scriptsize [+0.13, +0.34]} & {\scriptsize [$-$0.04, +0.10]} & {\scriptsize [+0.15, +0.26]} & {\scriptsize [+0.11, +0.23]} \\
Energy & 0.74 & 0.61 & 0.65 & 0.91 & 0.73 & 0.77 \\
 & {\scriptsize [$-$0.06, +0.24]} & {\scriptsize [+0.12, +0.39]} & {\scriptsize [+0.10, +0.35]} & {\scriptsize [$-$0.10, +0.03]} & {\scriptsize [+0.10, +0.20]} & {\scriptsize [+0.06, +0.16]} \\
Mahalanobis & 0.56 & 0.62 & 0.62 & 0.62 & 0.60 & 0.61 \\
 & {\scriptsize [+0.10, +0.50]} & {\scriptsize [+0.14, +0.39]} & {\scriptsize [+0.14, +0.40]} & {\scriptsize [+0.13, +0.36]} & {\scriptsize [+0.19, +0.37]} & {\scriptsize [+0.19, +0.37]} \\
kNN & 0.56 & 0.54 & 0.55 & 0.58 & 0.58 & 0.58 \\
 & {\scriptsize [+0.07, +0.52]} & {\scriptsize [+0.21, +0.49]} & {\scriptsize [+0.22, +0.46]} & {\scriptsize [+0.14, +0.44]} & {\scriptsize [+0.20, +0.42]} & {\scriptsize [+0.20, +0.40]} \\
\midrule
\multicolumn{7}{@{}l}{\textit{Per-image scores averaged over the set, centered}} \\
MSP & 0.67 & 0.63 & 0.64 & 0.81 & 0.71 & 0.72 \\
 & {\scriptsize [+0.02, +0.29]} & {\scriptsize [+0.12, +0.33]} & {\scriptsize [+0.11, +0.33]} & {\scriptsize [$-$0.03, +0.15]} & {\scriptsize [+0.10, +0.25]} & {\scriptsize [+0.09, +0.24]} \\
Energy & 0.64 & 0.64 & 0.64 & 0.82 & 0.76 & 0.77 \\
 & {\scriptsize [+0.04, +0.34]} & {\scriptsize [+0.09, +0.37]} & {\scriptsize [+0.09, +0.37]} & {\scriptsize [$-$0.06, +0.15]} & {\scriptsize [+0.04, +0.20]} & {\scriptsize [+0.04, +0.19]} \\
Mahalanobis & 0.57 & 0.62 & 0.62 & 0.63 & 0.62 & 0.62 \\
 & {\scriptsize [+0.10, +0.50]} & {\scriptsize [+0.13, +0.40]} & {\scriptsize [+0.14, +0.40]} & {\scriptsize [+0.12, +0.34]} & {\scriptsize [+0.18, +0.35]} & {\scriptsize [+0.18, +0.36]} \\
kNN & 0.51 & 0.51 & 0.52 & 0.55 & 0.55 & 0.55 \\
 & {\scriptsize [+0.18, +0.51]} & {\scriptsize [+0.25, +0.49]} & {\scriptsize [+0.25, +0.48]} & {\scriptsize [+0.19, +0.48]} & {\scriptsize [+0.24, +0.45]} & {\scriptsize [+0.23, +0.45]} \\
\midrule
\multicolumn{7}{@{}l}{\textit{Set-mean scores}} \\
MSP & 0.81 & 0.65 & 0.71 & 0.96 & 0.79 & 0.85 \\
 & {\scriptsize [$-$0.11, +0.15]} & {\scriptsize [+0.09, +0.32]} & {\scriptsize [+0.05, +0.27]} & {\scriptsize [$-$0.14, $-$0.04]} & {\scriptsize [+0.03, +0.14]} & {\scriptsize [$-$0.02, +0.08]} \\
Energy & 0.81 & 0.67 & 0.72 & 0.96 & 0.82 & 0.86 \\
 & {\scriptsize [$-$0.12, +0.17]} & {\scriptsize [+0.08, +0.31]} & {\scriptsize [+0.04, +0.27]} & {\scriptsize [$-$0.14, $-$0.04]} & {\scriptsize [+0.02, +0.11]} & {\scriptsize [$-$0.02, +0.06]} \\
Mahalanobis & 0.74 & 0.67 & 0.71 & 0.75 & 0.62 & 0.65 \\
 & {\scriptsize [+0.02, +0.17]} & {\scriptsize [+0.14, +0.24]} & {\scriptsize [+0.11, +0.19]} & {\scriptsize [+0.04, +0.20]} & {\scriptsize [+0.21, +0.32]} & {\scriptsize [+0.16, +0.29]} \\
kNN & 0.56 & 0.52 & 0.53 & 0.59 & 0.57 & 0.58 \\
 & {\scriptsize [+0.06, +0.54]} & {\scriptsize [+0.19, +0.53]} & {\scriptsize [+0.21, +0.50]} & {\scriptsize [+0.14, +0.42]} & {\scriptsize [+0.20, +0.42]} & {\scriptsize [+0.21, +0.40]} \\
Nearest centroid & 0.59 & 0.64 & 0.65 & 0.50 & 0.58 & 0.57 \\
 & {\scriptsize [$-$0.03, +0.43]} & {\scriptsize [+0.14, +0.29]} & {\scriptsize [+0.13, +0.30]} & {\scriptsize [+0.22, +0.50]} & {\scriptsize [+0.22, +0.39]} & {\scriptsize [+0.22, +0.39]} \\
\midrule
\multicolumn{7}{@{}l}{\textit{Set-mean scores, centered}} \\
MSP & 0.77 & 0.76 & 0.77 & 0.94 & 0.89 & 0.90 \\
 & {\scriptsize [$-$0.07, +0.21]} & {\scriptsize [$-$0.01, +0.21]} & {\scriptsize [$-$0.01, +0.21]} & {\scriptsize [$-$0.13, $-$0.01]} & {\scriptsize [$-$0.08, +0.05]} & {\scriptsize [$-$0.08, +0.05]} \\
Energy & 0.73 & 0.74 & 0.74 & 0.93 & 0.89 & 0.90 \\
 & {\scriptsize [$-$0.05, +0.27]} & {\scriptsize [+0.01, +0.25]} & {\scriptsize [+0.01, +0.26]} & {\scriptsize [$-$0.13, 0.00]} & {\scriptsize [$-$0.07, +0.04]} & {\scriptsize [$-$0.08, +0.04]} \\
Mahalanobis & 0.76 & 0.74 & 0.76 & 0.75 & 0.67 & 0.69 \\
 & {\scriptsize [+0.01, +0.14]} & {\scriptsize [+0.05, +0.17]} & {\scriptsize [+0.06, +0.15]} & {\scriptsize [+0.04, +0.21]} & {\scriptsize [+0.13, +0.30]} & {\scriptsize [+0.11, +0.28]} \\
kNN & 0.47 & 0.48 & 0.48 & 0.53 & 0.52 & 0.53 \\
 & {\scriptsize [+0.20, +0.55]} & {\scriptsize [+0.26, +0.55]} & {\scriptsize [+0.27, +0.54]} & {\scriptsize [+0.20, +0.48]} & {\scriptsize [+0.23, +0.49]} & {\scriptsize [+0.23, +0.49]} \\
Nearest centroid & 0.78 & 0.73 & 0.75 & 0.68 & 0.66 & 0.66 \\
 & {\scriptsize [$-$0.03, +0.12]} & {\scriptsize [+0.07, +0.18]} & {\scriptsize [+0.07, +0.17]} & {\scriptsize [+0.07, +0.30]} & {\scriptsize [+0.13, +0.31]} & {\scriptsize [+0.12, +0.31]} \\
\midrule
\multicolumn{7}{@{}l}{\textit{LineagePrint variants}} \\
LineagePrint (LDA), uncentered & 0.81 & 0.69 & 0.76 & 0.82 & 0.71 & 0.78 \\
 & {\scriptsize [$-$0.02, +0.03]} & {\scriptsize [+0.13, +0.20]} & {\scriptsize [+0.08, +0.12]} & {\scriptsize [0.00, +0.08]} & {\scriptsize [+0.13, +0.20]} & {\scriptsize [+0.07, +0.14]} \\
LineagePrint (PLDA), uncentered & 0.86 & 0.72 & 0.79 & 0.96 & 0.81 & 0.87 \\
 & {\scriptsize [$-$0.11, +0.04]} & {\scriptsize [+0.07, +0.20]} & {\scriptsize [+0.02, +0.14]} & {\scriptsize [$-$0.13, $-$0.06]} & {\scriptsize [+0.03, +0.11]} & {\scriptsize [$-$0.03, +0.05]} \\
LineagePrint (PLDA) & 0.87 & 0.85 & 0.86 & 0.97 & 0.93 & 0.94 \\
 & {\scriptsize [$-$0.13, +0.06]} & {\scriptsize [$-$0.05, +0.09]} & {\scriptsize [$-$0.06, +0.08]} & {\scriptsize [$-$0.14, $-$0.06]} & {\scriptsize [$-$0.08, $-$0.03]} & {\scriptsize [$-$0.09, $-$0.03]} \\
\bottomrule
\end{tabular}
\end{table*}
\begin{table*}[t]
\centering
\caption{\textbf{Discovery per unseen model.} The 25-model test on the Family-Tree captions: fit on COCO-dataset captions outside the 400 Family-Tree captions (12 registered generators), test on those 400 captions, CF features, 42 disjoint sets of $k=8$ per model against 42 per registered generator, one run, chance 0.50. AUROC of each model's sets against the registered sets on clean images, on the six new pipelines (New) and on all ten (All), for LineagePrint (LDA) and for Energy on the centered set mean. Family: the family used by the bootstrap over families. Relative: the registered generator the model derives from or is a sibling of. Shared VAE: autoencoder weights identical to those of a registered generator by tensor hash (yes), documented as identical (yes$^{d}$), different (no) or not checked (?). The last two blocks are not counted in the mean: the checkpoints of registered generators with other sampler settings, expected near 0.50, and the weight merges of the Family-Tree set. SD2.1 is the 512-pixel base and SD2.1-768 its 768-pixel relative. DS-8: DreamShaper-8, RV-5.1: RealisticVision-5.1, Anim-XL: Animagine-XL-3.1, $w$: merge weight on the second parent. FLUX.1-dev$\times$Krea-dev is our weight merge and is one of the 25 models.}
\label{tab:supp_unseen_models}
\small
\setlength{\tabcolsep}{3.5pt}
\begin{tabular}{@{}llll ccc@{\hspace{8pt}}ccc@{}}
\toprule
& & & & \multicolumn{3}{c}{LineagePrint (LDA)} & \multicolumn{3}{c}{Energy, centered} \\
\cmidrule(lr){5-7}\cmidrule(l){8-10}
Model & Family & Relative & Shared VAE & Clean & New & All & Clean & New & All \\
\midrule
\multicolumn{10}{@{}l}{\textit{Unseen, derived from or a sibling of a registered generator}} \\
Anything-v5 & SD1 & SD1.5 & no & 0.94 & 0.95 & 0.96 & 0.89 & 0.73 & 0.75 \\
DreamShaper-8 & SD1 & SD1.5 & no & 0.97 & 0.98 & 0.98 & 1.00 & 0.99 & 0.99 \\
Openjourney & SD1 & SD1.5 & no & 0.38 & 0.60 & 0.59 & 0.25 & 0.33 & 0.31 \\
RealisticVision-5.1 & SD1 & SD1.5 & no & 0.82 & 0.98 & 0.96 & 0.74 & 0.96 & 0.92 \\
SD1.4 & SD1 & SD1.5 & yes & 0.79 & 0.58 & 0.63 & 0.05 & 0.08 & 0.08 \\
Animagine-XL-3.1 & SDXL & SDXL & yes$^{d}$ & 0.96 & 0.98 & 0.98 & 0.94 & 0.85 & 0.89 \\
DreamShaper-XL & SDXL & SDXL & yes & 1.00 & 0.98 & 0.99 & 0.23 & 0.40 & 0.39 \\
SDXL-Turbo & SDXL & SDXL & no & 1.00 & 0.99 & 0.99 & 0.39 & 0.67 & 0.64 \\
Playground v2 & Playground & Playground v2.5 & yes & 0.92 & 0.93 & 0.95 & 0.98 & 0.92 & 0.94 \\
AWPortrait-FL & FLUX & FLUX.1-dev & yes & 0.98 & 0.93 & 0.95 & 0.86 & 0.89 & 0.88 \\
FLUX.1-Krea-dev & FLUX & FLUX.1-dev & yes & 0.98 & 0.97 & 0.98 & 0.91 & 0.94 & 0.94 \\
FLUX.1-dev + Turbo-Alpha & FLUX & FLUX.1-dev & yes & 0.67 & 0.61 & 0.64 & 0.69 & 0.68 & 0.68 \\
FLUX.1-dev$\times$Krea-dev ($w$=0.50) & FLUX & FLUX.1-dev & yes & 0.63 & 0.70 & 0.69 & 0.82 & 0.84 & 0.83 \\
FLUX.1-lite-8B & FLUX & FLUX.1-dev & yes & 0.52 & 0.47 & 0.46 & 0.56 & 0.56 & 0.55 \\
FLUX.1-schnell & FLUX & FLUX.1-dev & yes & 0.66 & 0.80 & 0.77 & 0.91 & 0.89 & 0.90 \\
Flex.1-alpha & FLUX & FLUX.1-dev & yes & 0.99 & 0.97 & 0.98 & 0.97 & 0.88 & 0.90 \\
SANA 1.5 & SANA & SANA & ? & 0.43 & 0.75 & 0.72 & 0.89 & 0.86 & 0.87 \\
\midrule
\multicolumn{10}{@{}l}{\textit{Unseen, no registered relative}} \\
SD2.1 & SD2 & -- & no & 0.36 & 0.57 & 0.58 & 0.19 & 0.28 & 0.27 \\
SD2.1-768 & SD2 & -- & no & 0.75 & 0.87 & 0.88 & 0.49 & 0.53 & 0.53 \\
SD3.5-large & SD3.5 & -- & no & 0.98 & 0.98 & 0.98 & 0.98 & 0.98 & 0.98 \\
SD3.5-large-TurboX & SD3.5 & -- & no & 0.96 & 0.99 & 0.99 & 0.93 & 0.97 & 0.97 \\
SD3.5-large-turbo & SD3.5 & -- & no & 0.95 & 0.95 & 0.96 & 0.98 & 0.95 & 0.96 \\
SD3.5-medium & SD3.5 & -- & no & 0.98 & 0.98 & 0.98 & 1.00 & 0.99 & 0.99 \\
SD3.5-medium-turbo & SD3.5 & -- & no & 0.99 & 0.96 & 0.97 & 0.75 & 0.46 & 0.49 \\
Z-Image-Turbo & Z-Image & -- & yes & 0.97 & 0.96 & 0.97 & 0.89 & 0.90 & 0.90 \\
\midrule
\multicolumn{10}{@{}l}{\textit{Not counted: same checkpoint as a registered generator, other sampler settings}} \\
SD1.5 (Family-Tree settings) & SD1 & SD1.5 & yes & 0.74 & 0.57 & 0.63 & 0.05 & 0.09 & 0.09 \\
SDXL (Family-Tree settings) & SDXL & SDXL & yes & 0.64 & 0.55 & 0.57 & 0.39 & 0.45 & 0.44 \\
FLUX.1-dev (28 steps) & FLUX & FLUX.1-dev & yes & 0.53 & 0.43 & 0.43 & 0.60 & 0.62 & 0.61 \\
\midrule
\multicolumn{10}{@{}l}{\textit{Not counted: weight merges of the Family-Tree set}} \\
DS-8$\times$RV-5.1 ($w$=0.50) & SD1 & SD1.5 & no & 0.95 & 0.99 & 0.99 & 0.98 & 1.00 & 0.99 \\
SD1.5$\times$DS-8 ($w$=0.25) & SD1 & SD1.5 & yes & 0.61 & 0.53 & 0.58 & 0.09 & 0.20 & 0.20 \\
SD1.5$\times$DS-8 ($w$=0.50) & SD1 & SD1.5 & yes & 0.48 & 0.74 & 0.72 & 0.19 & 0.47 & 0.45 \\
SD1.5$\times$DS-8 ($w$=0.75) & SD1 & SD1.5 & yes & 0.59 & 0.92 & 0.87 & 0.42 & 0.76 & 0.72 \\
SDXL$\times$Anim-XL ($w$=0.50) & SDXL & SDXL & yes & 0.48 & 0.58 & 0.59 & 0.50 & 0.48 & 0.48 \\
\midrule
\multicolumn{4}{@{}l}{Mean over the 25 counted models} & 0.82 & 0.86 & 0.86 & 0.73 & 0.74 & 0.74 \\
\bottomrule
\end{tabular}
\end{table*}
\begin{table*}[t]
\centering
\caption{\textbf{Discovery per unseen generator across datasets.} The 12-generator test of \cref{tab:unseen}: fit in the COCO dataset (up to 600 images per generator), test in the Flickr dataset, CF features, 120 disjoint sets of $k=8$ per unseen generator against 120 per registered generator, one run, chance 0.50. Held out: one of the ten generators present in both datasets, removed from the fit and from the registered generators for its own test (nine registered). Flickr only: AuraFlow and SD3.5-large, never in the fit (ten registered). Columns as in \cref{tab:unseen}.}
\label{tab:supp_unseen_generators}
\small
\setlength{\tabcolsep}{3.5pt}
\begin{tabular}{@{}l ccc@{\hspace{8pt}}ccc@{\hspace{8pt}}ccc@{}}
\toprule
& \multicolumn{3}{c}{LineagePrint (LDA)} & \multicolumn{3}{c}{LineagePrint (PLDA)} & \multicolumn{3}{c}{Energy, centered} \\
\cmidrule(lr){2-4}\cmidrule(lr){5-7}\cmidrule(l){8-10}
Generator & Clean & New & All & Clean & New & All & Clean & New & All \\
\midrule
\multicolumn{10}{@{}l}{\textit{Held out of the fit in turn}} \\
CogView4 & 0.83 & 0.83 & 0.84 & 0.96 & 0.91 & 0.93 & 0.78 & 0.77 & 0.79 \\
FLUX.1-dev & 0.97 & 0.97 & 0.97 & 1.00 & 0.98 & 0.99 & 0.89 & 0.92 & 0.92 \\
Hunyuan-DiT & 0.81 & 0.91 & 0.89 & 0.98 & 0.95 & 0.96 & 0.98 & 0.86 & 0.88 \\
Kolors & 0.75 & 0.86 & 0.84 & 0.92 & 0.90 & 0.91 & 0.97 & 0.93 & 0.94 \\
Lumina-Image 2.0 & 0.89 & 0.90 & 0.89 & 0.98 & 0.95 & 0.95 & 0.94 & 0.95 & 0.96 \\
PixArt-$\Sigma$ & 0.76 & 0.66 & 0.68 & 0.95 & 0.82 & 0.85 & 0.97 & 0.88 & 0.88 \\
Playground v2.5 & 0.98 & 0.93 & 0.94 & 1.00 & 0.96 & 0.97 & 0.99 & 0.95 & 0.97 \\
SANA & 0.81 & 0.82 & 0.83 & 0.96 & 0.90 & 0.92 & 0.96 & 0.91 & 0.92 \\
SD1.5 & 0.99 & 0.96 & 0.97 & 1.00 & 0.98 & 0.99 & 0.97 & 0.90 & 0.90 \\
SDXL & 0.98 & 0.97 & 0.97 & 1.00 & 0.97 & 0.97 & 0.94 & 0.81 & 0.80 \\
\midrule
\multicolumn{10}{@{}l}{\textit{Flickr only}} \\
AuraFlow & 0.77 & 0.82 & 0.83 & 0.96 & 0.93 & 0.94 & 0.99 & 0.97 & 0.97 \\
SD3.5-large & 0.87 & 0.93 & 0.93 & 0.92 & 0.93 & 0.93 & 0.80 & 0.88 & 0.87 \\
\midrule
Mean over the 12 & 0.87 & 0.88 & 0.88 & 0.97 & 0.93 & 0.94 & 0.93 & 0.89 & 0.90 \\
\bottomrule
\end{tabular}
\end{table*}
 
\noindent\textbf{The two larger discovery tests.}
\Cref{tab:supp_unseen_scores} gives every score of the 25-model test on the Family-Tree captions and of the 12-generator test with the interval of its difference to LineagePrint (LDA).
On the 25 unseen models LineagePrint (LDA) leads Energy on the centered set mean over all ten pipelines and on the six new ones, and is ahead for 17 of the 25 models over all ten.
On clean images the two are level (0.82 against 0.73), and LineagePrint (LDA) is level with MSP on the centered set mean in every column.
\Cref{tab:supp_unseen_models} lists every model.
Energy falls below 0.50 on five models, three with a registered relative (SD1.4 0.08, Openjourney 0.31, DreamShaper-XL 0.39) and two without (SD2.1 0.27, SD3.5-medium-turbo 0.49).
LineagePrint (LDA) is below 0.70 on six models: FLUX.1-lite-8B (0.46), SD2.1 (0.58), Openjourney (0.59), SD1.4 (0.63), FLUX.1-dev with the fused Turbo-Alpha LoRA (0.64) and the FLUX.1 merge (0.69).
Three checkpoints of registered generators sampled with other settings score 0.43 to 0.63 where 0.50 is expected, so a per-model AUROC this close to 0.50 does not show a new generator.
On the 12 unseen generators of the cross-dataset test (\cref{tab:supp_unseen_generators}) LineagePrint (LDA) is level with Energy on the centered set mean on the six new pipelines and over all ten.
On clean images it reaches 0.87, below Energy without centering (0.96), MSP on the centered set mean (0.94) and LineagePrint (PLDA) (0.97).
LineagePrint (PLDA) has the highest mean in every column of this test, and on the 25 models the two variants are level (0.86 both).
With CLIP ViT-B/32 features LineagePrint (LDA) is level with Energy on the centered set mean in both tests (0.82 against 0.80 on the 25 models over all ten pipelines and 0.81 against 0.81 on the 12 generators on the six new ones).
Both tests come from one run.

\begin{table}[t]
\centering
\caption{\textbf{Discovery across datasets and processing.} Fit on 12 COCO-dataset generators (augmented fit). Test in the Flickr dataset: sets of the unseen AuraFlow and SD3.5 against sets of 10 registered generators, $k=8$ images, 200 sets per generator, 3 splits, chance 0.50. New / All: mean over the six pipelines never used in fitting / all ten. Per-image scores are averaged over the set. Set-mean baseline rows use the same set averaging and pool as LineagePrint. Pool: a fifth of the Flickr-dataset outputs of every test generator. Centered: pool mean minus fitting mean subtracted. Nearest centroid: standardized raw features. Pixel fingerprints: the noise residual averaged over the set and correlated with each registered generator's mean residual~\cite{marra2019gans}, and DIF~\cite{sinitsa2024deep} trained for each registered generator with the other registered generators as negatives. Best in bold, second best underlined. Over 20 split seeds on New (CF), LineagePrint (LDA) scores 0.91 $\pm$ 0.02 against 0.82 $\pm$ 0.06 for Energy, centered, and leads in 20 of 20.}
\label{tab:discovery}
\fontsize{8}{9.5}\selectfont
\setlength{\tabcolsep}{2.2pt}
\begin{tabular}{@{}lcccccc@{}}
\toprule
& \multicolumn{6}{c}{Discovery AUROC $\uparrow$} \\
\cmidrule(l){2-7}
& \multicolumn{3}{c}{CF ViT-S/16} & \multicolumn{3}{c}{CLIP ViT-B/32} \\
\cmidrule(lr){2-4}\cmidrule(l){5-7}
Method & Clean & New & All & Clean & New & All \\
\midrule
\multicolumn{7}{@{}l}{\textit{Standard per-image scores}} \\
MSP~\cite{hendrycks2017msp} & 0.61 & 0.59 & 0.62 & 0.53 & 0.56 & 0.55 \\
Energy~\cite{liu2020energy} & 0.71 & 0.70 & 0.72 & 0.64 & 0.61 & 0.63 \\
Mahalanobis~\cite{lee2018mahalanobis} & 0.86 & 0.79 & 0.80 & 0.47 & 0.47 & 0.47 \\
kNN~\cite{sun2022knn} & 0.75 & 0.75 & 0.72 & 0.38 & 0.38 & 0.39 \\
\midrule
\multicolumn{7}{@{}l}{\textit{Set-mean baselines}} \\
Energy & 0.88 & 0.81 & 0.84 & \underline{0.72} & 0.66 & \underline{0.69} \\
Energy, centered & 0.81 & 0.84 & 0.84 & \textbf{0.78} & \textbf{0.76} & \textbf{0.76} \\
Mahalanobis, centered & 0.80 & 0.63 & 0.68 & 0.54 & 0.52 & 0.53 \\
Nearest centroid, centered & 0.73 & 0.63 & 0.65 & 0.56 & 0.55 & 0.56 \\
\midrule
\multicolumn{7}{@{}l}{\textit{Pixel fingerprints on the same 256-pixel images}} \\
Noise residual~\cite{marra2019gans} & 0.72 & 0.65 & 0.68 & -- & -- & -- \\
Noise residual, centered & 0.37 & 0.37 & 0.38 & -- & -- & -- \\
DIF~\cite{sinitsa2024deep} & 0.43 & 0.44 & 0.46 & -- & -- & -- \\
DIF, centered & 0.44 & 0.47 & 0.49 & -- & -- & -- \\
\midrule
LineagePrint (LDA) & \textbf{0.97} & \textbf{0.91} & \textbf{0.92} & 0.62 & 0.61 & 0.62 \\
LineagePrint (PLDA) & \underline{0.94} & \underline{0.89} & \underline{0.90} & 0.69 & \underline{0.68} & 0.68 \\
\bottomrule
\end{tabular}
\end{table}
 
\begin{figure}[t]
  \centering
  \includegraphics[width=\columnwidth]{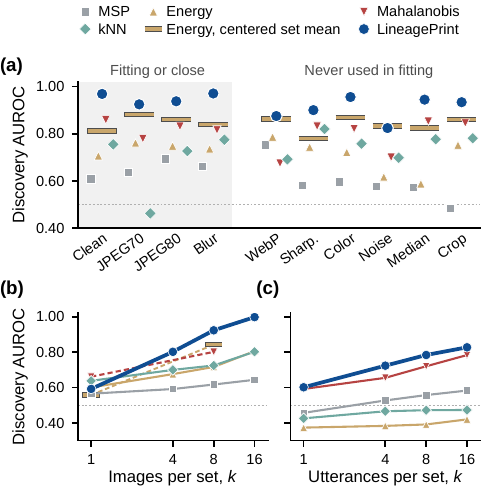}
  \caption{\textbf{Discovery across processing and set size with AuraFlow and SD3.5 unseen.}
  (a) AUROC per pipeline, CF features, augmented fit, $k=8$. Gray: pipelines used in fitting or close to them. Markers: per-image scores averaged over the set. Bars: Energy of the pool-centered set mean.
  (b) Images, AUROC vs set size $k$, mean over the ten pipelines, with per-image Mahalanobis and Energy of the centered set mean at $k=1$ and 8 (dashed). Single images give 0.56 to 0.66 for every method. From $k=4$ on LineagePrint leads MSP, Energy and kNN, and at $k=8$ it leads every score.
  (c) Speech, mean over the seven channels (\cref{sec:supp_speech}).
  LineagePrint is LDA for images and PLDA for speech.
  3 splits, chance 0.50 (dotted).}
  \label{fig:discovery}
\end{figure}

\noindent\textbf{The test with AuraFlow and SD3.5 unseen.}
\Cref{tab:discovery} is the test with two unseen generators that the false-alarm calibration (\cref{tab:calibration}), the pool study, the encoder sweep and the analysis of each part (\cref{tab:ladder}) use.
We fit on the 12 COCO-dataset generators and test in the Flickr dataset, where the 10 shared generators are registered and AuraFlow and SD3.5 are unseen.
On the six new pipelines LineagePrint (LDA) reaches a discovery AUROC of 0.91, against 0.84 for Energy on the centered set mean and 0.79 for standard per-image scores.
Over 20 split seeds it scores 0.91 $\pm$ 0.02 against 0.82 $\pm$ 0.06 and is ahead in 20 of 20 seeds. These seeds share test sets: with disjoint test sets and a pool of registered generators only, LineagePrint (LDA) reaches 0.91 against 0.80 over all ten pipelines, a difference whose 95\% interval includes zero.
It reaches 0.97 against 0.88 on clean images and 0.92 against 0.84 over all ten pipelines, and it is ahead of Energy on the centered set mean on nine of the ten (\cref{fig:discovery}a).
On CF-Eval (21 generators outside the CF detector's training set, 15 registered, $k=8$, 30 runs), LineagePrint (LDA) reaches AUROC 0.97 and top-1 0.99. With four registered generators it already reaches 0.97 and 1.00.

\begin{table}[t]
\centering
\caption{\textbf{False alarms at a chosen rate.} Realized false-alarm rate (FPR, mean over pipelines and splits), its worst pipeline (Worst) and detection rate (TPR) at a threshold set for $\alpha=0.05$ from same-pipeline anchor sets (Same pipeline) or from another source. The test with AuraFlow and SD3.5 unseen (\cref{tab:discovery}), CF features, sets of $k=8$, 50 anchor sets. Same-pipeline anchors keep the FPR close to $\alpha$.}
\label{tab:calibration}
\small
\setlength{\tabcolsep}{2.5pt}
\begin{tabular}{@{}lcccccc@{}}
\toprule
& \multicolumn{3}{c}{LineagePrint (LDA)} & \multicolumn{3}{c}{LineagePrint (PLDA)} \\
\cmidrule(lr){2-4}\cmidrule(l){5-7}
Threshold from & FPR & Worst & TPR $\uparrow$ & FPR & Worst & TPR $\uparrow$ \\
\midrule
\multicolumn{7}{@{}l}{\textit{Images, new dataset, 10 pipelines}} \\
Clean images & 0.08 & 0.19 & 0.55 & 0.08 & 0.15 & 0.50 \\
Same pipeline & 0.04 & 0.08 & 0.56 & 0.05 & 0.09 & 0.44 \\
\bottomrule
\end{tabular}
\end{table}
 
\noindent\textbf{False alarms.}
\Cref{tab:supp_fresh_calibration} gives the false-alarm rates of \cref{sec:exp_calibration}, and \cref{tab:calibration} those of the test across datasets.
In the test with AuraFlow and SD3.5 unseen, same-pipeline anchors keep the mean FPR at 0.04 (LDA) and 0.05 (PLDA), and thresholds from clean images give 0.08 with 0.19 and 0.15 on the worst pipeline.

\begin{table*}[t]
\centering
\caption{\textbf{Every method on native-resolution center crops.} 25 unseen models from 8 families against the 12 registered COCO-dataset generators on the Family-Tree captions (\cref{tab:supp_unseen_models}), disjoint sets of $k=8$, pool of registered generators only, one run. Discovery AUROC of a model: its sets against the sets of all 12 registered generators, chance 0.50. Input: standard 256 = the 256-pixel images of the main protocol with the augmented fit, crop 256 / 512 = center crop of 256 or 512 pixels at native resolution without resizing, with a fit on clean crops of the same fitting images. JPEG 85 is applied to the crop, and for the standard input to the full image before resizing. Centered: pool centering as in the main text. Set mean: features averaged over the set. Set residual: the set's mean noise residual correlated with each generator's mean residual. Image mean: each image scored and the scores averaged. Raw corr.: DIF without its normalized correlation. AuraFlow and SD3.5: the test with these two models unseen (\cref{tab:discovery}), mean over three splits, centered. DIF trained without failure, but for 5 of the 12 registered generators at 256 pixels and 7 at 512 pixels its fingerprint does not score its own training crops above the negatives.}
\label{tab:supp_native}
\fontsize{8}{9.5}\selectfont
\setlength{\tabcolsep}{3.5pt}
\begin{tabular}{@{}lllcccccc@{}}
\toprule
& & & \multicolumn{4}{c}{25 unseen models} & \multicolumn{2}{c}{AuraFlow and SD3.5} \\
\cmidrule(lr){4-7}\cmidrule(l){8-9}
& & & \multicolumn{2}{c}{Clean} & \multicolumn{2}{c}{JPEG 85} & Clean & JPEG 85 \\
\cmidrule(lr){4-5}\cmidrule(lr){6-7}\cmidrule(lr){8-8}\cmidrule(l){9-9}
Method & Input & Aggregation & Uncentered & Centered & Uncentered & Centered & Centered & Centered \\
\midrule
LineagePrint (LDA) & standard 256 & set mean & 0.81 & 0.82 & 0.82 & 0.83 & 0.95 & 0.93 \\
 & crop 256 & set mean & 0.96 & 0.96 & 0.38 & 0.71 & 0.85 & 0.77 \\
 & crop 512 & set mean & 0.94 & 0.94 & 0.58 & 0.83 & 0.85 & 0.85 \\
\midrule
Energy & standard 256 & set mean & 0.81 & 0.73 & 0.76 & 0.67 & 0.73 & 0.76 \\
 & crop 256 & set mean & 0.93 & 0.93 & 0.59 & 0.71 & 0.84 & 0.84 \\
 & crop 512 & set mean & 0.86 & 0.85 & 0.73 & 0.72 & 0.85 & 0.75 \\
\midrule
Noise residual~\cite{marra2019gans} & crop 256 & set residual & 0.77 & 0.75 & 0.71 & 0.70 & 0.82 & 0.64 \\
 & crop 256 & image mean & 0.76 & 0.74 & 0.70 & 0.69 & 0.80 & 0.61 \\
 & crop 512 & set residual & 0.79 & 0.79 & 0.70 & 0.71 & 0.81 & 0.69 \\
 & crop 512 & image mean & 0.79 & 0.78 & 0.70 & 0.71 & 0.80 & 0.70 \\
\midrule
DIF~\cite{sinitsa2024deep} & crop 256 & image mean & 0.76 & 0.77 & 0.43 & 0.45 & 0.78 & 0.44 \\
 & crop 256 & image mean, raw corr. & 0.72 & 0.72 & 0.44 & 0.38 & 0.51 & 0.40 \\
 & crop 256 & set residual & 0.71 & 0.72 & 0.45 & 0.45 & 0.75 & 0.46 \\
 & crop 256 & set residual, raw corr. & 0.73 & 0.73 & 0.46 & 0.39 & 0.50 & 0.38 \\
 & crop 512 & image mean & 0.75 & 0.75 & 0.41 & 0.44 & 0.85 & 0.43 \\
 & crop 512 & image mean, raw corr. & 0.58 & 0.58 & 0.44 & 0.47 & 0.70 & 0.26 \\
 & crop 512 & set residual & 0.75 & 0.75 & 0.42 & 0.44 & 0.81 & 0.43 \\
 & crop 512 & set residual, raw corr. & 0.58 & 0.58 & 0.44 & 0.49 & 0.67 & 0.29 \\
\bottomrule
\end{tabular}

\end{table*}
\begin{table*}[t]
\centering
\caption{\textbf{Forensic Self-Descriptions on native crops.} The 25 unseen models, sets and 512-pixel center crops of \cref{tab:supp_native}. FSD descriptors~\cite{nguyen2025forensic} come from the released extractor and transforms. FSD mixtures: the open-set rule of the method, one Gaussian mixture per registered generator with the released settings, fitted on the fit crops, scored per image and averaged over the set, with the pool mean subtracted, or on the set-mean descriptor. Difference: LineagePrint (LDA) with CF features minus the row. $^\dagger$The same descriptors with the generator subspace and nearest-fingerprint distance of LineagePrint (LDA).}
\label{tab:supp_fsd}
\footnotesize
\setlength{\tabcolsep}{6pt}
\begin{tabular}{@{}lccl@{}}
\toprule
Method & Clean & JPEG 85 & Difference, clean \\
\midrule
FSD mixtures, per image & 0.54 & 0.53 & +0.41 \\
FSD mixtures, per image, centered & 0.43 & 0.47 & +0.51 \\
FSD mixtures, set mean & 0.59 & 0.54 & +0.35 \\
FSD, LineagePrint back end$^\dagger$ & 0.82 & 0.63 & +0.13 \\
\midrule
LineagePrint (LDA), CF features & 0.94 & 0.83 & \\
\bottomrule
\end{tabular}
\end{table*}
 
\noindent\textbf{Pixel fingerprints on native-resolution crops.}
\label{sec:supp_native}
\Cref{tab:supp_native} gives every method the same native-resolution center crops, on the 25 unseen models of \cref{tab:supp_unseen_models} against the 12 registered generators, with disjoint sets of 8, a pool of registered generators only and one run.
Each image is cut to a 256- or 512-pixel center crop without resizing, JPEG at quality 85 is applied to the crop, and LineagePrint and Energy on crops are fit on clean crops of the same fitting images.
On the same 512-pixel crops LineagePrint (LDA) leads the noise residual and DIF on clean images (0.94 against 0.79 and 0.75) and under JPEG compression (0.83 against 0.71 and 0.44).
On the 8 models without a registered relative the two pixel fingerprints match LineagePrint (LDA) with the standard 256-pixel input on clean images (0.87 and 0.92 against 0.87).

\noindent\textbf{Forensic Self-Descriptions.}
\Cref{tab:supp_fsd} adds Forensic Self-Descriptions (FSD)~\cite{nguyen2025forensic}, a published method for open-set source attribution, on the same 512-pixel crops, sets and fit images.
We extract its descriptors with the released extractor and transforms and apply its open-set rule, one Gaussian mixture per registered generator with the released settings.
On clean crops LineagePrint (LDA) is 0.51 above the pool-centered per-image form.
The rule reaches at most 0.59 in its three forms, with about 510 fit crops per generator for mixtures in 960 dimensions.
The descriptors themselves separate the registered generators: with the generator subspace and the nearest-fingerprint distance of LineagePrint (LDA), they attribute registered sets with top-1 0.95 and discover the unseen models with AUROC 0.82 on clean crops, above the noise residual and DIF, and 0.63 under JPEG compression.
LineagePrint (LDA) with CF features is 0.13 above this form on clean crops.
On the same 256-pixel crops it reaches 0.96 on clean images against 0.75 and 0.77, and 0.71 under JPEG compression against 0.70 for the noise residual and 0.45 for DIF.
With the standard 256-pixel input, LineagePrint (LDA) reaches 0.82 against 0.79 and 0.77 for the two pixel fingerprints on native crops of clean images, and 0.83 against 0.71 and 0.45 under JPEG compression.
With AuraFlow and SD3.5 unseen, native crops lower LineagePrint (LDA) from 0.95 to 0.85 on clean images.

\begin{table}[t]
\centering
\caption{\textbf{Attribution to the base model.} Family-Tree set (13 public models and 5 weight merges, 400 COCO captions), CF features with the subspace learned on CF-Small after removing the 64 generator names that contain a Family-Tree model name, sets of $k=16$ images, 72 sets per model. The gallery holds four base models (SD1.5, SD2.1, SDXL, Playground v2). Base: the registered base model of its family. To base: sets out of 72 whose nearest gallery fingerprint is this base. Nearest model: the closest of the other 17 models by fingerprint distance. A merge $A \times B$ ($w$) has weights $(1-w)A + wB$; DS-8 is DreamShaper-8 and Anim-XL is Animagine-XL-3.1. Every model is attributed to its base by majority.}
\label{tab:tree}
\small
\setlength{\tabcolsep}{3pt}
\begin{tabular}{@{}llcl@{}}
\toprule
Model & Base & To base & Nearest model \\
\midrule
\multicolumn{4}{@{}l}{\textit{Sibling version}} \\
SD1.4 & SD1.5 & 72 & SD1.5 \\
\addlinespace[2pt]
\multicolumn{4}{@{}l}{\textit{768-pixel relative}} \\
SD2.1-768 & SD2.1 & 72 & SD2.1 \\
\addlinespace[2pt]
\multicolumn{4}{@{}l}{\textit{Distillation}} \\
SDXL-Turbo & SDXL & 72 & SDXL \\
\addlinespace[2pt]
\multicolumn{4}{@{}l}{\textit{Light fine-tune}} \\
DreamShaper-XL & SDXL & 72 & SDXL \\
\addlinespace[2pt]
\multicolumn{4}{@{}l}{\textit{Weight merge}} \\
SD1.5$\times$DS-8 (0.25) & SD1.5 & 72 & SD1.5$\times$DS-8 (0.50) \\
SD1.5$\times$DS-8 (0.50) & SD1.5 & 70 & SD1.5$\times$DS-8 (0.25) \\
SD1.5$\times$DS-8 (0.75) & SD1.5 & 50 & SD1.5$\times$DS-8 (0.50) \\
SDXL$\times$Anim-XL (0.50) & SDXL & 71 & SDXL \\
\midrule
\multicolumn{2}{@{}l}{All eight models} & 551/576 & \\
\bottomrule
\end{tabular}
\end{table}
\begin{table*}[t]
\centering
\caption{\textbf{Lineage at scale, all methods.} The 495 CF-Small community fine-tunes with a known base model form two families (SD1/SD2: 380, SDXL: 115). CF features, mean over 10 draws, chance 0.50. Top-1: family attribution of unseen fine-tunes with 32 generators of each family registered (LineagePrint uses the subspace fit with family labels, or with generator labels where stated). Discovery AUROC with 32 generators of one family registered (the gallery): other family = sets of 83 unseen fine-tunes of the unregistered family, same family = sets of 83 unseen fine-tunes of the registered family, 20 sets per generator. Every method uses the same fitting images, 8 per registered generator. The generator-label classifier gives its family prediction by summing its generator posteriors. The family-label classifier has one class when one family is registered, so it gives no discovery score. kNN on the set mean compares it with means of 4-image gallery sets. Nearest centroid: standardized raw features. PCA space and centroid span have the dimension of the LDA subspace. Family direction: one direction between the two family means. -: not defined. Best in bold, second best underlined.}
\label{tab:lineage}
\small
\setlength{\tabcolsep}{3pt}
\begin{tabular}{@{}lccccc@{\hspace{8pt}}ccccc@{}}
\toprule
& \multicolumn{5}{c}{Sets of $k=8$ images} & \multicolumn{5}{c}{Sets of $k=16$ images} \\
\cmidrule(lr){2-6}\cmidrule(l){7-11}
& Family & \multicolumn{2}{c}{Other family} & \multicolumn{2}{c}{Same family} & Family & \multicolumn{2}{c}{Other family} & \multicolumn{2}{c}{Same family} \\
\cmidrule(lr){3-4}\cmidrule(lr){5-6}\cmidrule(lr){8-9}\cmidrule(l){10-11}
Method & top-1 & SD1/SD2 & SDXL & SD1/SD2 & SDXL & top-1 & SD1/SD2 & SDXL & SD1/SD2 & SDXL \\
\midrule
\multicolumn{11}{@{}l}{\textit{Confidence of a linear classifier on generator labels}} \\
MSP, per image & \textbf{0.88} & 0.43 & 0.62 & 0.51 & 0.52 & \textbf{0.90} & 0.40 & 0.65 & 0.51 & 0.53 \\
MSP, set mean & 0.86 & 0.40 & 0.63 & 0.51 & 0.56 & \underline{0.88} & 0.38 & 0.64 & 0.51 & 0.56 \\
Energy, per image & \textbf{0.88} & 0.35 & 0.70 & 0.49 & 0.53 & \textbf{0.90} & 0.32 & 0.73 & 0.50 & 0.53 \\
Energy, set mean & 0.86 & 0.33 & 0.64 & 0.48 & 0.54 & \underline{0.88} & 0.32 & 0.64 & 0.48 & 0.54 \\
\midrule
\multicolumn{11}{@{}l}{\textit{Linear classifier on family labels (attribution only)}} \\
Family classifier, per image & 0.86 & -- & -- & -- & -- & \underline{0.88} & -- & -- & -- & -- \\
Family classifier, set mean & 0.86 & -- & -- & -- & -- & \underline{0.88} & -- & -- & -- & -- \\
\midrule
\multicolumn{11}{@{}l}{\textit{Distance to the gallery}} \\
Mahalanobis, per image & 0.83 & 0.73 & 0.52 & 0.53 & 0.52 & 0.85 & 0.76 & 0.53 & 0.54 & 0.52 \\
Mahalanobis, set mean & 0.78 & 0.88 & 0.69 & 0.60 & 0.58 & 0.79 & 0.93 & 0.78 & 0.63 & 0.62 \\
Mahalanobis, set mean, family mean & 0.86 & 0.87 & 0.64 & 0.57 & 0.56 & \underline{0.88} & 0.92 & 0.72 & 0.58 & 0.58 \\
kNN, per image & -- & 0.62 & 0.59 & 0.50 & 0.54 & -- & 0.65 & 0.60 & 0.50 & 0.55 \\
kNN, set mean, 1 neighbour & 0.83 & \underline{0.90} & 0.71 & 0.60 & 0.61 & 0.85 & \underline{0.94} & 0.76 & 0.63 & 0.63 \\
kNN, set mean, 10 neighbours & 0.85 & \textbf{0.91} & 0.71 & 0.60 & 0.60 & 0.87 & \underline{0.94} & 0.77 & 0.62 & 0.63 \\
Nearest centroid, set mean & 0.84 & \textbf{0.91} & \underline{0.78} & \textbf{0.62} & \underline{0.63} & 0.86 & \textbf{0.95} & \underline{0.85} & \underline{0.65} & \underline{0.67} \\
Nearest centroid, set mean, family & 0.82 & 0.83 & 0.60 & 0.55 & 0.53 & 0.86 & 0.86 & 0.64 & 0.55 & 0.53 \\
\midrule
\multicolumn{11}{@{}l}{\textit{Component checks}} \\
Nearest centroid, PCA space & 0.83 & 0.88 & 0.77 & \underline{0.61} & 0.62 & 0.86 & 0.92 & 0.84 & 0.64 & 0.66 \\
Nearest centroid, centroid span & 0.84 & 0.88 & 0.74 & \textbf{0.62} & \underline{0.63} & 0.86 & 0.92 & 0.79 & \textbf{0.66} & 0.66 \\
Family direction, 1-d PCA & 0.67 & -- & -- & -- & -- & 0.69 & -- & -- & -- & -- \\
Family direction, 1-d mean difference & 0.82 & -- & -- & -- & -- & 0.86 & -- & -- & -- & -- \\
\midrule
LineagePrint (LDA) & \underline{0.87} & 0.87 & 0.75 & \textbf{0.62} & 0.62 & \underline{0.88} & 0.90 & 0.79 & \underline{0.65} & 0.65 \\
LineagePrint (PLDA) & 0.85 & 0.81 & \textbf{0.83} & \underline{0.61} & \textbf{0.66} & 0.87 & 0.86 & \textbf{0.86} & 0.64 & \textbf{0.69} \\
LineagePrint (LDA), generator subspace & 0.85 & -- & -- & -- & -- & 0.87 & -- & -- & -- & -- \\
LineagePrint (PLDA), generator subspace & 0.86 & -- & -- & -- & -- & 0.87 & -- & -- & -- & -- \\
\bottomrule
\end{tabular}
\end{table*}
 
\noindent\textbf{Lineage at scale.}
\Cref{tab:lineage} gives the full results of \cref{sec:exp_scale} for both set sizes and both LineagePrint variants.
Each of 10 draws registers 32 generators of one family and queries 83 unseen generators per family, with sets of $k=8$ or 16.
For family attribution both families register 32 generators and LineagePrint uses family labels.
Baselines are linear classifiers and set-mean distances on the same fitting images and sets.
Distances of the set mean flag the other family in both directions.
With SD1/SD2 registered, the raw set-mean nearest centroid leads LineagePrint (LDA) in every draw (0.91 against 0.87 at $k=8$, 0.95 against 0.90 at $k=16$).
With SDXL registered, LineagePrint (PLDA) leads the raw centroid in 10 of 10 draws at $k=8$ (0.83 against 0.78) and is on par at $k=16$ (0.86 against 0.85, ahead in 8 of 10 draws).
Classifier confidence stays at 0.32 to 0.43 with SD1/SD2 registered and 0.62 to 0.73 with SDXL registered.
On family attribution LineagePrint (0.85 to 0.88) is on par with the linear classifiers (0.86 to 0.90), and the generator-label classifier is the best method (0.88 and 0.90).
Unseen fine-tunes of the registered family score 0.61 to 0.69 for LineagePrint and 0.58 to 0.69 for every set-mean distance to the registered generators.

\noindent\textbf{Attribution to the base model.}
\Cref{tab:tree} lists the eight models of \cref{sec:exp_tree} that are attributed to the base model of their family.
Each one is attributed to its base by majority, 551 of 576 sets in all.

\begin{table*}[t]
\centering
\caption{\textbf{Discovery per processing pipeline.} Setting of \cref{tab:discovery}: subspace fit in the COCO dataset, discovery of AuraFlow and SD3.5 among 10 registered generators in the Flickr dataset, sets of $k=8$, mean over 3 splits, chance 0.50. The augmented fit uses 12 versions of every fitting image (\cref{sec:supp_worlds}). The clean fit uses clean images only, so every pipeline except Clean is new for it. Pipelines: JPEG at quality 70 and 80, Gaussian blur ($\sigma=1$), WebP at quality 75, unsharp masking then JPEG 85 (Sharp.), gamma 0.85 and saturation 1.2 then JPEG 85 (Color), Gaussian noise ($\sigma=3$), $3\times3$ median filter, and a 4-pixel crop then JPEG 85. MSP, Energy and kNN are per-image scores averaged over the set, as in \cref{tab:discovery}. Mahalanobis, per image: the standard per-image Mahalanobis row of \cref{tab:discovery}. Mahalanobis, set mean, centered: the score of the pool-centered set mean (the clean fit has no per-image Mahalanobis run). Best in bold, second best underlined within each block.}
\label{tab:supp_pipelines}
\small
\setlength{\tabcolsep}{3.6pt}
\begin{tabular}{@{}lcccc@{\hspace{8pt}}cccccc@{\hspace{8pt}}cc@{}}
\toprule
& \multicolumn{4}{c}{Fitting or close} & \multicolumn{6}{c}{Never used in fitting} & \multicolumn{2}{c}{Mean} \\
\cmidrule(lr){2-5}\cmidrule(lr){6-11}\cmidrule(l){12-13}
Method & Clean & JPEG70 & JPEG80 & Blur & WebP & Sharp. & Color & Noise & Median & Crop & New 6 & All 10 \\
\midrule
\multicolumn{13}{@{}l}{\textit{CF ViT-S/16, augmented fit}} \\
MSP & 0.61 & 0.64 & 0.69 & 0.66 & 0.75 & 0.58 & 0.60 & 0.58 & 0.57 & 0.48 & 0.59 & 0.62 \\
Energy & 0.71 & 0.76 & 0.75 & 0.74 & 0.79 & 0.74 & 0.72 & 0.62 & 0.59 & 0.75 & 0.70 & 0.72 \\
Mahalanobis, per image & 0.86 & 0.78 & 0.83 & 0.82 & 0.68 & 0.83 & 0.82 & 0.70 & 0.85 & 0.85 & 0.79 & 0.80 \\
kNN & 0.75 & 0.46 & 0.73 & 0.77 & 0.69 & 0.82 & 0.76 & 0.70 & 0.78 & 0.78 & 0.75 & 0.72 \\
\cmidrule(l){1-13}
LineagePrint (LDA) & \textbf{0.97} & \textbf{0.92} & \textbf{0.94} & \textbf{0.97} & \textbf{0.87} & \textbf{0.90} & \textbf{0.96} & \underline{0.82} & \textbf{0.94} & \textbf{0.93} & \textbf{0.91} & \textbf{0.92} \\
LineagePrint (PLDA) & \underline{0.94} & \underline{0.89} & \underline{0.90} & \underline{0.94} & \underline{0.84} & \underline{0.89} & \underline{0.94} & \textbf{0.83} & \underline{0.92} & \underline{0.91} & \underline{0.89} & \underline{0.90} \\
\midrule
\multicolumn{13}{@{}l}{\textit{CF ViT-S/16, clean fit}} \\
MSP & 0.81 & 0.48 & 0.45 & 0.50 & 0.62 & 0.59 & 0.51 & 0.52 & 0.50 & 0.47 & 0.54 & 0.54 \\
Energy & 0.81 & 0.38 & 0.48 & 0.54 & 0.62 & 0.63 & 0.58 & 0.48 & 0.46 & 0.50 & 0.54 & 0.55 \\
Mahalanobis, set mean, centered & 0.84 & 0.51 & 0.55 & \underline{0.77} & 0.41 & 0.63 & 0.55 & 0.56 & 0.74 & 0.57 & 0.58 & 0.61 \\
kNN & 0.68 & 0.38 & 0.43 & 0.64 & 0.45 & 0.51 & 0.47 & 0.48 & 0.70 & 0.46 & 0.51 & 0.52 \\
\cmidrule(l){1-13}
LineagePrint (LDA) & \textbf{0.95} & \textbf{0.79} & \textbf{0.79} & \textbf{0.91} & \textbf{0.72} & \textbf{0.78} & \textbf{0.79} & \textbf{0.87} & \textbf{0.94} & \textbf{0.78} & \textbf{0.81} & \textbf{0.83} \\
LineagePrint (PLDA) & \underline{0.92} & \underline{0.68} & \underline{0.68} & 0.76 & \underline{0.71} & \underline{0.71} & \underline{0.67} & \underline{0.63} & \underline{0.91} & \underline{0.64} & \underline{0.71} & \underline{0.73} \\
\midrule
\multicolumn{13}{@{}l}{\textit{CLIP ViT-B/32, augmented fit}} \\
MSP & 0.53 & 0.49 & 0.63 & 0.52 & 0.55 & \underline{0.62} & 0.57 & 0.52 & 0.56 & 0.56 & 0.56 & 0.55 \\
Energy & \underline{0.64} & \underline{0.62} & \underline{0.66} & \underline{0.68} & \underline{0.62} & 0.54 & \underline{0.65} & 0.56 & \underline{0.65} & \underline{0.65} & \underline{0.61} & \underline{0.63} \\
Mahalanobis, per image & 0.47 & 0.49 & 0.48 & 0.45 & 0.48 & 0.46 & 0.46 & 0.49 & 0.46 & 0.45 & 0.47 & 0.47 \\
kNN & 0.38 & 0.42 & 0.39 & 0.41 & 0.40 & 0.37 & 0.39 & 0.38 & 0.37 & 0.35 & 0.38 & 0.39 \\
\cmidrule(l){1-13}
LineagePrint (LDA) & 0.62 & 0.61 & 0.62 & 0.62 & 0.61 & 0.59 & 0.64 & \underline{0.61} & 0.64 & 0.61 & \underline{0.61} & 0.62 \\
LineagePrint (PLDA) & \textbf{0.69} & \textbf{0.67} & \textbf{0.68} & \textbf{0.69} & \textbf{0.67} & \textbf{0.63} & \textbf{0.70} & \textbf{0.67} & \textbf{0.70} & \textbf{0.68} & \textbf{0.68} & \textbf{0.68} \\
\midrule
\multicolumn{13}{@{}l}{\textit{CLIP ViT-B/32, clean fit}} \\
MSP & 0.61 & 0.57 & 0.60 & 0.67 & 0.61 & 0.53 & 0.50 & 0.48 & 0.69 & 0.63 & 0.57 & 0.59 \\
Energy & \underline{0.76} & 0.63 & 0.68 & \textbf{0.75} & 0.63 & 0.57 & 0.61 & 0.57 & \underline{0.73} & \underline{0.70} & 0.64 & \underline{0.66} \\
Mahalanobis, set mean, centered & 0.54 & 0.53 & 0.50 & 0.48 & 0.51 & 0.51 & 0.53 & 0.52 & 0.52 & 0.51 & 0.52 & 0.52 \\
kNN & 0.41 & 0.44 & 0.42 & 0.40 & 0.43 & 0.43 & 0.45 & 0.42 & 0.43 & 0.41 & 0.43 & 0.42 \\
\cmidrule(l){1-13}
LineagePrint (LDA) & 0.71 & \underline{0.64} & \underline{0.69} & 0.65 & \underline{0.68} & \underline{0.64} & \underline{0.66} & \underline{0.63} & 0.69 & 0.66 & \underline{0.66} & \underline{0.66} \\
LineagePrint (PLDA) & \textbf{0.77} & \textbf{0.71} & \textbf{0.77} & \underline{0.73} & \textbf{0.76} & \textbf{0.72} & \textbf{0.75} & \textbf{0.73} & \textbf{0.74} & \textbf{0.75} & \textbf{0.74} & \textbf{0.74} \\
\bottomrule
\end{tabular}
\end{table*}
\begin{table*}[t]
\centering
\caption{\textbf{Effect of the encoder.} Discovery AUROC in the setting of \cref{tab:discovery}, mean over all ten pipelines, sets of $k=8$, mean over 3 splits, chance 0.50, for the clean fit (C) and the augmented fit (A). MSP, Energy and kNN are per-image scores averaged over the set. Mahalanobis, per image: the standard per-image score of \cref{tab:discovery}, run for CF and CLIP ViT-B/32 in the augmented fit (-- elsewhere). Mahalanobis, set mean, centered: the score of the pool-centered set mean. Best in bold, second best underlined. LineagePrint beats every score of this table with CF, CLIP ViT-B/32 and SigLIP features in both fits and with CLIP ViT-L/14 in the clean fit. With DINOv2-L features the baselines score higher. The forensic CF encoder gives the largest margin.}
\label{tab:supp_encoders}
\small
\setlength{\tabcolsep}{4pt}
\begin{tabular}{@{}lcccccccccc@{}}
\toprule
& \multicolumn{2}{c}{CF ViT-S/16} & \multicolumn{2}{c}{CLIP ViT-L/14} & \multicolumn{2}{c}{CLIP ViT-B/32} & \multicolumn{2}{c}{DINOv2-L} & \multicolumn{2}{c}{SigLIP} \\
\cmidrule(lr){2-3}\cmidrule(lr){4-5}\cmidrule(lr){6-7}\cmidrule(lr){8-9}\cmidrule(l){10-11}
Method & C & A & C & A & C & A & C & A & C & A \\
\midrule
MSP & 0.54 & 0.62 & 0.59 & \underline{0.60} & 0.59 & 0.55 & \underline{0.63} & \underline{0.57} & 0.56 & 0.58 \\
Energy & 0.55 & 0.72 & \underline{0.63} & \textbf{0.68} & \underline{0.66} & \underline{0.63} & \textbf{0.64} & \textbf{0.68} & 0.57 & 0.64 \\
kNN & 0.52 & 0.72 & 0.44 & 0.42 & 0.42 & 0.39 & 0.57 & 0.43 & 0.46 & 0.44 \\
Mahalanobis, per image & -- & 0.80 & -- & -- & -- & 0.47 & -- & -- & -- & -- \\
Mahalanobis, set mean, centered & 0.61 & 0.68 & 0.49 & 0.48 & 0.52 & 0.53 & 0.54 & 0.53 & 0.50 & 0.50 \\
\midrule
LineagePrint (LDA) & \textbf{0.83} & \textbf{0.92} & 0.60 & 0.53 & \underline{0.66} & 0.62 & 0.53 & 0.53 & \underline{0.74} & \underline{0.71} \\
LineagePrint (PLDA) & \underline{0.73} & \underline{0.90} & \textbf{0.70} & \underline{0.60} & \textbf{0.74} & \textbf{0.68} & 0.55 & 0.56 & \textbf{0.76} & \textbf{0.73} \\
\bottomrule
\end{tabular}
\end{table*}
 
\noindent\textbf{Discovery per pipeline.}
\Cref{tab:supp_pipelines} breaks \cref{tab:discovery} down by pipeline.
In the augmented fit with CF features LineagePrint (LDA) beats the four standard per-image scores on each of the 10 pipelines.
It ranges from 0.82 (noise) to 0.97 (clean and blur), and per-image Mahalanobis ranges from 0.68 (WebP) to 0.86 (clean).
In the clean fit every processed pipeline is new, and LineagePrint (LDA) beats the four scores of this block on each pipeline.
It ranges from 0.72 (WebP) to 0.95 (clean), with 0.82 on the nine processed pipelines against 0.59 for Mahalanobis on the centered set mean, the best other score of this block.
With CLIP ViT-B/32 features LineagePrint (PLDA) leads the four scores of its block on all 10 pipelines in the augmented fit and on 9 of 10 in the clean fit. The comparison with set-mean baselines is in \cref{tab:supp_equal_treatment}.

\noindent\textbf{Encoder sweep.}
\Cref{tab:supp_encoders} compares five encoders on the ten-pipeline mean.
LineagePrint beats every score of this table with CF, CLIP ViT-B/32 and SigLIP features in both fits, and with CLIP ViT-L/14 features in the clean fit.
The margin of the best LineagePrint score over the best score of this table is +0.12 (augmented fit) and +0.22 (clean fit) with CF features.
It is +0.09 and +0.19 with SigLIP, +0.05 and +0.08 with CLIP ViT-B/32, and +0.07 with CLIP ViT-L/14 in the clean fit.
With DINOv2-L features the Energy and MSP baselines score highest, with Energy ahead of LineagePrint by 0.12 (augmented fit) and 0.09 (clean fit), and with CLIP ViT-L/14 features in the augmented fit Energy leads by 0.08.
The forensic CF encoder gives the largest margin in this test.
Per-image Mahalanobis and the comparison with set-mean baselines (\cref{tab:supp_equal_treatment}) cover CF and CLIP ViT-B/32.

\begin{table*}[t]
\centering
\caption{\textbf{Every baseline with the same set averaging and pool.} Setting of \cref{tab:discovery} (sets of $k=8$, mean over 3 splits, chance 0.50). Per image: the per-image score averaged over the set. Set mean: the score of the set's mean feature. Set: a two-sample test between the set and each registered generator's fitting images, scored by the smallest statistic. Centered: the pool mean minus the fitting mean is subtracted before scoring, with the pool of \cref{tab:discovery}. Every classifier, scaler, neighbour bank and covariance is fit on the same images as LineagePrint. Best in bold, second best underlined in each column.}
\label{tab:supp_equal_treatment}
\small
\setlength{\tabcolsep}{2.8pt}
\begin{tabular}{@{}lccc@{\hspace{6pt}}ccc@{\hspace{6pt}}ccc@{\hspace{6pt}}ccc@{}}
\toprule
& \multicolumn{6}{c}{CF ViT-S/16} & \multicolumn{6}{c}{CLIP ViT-B/32} \\
\cmidrule(lr){2-7}\cmidrule(l){8-13}
& \multicolumn{3}{c}{Uncentered} & \multicolumn{3}{c}{Centered} & \multicolumn{3}{c}{Uncentered} & \multicolumn{3}{c}{Centered} \\
\cmidrule(lr){2-4}\cmidrule(lr){5-7}\cmidrule(lr){8-10}\cmidrule(l){11-13}
Method & Clean & New & All & Clean & New & All & Clean & New & All & Clean & New & All \\
\midrule
MSP, per image & 0.61 & 0.59 & 0.62 & 0.66 & 0.55 & 0.58 & 0.53 & 0.56 & 0.55 & 0.56 & 0.53 & 0.54 \\
MSP, set mean & 0.76 & 0.71 & 0.74 & 0.81 & 0.77 & 0.78 & 0.62 & 0.58 & 0.60 & \underline{0.69} & \underline{0.68} & \underline{0.68} \\
Energy, per image & 0.71 & 0.70 & 0.72 & 0.59 & 0.62 & 0.63 & 0.64 & 0.61 & 0.63 & 0.65 & 0.64 & 0.64 \\
Energy, set mean & 0.88 & \textbf{0.81} & \textbf{0.84} & 0.81 & 0.84 & 0.84 & \textbf{0.72} & \textbf{0.66} & \textbf{0.69} & \textbf{0.78} & \textbf{0.76} & \textbf{0.76} \\
Mahalanobis, per image & 0.86 & \underline{0.79} & 0.80 & 0.86 & 0.80 & 0.81 & 0.47 & 0.47 & 0.47 & 0.46 & 0.46 & 0.46 \\
Mahalanobis, set mean & 0.78 & 0.60 & 0.65 & 0.80 & 0.63 & 0.68 & 0.54 & 0.54 & 0.55 & 0.54 & 0.52 & 0.53 \\
kNN, per image & 0.75 & 0.75 & 0.72 & 0.69 & 0.67 & 0.66 & 0.38 & 0.38 & 0.39 & 0.39 & 0.39 & 0.40 \\
kNN, set mean & 0.72 & 0.72 & 0.69 & 0.59 & 0.59 & 0.58 & 0.52 & 0.49 & 0.50 & 0.50 & 0.49 & 0.50 \\
\midrule
RBF-MMD, set & 0.17 & 0.37 & 0.35 & 0.59 & 0.51 & 0.51 & 0.51 & 0.55 & 0.55 & 0.55 & 0.55 & 0.56 \\
Energy distance, set & 0.17 & 0.36 & 0.34 & 0.55 & 0.49 & 0.49 & 0.50 & 0.55 & 0.55 & 0.55 & 0.55 & 0.55 \\
\midrule
LineagePrint (LDA) & \textbf{0.92} & 0.76 & 0.82 & \textbf{0.97} & \textbf{0.91} & \textbf{0.92} & 0.57 & 0.54 & 0.55 & 0.62 & 0.61 & 0.62 \\
LineagePrint (PLDA) & \underline{0.91} & \underline{0.79} & \underline{0.83} & \underline{0.94} & \underline{0.89} & \underline{0.90} & \underline{0.67} & \underline{0.63} & \underline{0.64} & \underline{0.69} & \underline{0.68} & \underline{0.68} \\
\bottomrule
\end{tabular}
\end{table*}
\begin{table}[t]
\centering
\caption{\textbf{What each part of LineagePrint adds.} Setting of \cref{tab:discovery}, CF features, discovery AUROC, mean over 3 splits, chance 0.50. $k$: images per set. Centered: the pool mean minus the fitting mean is subtracted. Distance in: raw features (standardized), PCA space (11 dimensions) and centroid span (the span of the 12 generator centroids) use the distance of the set mean to the nearest centroid without whitening. Full space, whitened: Mahalanobis distance with the within-generator covariance. LDA subspace: LineagePrint (LDA). Best in bold, second best underlined.}
\label{tab:ladder}
\small
\setlength{\tabcolsep}{3pt}
\begin{tabular}{@{}cclccc@{}}
\toprule
& & & \multicolumn{3}{c}{Discovery AUROC $\uparrow$} \\
\cmidrule(l){4-6}
$k$ & Centered & Distance in & Clean & New & All \\
\midrule
$1$ & yes & LDA subspace & 0.64 & 0.59 & 0.59 \\
$8$ & no & LDA subspace & \underline{0.92} & \underline{0.76} & \underline{0.82} \\
$8$ & yes & raw features & 0.73 & 0.63 & 0.65 \\
$8$ & yes & PCA space & 0.60 & 0.57 & 0.55 \\
$8$ & yes & centroid span & 0.68 & 0.64 & 0.64 \\
$8$ & yes & full space, whitened & 0.80 & 0.63 & 0.68 \\
\midrule
$8$ & yes & LDA subspace & \textbf{0.97} & \textbf{0.91} & \textbf{0.92} \\
\bottomrule
\end{tabular}
\end{table}
 
\noindent\textbf{Every baseline with the same set averaging and pool.}
\Cref{tab:supp_equal_treatment} gives every baseline both aggregations, with and without the pool centering of LineagePrint.
With CF features the strongest baseline is Energy on the set mean, 0.88 on clean images without centering and 0.84 on the six new pipelines with centering.
Centering helps Energy on the set mean only on the six new pipelines (0.81 to 0.84) and lowers it on clean images (0.88 to 0.81).
LineagePrint (LDA) with centering has the highest mean in every column (0.97, 0.91 and 0.92).
Over 20 split seeds it is ahead of the strongest baseline in 20 of 20 seeds on the six new pipelines (mean paired difference 0.09), in 20 of 20 over all ten pipelines and in 15 of 20 on clean images.
With non-overlapping test sets, 2 to 5 disjoint sets per generator, it scores 0.91 on the six new pipelines and is ahead in 17 of 20 seeds.
With CLIP ViT-B/32 features Energy on the centered set mean is the strongest score (0.78, 0.76 and 0.76), ahead of LineagePrint (PLDA) (0.69, 0.68 and 0.68), and over 20 seeds on the six new pipelines LineagePrint (PLDA) leads it in 4 of 20 (0.70 against 0.74).
In this configuration, the RBF-MMD and energy-distance two-sample scores give 0.51 and 0.49 on the six new pipelines with centering, and 0.37 and 0.36 without.

\noindent\textbf{What each part adds.}
\Cref{tab:ladder} changes one part of LineagePrint at a time (CF, six new pipelines).
One image per set gives 0.59, and the set mean without pool centering gives 0.76, so set averaging adds 0.17 and centering 0.15.
With centering, the nearest centroid of the set mean gives 0.63 in raw features, 0.57 in a PCA space of 11 dimensions, 0.64 in the span of the 12 generator centroids and 0.63 with the whitened Mahalanobis metric, against 0.91 in the LDA subspace.

\begin{table*}[t]
\centering
\caption{\textbf{Discovery per unseen generator over 20 split seeds.} Setting and rows of \cref{tab:discovery}, CF features, sets of $k=8$, mean over 20 split seeds (SD over seeds on the six new pipelines), chance 0.50. Both unseen: AuraFlow and SD3.5 sets together, as in \cref{tab:discovery}. Per generator: sets of one unseen generator against the same registered sets. Best in bold, second best underlined.}
\label{tab:supp_per_unseen}
\small
\setlength{\tabcolsep}{3pt}
\begin{tabular}{@{}l ccc@{\hspace{8pt}}ccc@{\hspace{8pt}}ccc@{}}
\toprule
& \multicolumn{3}{c}{Both unseen} & \multicolumn{3}{c}{AuraFlow} & \multicolumn{3}{c}{SD3.5} \\
\cmidrule(lr){2-4}\cmidrule(lr){5-7}\cmidrule(l){8-10}
Method & Clean & New & All & Clean & New & All & Clean & New & All \\
\midrule
\multicolumn{10}{@{}l}{\textit{Standard per-image scores}} \\
MSP~\cite{hendrycks2017msp} & 0.67 & 0.61\,{\scriptsize$\pm$0.05} & 0.63 & 0.70 & 0.63\,{\scriptsize$\pm$0.06} & 0.65 & 0.64 & 0.59\,{\scriptsize$\pm$0.07} & 0.61 \\
Energy~\cite{liu2020energy} & 0.73 & 0.69\,{\scriptsize$\pm$0.05} & 0.70 & 0.80 & 0.69\,{\scriptsize$\pm$0.06} & 0.72 & 0.65 & 0.68\,{\scriptsize$\pm$0.07} & 0.69 \\
Mahalanobis~\cite{lee2018mahalanobis} & 0.88 & 0.80\,{\scriptsize$\pm$0.04} & 0.82 & 0.76 & 0.69\,{\scriptsize$\pm$0.07} & 0.71 & \textbf{1.00} & 0.90\,{\scriptsize$\pm$0.03} & 0.92 \\
kNN~\cite{sun2022knn} & 0.77 & 0.76\,{\scriptsize$\pm$0.04} & 0.74 & 0.53 & 0.59\,{\scriptsize$\pm$0.08} & 0.60 & \textbf{1.00} & \underline{0.92}\,{\scriptsize$\pm$0.02} & 0.88 \\
\midrule
\multicolumn{10}{@{}l}{\textit{Set-mean baselines}} \\
Energy & 0.92 & 0.81\,{\scriptsize$\pm$0.03} & 0.85 & \textbf{0.95} & 0.80\,{\scriptsize$\pm$0.03} & 0.85 & 0.88 & 0.81\,{\scriptsize$\pm$0.05} & 0.85 \\
Energy, centered & 0.81 & 0.82\,{\scriptsize$\pm$0.06} & 0.83 & \textbf{0.95} & \textbf{0.91}\,{\scriptsize$\pm$0.04} & \textbf{0.92} & 0.68 & 0.72\,{\scriptsize$\pm$0.10} & 0.73 \\
Mahalanobis, centered & 0.82 & 0.67\,{\scriptsize$\pm$0.05} & 0.71 & 0.65 & 0.45\,{\scriptsize$\pm$0.09} & 0.50 & \underline{0.99} & 0.90\,{\scriptsize$\pm$0.03} & 0.92 \\
Nearest centroid, centered & 0.73 & 0.64\,{\scriptsize$\pm$0.05} & 0.66 & 0.50 & 0.42\,{\scriptsize$\pm$0.08} & 0.46 & 0.95 & 0.86\,{\scriptsize$\pm$0.05} & 0.86 \\
\midrule
LineagePrint (LDA) & \textbf{0.96} & \textbf{0.91}\,{\scriptsize$\pm$0.02} & \textbf{0.92} & 0.93 & 0.85\,{\scriptsize$\pm$0.04} & 0.87 & \underline{0.99} & \textbf{0.96}\,{\scriptsize$\pm$0.02} & \textbf{0.97} \\
LineagePrint (PLDA) & \underline{0.94} & \underline{0.90}\,{\scriptsize$\pm$0.02} & \underline{0.91} & \underline{0.94} & \underline{0.88}\,{\scriptsize$\pm$0.03} & \underline{0.90} & 0.95 & \underline{0.92}\,{\scriptsize$\pm$0.04} & \underline{0.93} \\
\bottomrule
\end{tabular}
\end{table*}
 
\noindent\textbf{Each unseen generator.}
\Cref{tab:supp_per_unseen} scores AuraFlow and SD3.5 separately over 20 split seeds.
On the six new pipelines LineagePrint (LDA) discovers SD3.5 with AUROC 0.96, against 0.92 for per-image kNN and 0.93 for per-image Mahalanobis with the pool centering (20 seeds).
AuraFlow is discovered with 0.85 (LDA) and 0.88 (PLDA), and Energy on the centered set mean reaches 0.91 on it.

\noindent\textbf{Leave one generator out.}
Ten of the 12 unseen generators of the cross-dataset test are the shared generators, each held out of the fit in turn (\cref{tab:supp_unseen_generators}).
On the six new pipelines LineagePrint (LDA) ranges from 0.66 (PixArt-$\Sigma$) to 0.97 (FLUX.1-dev and SDXL) over the ten.

\begin{table*}[t]
\centering
\caption{\textbf{Which pool centers the fingerprint.} Setting of \cref{tab:discovery}, CF features, discovery AUROC on the six pipelines never used in fitting, sets of $k=8$, mean over 3 splits, chance 0.50. Each pool passes through the test pipeline, and every method subtracts the same pool mean minus fitting mean. The second row is the pool of \cref{tab:discovery}. Images: pool size per split and pipeline. Pools of 100, 30 and 10 images are drawn from the registered-only pool (mean over 5 draws), and one registered generator is a mean over the 10 registered generators. Maha.: Mahalanobis. MMD: RBF-kernel maximum mean discrepancy. E-dist.: energy distance. The two set-level tests compare the set with each registered generator's fitting images. Best in each row in bold, second best underlined.}
\label{tab:supp_pool}
\small
\setlength{\tabcolsep}{3.4pt}
\begin{tabular}{@{}lc cc@{\hspace{8pt}}ccc@{\hspace{8pt}}ccc@{\hspace{8pt}}cc@{}}
\toprule
& & \multicolumn{2}{c}{LineagePrint} & \multicolumn{3}{c}{Set mean} & \multicolumn{3}{c}{Per image} & \multicolumn{2}{c}{Set level} \\
\cmidrule(lr){3-4}\cmidrule(lr){5-7}\cmidrule(lr){8-10}\cmidrule(l){11-12}
Pool & Images & LDA & PLDA & Energy & MSP & Maha. & Energy & Maha. & kNN & MMD & E-dist. \\
\midrule
No centering & -- & 0.76 & \underline{0.79} & \textbf{0.81} & 0.71 & 0.60 & 0.70 & \underline{0.79} & 0.75 & 0.37 & 0.36 \\
\midrule
20\% of every test generator & 180--213 & \textbf{0.91} & \underline{0.89} & 0.84 & 0.77 & 0.63 & 0.62 & 0.80 & 0.67 & 0.51 & 0.49 \\
Registered generators only & 153--181 & \textbf{0.90} & \underline{0.88} & 0.79 & 0.74 & 0.67 & 0.58 & 0.81 & 0.64 & 0.55 & 0.54 \\
Registered only, 100 images & 100 & \textbf{0.89} & \underline{0.87} & 0.80 & 0.74 & 0.67 & 0.59 & 0.81 & 0.64 & 0.56 & 0.54 \\
Registered only, 30 images & 30 & \textbf{0.87} & \underline{0.85} & 0.77 & 0.72 & 0.69 & 0.58 & 0.81 & 0.64 & 0.57 & 0.55 \\
Registered only, 10 images & 10 & \underline{0.79} & 0.78 & 0.71 & 0.69 & 0.65 & 0.55 & \textbf{0.81} & 0.62 & 0.52 & 0.51 \\
One registered generator & 10--24 & 0.70 & \underline{0.73} & 0.65 & 0.62 & 0.62 & 0.55 & \textbf{0.80} & 0.61 & 0.51 & 0.50 \\
\midrule
Real photographs & 203--228 & 0.37 & 0.57 & \textbf{0.76} & 0.59 & 0.25 & 0.74 & \underline{0.75} & \textbf{0.76} & 0.19 & 0.19 \\
\bottomrule
\end{tabular}
\end{table*}
 
\noindent\textbf{Which pool centers the fingerprint.}
\Cref{tab:supp_pool} changes the centering pool (CF, six new pipelines).
A pool of registered generators only, with no output of the unseen generators, gives 0.90 against 0.91 for the pool of \cref{tab:discovery}, and over 20 seeds it leads the strongest baseline under the same pool (per-image Mahalanobis, centered) in 18 of 20.
Pools of 100 and 30 registered-generator images give 0.89 and 0.87.
With 10 pool images or the outputs of one registered generator LineagePrint (LDA) gives 0.79 and 0.70, and per-image Mahalanobis leads (0.81 and 0.80).
A pool of real photographs through the same pipeline gives 0.37.
With this pool every score except Mahalanobis on the set mean (0.25) and the two-sample scores (0.19) beats LineagePrint (LDA), up to 0.76 (per-image kNN and Energy on the set mean).
The pool must hold generated outputs from the test pipeline. Outputs of the registered generators that went through that pipeline suffice (\cref{tab:supp_poolchannel}), so the source of the anchor sets also provides the pool, as in the speech experiment. For an unlabeled pool collected in the wild, a real-versus-generated detector can first remove real photographs, a step we did not test.

\begin{table}[t]
\centering
\caption{\textbf{Learned projection head vs the closed form.} Discovery AUROC in the Flickr dataset, mean over the six pipelines never used in fitting, then over 3 seeds, chance 0.50. Learned heads are trained for 8 epochs on COCO-dataset images under the same 12 image versions as the closed form. This comparison uses its own data split, so its closed-form values differ slightly from \cref{tab:discovery}. Best in bold, second best underlined. With CF features the closed form matches the best learned head. With CLIP ViT-B/32 features a learned head adds about 0.06.}
\label{tab:supp_learned_head}
\small
\setlength{\tabcolsep}{1.8pt}
\begin{tabular}{@{}lcccc@{}}
\toprule
& \multicolumn{4}{c}{Discovery AUROC $\uparrow$} \\
\cmidrule(l){2-5}
& \multicolumn{2}{c}{CF ViT-S/16} & \multicolumn{2}{c}{CLIP ViT-B/32} \\
\cmidrule(lr){2-3}\cmidrule(l){4-5}
Projection & $k=8$ & $k=16$ & $k=8$ & $k=16$ \\
\midrule
\multicolumn{5}{@{}l}{\textit{Closed form}} \\
LineagePrint (LDA) & \textbf{0.92} & \textbf{0.98} & 0.70 & 0.82 \\
LineagePrint (PLDA) & 0.86 & 0.94 & 0.68 & 0.82 \\
\midrule
\multicolumn{5}{@{}l}{\textit{Learned projection head}} \\
SupCon head~\cite{khosla2020supcon} & \underline{0.91} & 0.96 & \underline{0.75} & \underline{0.86} \\
SupCon head + invariance & 0.90 & 0.95 & \textbf{0.76} & \textbf{0.88} \\
ArcFace head~\cite{deng2019arcface} & \textbf{0.92} & \underline{0.97} & 0.72 & 0.82 \\
\bottomrule
\end{tabular}
\end{table}
 
\noindent\textbf{Learned head vs closed form.}
\Cref{tab:supp_learned_head} trains three projection heads on the same COCO-dataset data as the closed form.
With CF features learned projection heads~\cite{khosla2020supcon,deng2019arcface} add less than 0.01.
With CF features the best learned head (ArcFace) matches LineagePrint (LDA) at 0.92 for $k=8$, and LineagePrint (LDA) leads at $k=16$ (0.98 against 0.97).
With CLIP ViT-B/32 features the SupCon head with the invariance term adds 0.06 at both set sizes.
With CF features the closed form therefore matches the best learned head with no gradient training.

\clearpage  %
\section{Confirmation on fresh captions}
\label{sec:supp_fresh}

\begin{table*}[t]
\centering
\caption{\textbf{Every score on fresh captions.} The confirmation run of \cref{sec:supp_fresh}: AUROC per unseen model, averaged over the 29 unseen models on the fresh captions (left) and over the 25 unseen models on the Family-Tree captions (right, the test of \cref{tab:supp_unseen_scores} rerun with the same fit and sets), CF features, $k=8$, columns as in \cref{tab:unseen}. Second line: 95\% bootstrap interval of LineagePrint (LDA) minus that row, over families and then models. Per-image scores are averaged over the set, and set-mean scores score the mean feature of the set. Centered: pool mean minus fitting mean subtracted. Nearest centroid: standardized raw features.}
\label{tab:supp_fresh_scores}
\footnotesize
\setlength{\tabcolsep}{4pt}
\begin{tabular}{@{}l ccc@{\hspace{10pt}}ccc@{}}
\toprule
& \multicolumn{3}{c}{Fresh captions, 29 models} & \multicolumn{3}{c}{Family-Tree captions, 25 models} \\
\cmidrule(lr){2-4}\cmidrule(l){5-7}
Method & Clean & New & All & Clean & New & All \\
\midrule
LineagePrint (LDA) & 0.81 & 0.86 & 0.86 & 0.82 & 0.86 & 0.86 \\
\midrule
\multicolumn{7}{@{}l}{\textit{Per-image scores averaged over the set}} \\
MSP & 0.77 & 0.61 & 0.66 & 0.72 & 0.59 & 0.63 \\
 & {\scriptsize [$-$0.09, +0.17]} & {\scriptsize [+0.15, +0.34]} & {\scriptsize [+0.11, +0.30]} & {\scriptsize [$-$0.02, +0.23]} & {\scriptsize [+0.15, +0.38]} & {\scriptsize [+0.13, +0.34]} \\
Energy & 0.80 & 0.65 & 0.69 & 0.74 & 0.61 & 0.65 \\
 & {\scriptsize [$-$0.12, +0.16]} & {\scriptsize [+0.11, +0.33]} & {\scriptsize [+0.07, +0.29]} & {\scriptsize [$-$0.06, +0.24]} & {\scriptsize [+0.12, +0.39]} & {\scriptsize [+0.10, +0.35]} \\
Mahalanobis & 0.57 & 0.61 & 0.62 & 0.56 & 0.62 & 0.62 \\
 & {\scriptsize [+0.08, +0.47]} & {\scriptsize [+0.17, +0.37]} & {\scriptsize [+0.16, +0.37]} & {\scriptsize [+0.10, +0.50]} & {\scriptsize [+0.14, +0.39]} & {\scriptsize [+0.14, +0.40]} \\
kNN & 0.57 & 0.54 & 0.55 & 0.56 & 0.54 & 0.55 \\
 & {\scriptsize [+0.05, +0.50]} & {\scriptsize [+0.22, +0.46]} & {\scriptsize [+0.23, +0.43]} & {\scriptsize [+0.07, +0.52]} & {\scriptsize [+0.21, +0.49]} & {\scriptsize [+0.22, +0.46]} \\
\midrule
\multicolumn{7}{@{}l}{\textit{Per-image scores averaged over the set, centered}} \\
MSP & 0.73 & 0.67 & 0.68 & 0.67 & 0.63 & 0.64 \\
 & {\scriptsize [$-$0.06, +0.20]} & {\scriptsize [+0.10, +0.27]} & {\scriptsize [+0.09, +0.26]} & {\scriptsize [+0.02, +0.29]} & {\scriptsize [+0.12, +0.33]} & {\scriptsize [+0.11, +0.33]} \\
Energy & 0.73 & 0.70 & 0.71 & 0.64 & 0.64 & 0.64 \\
 & {\scriptsize [$-$0.06, +0.23]} & {\scriptsize [+0.07, +0.26]} & {\scriptsize [+0.06, +0.26]} & {\scriptsize [+0.04, +0.34]} & {\scriptsize [+0.09, +0.37]} & {\scriptsize [+0.09, +0.37]} \\
Mahalanobis & 0.57 & 0.63 & 0.63 & 0.57 & 0.62 & 0.62 \\
 & {\scriptsize [+0.08, +0.47]} & {\scriptsize [+0.15, +0.37]} & {\scriptsize [+0.15, +0.37]} & {\scriptsize [+0.10, +0.50]} & {\scriptsize [+0.13, +0.40]} & {\scriptsize [+0.14, +0.40]} \\
kNN & 0.53 & 0.52 & 0.53 & 0.51 & 0.51 & 0.52 \\
 & {\scriptsize [+0.13, +0.47]} & {\scriptsize [+0.25, +0.45]} & {\scriptsize [+0.25, +0.45]} & {\scriptsize [+0.18, +0.51]} & {\scriptsize [+0.25, +0.49]} & {\scriptsize [+0.25, +0.48]} \\
\midrule
\multicolumn{7}{@{}l}{\textit{Set-mean scores}} \\
MSP & 0.85 & 0.69 & 0.75 & 0.81 & 0.65 & 0.71 \\
 & {\scriptsize [$-$0.17, +0.08]} & {\scriptsize [+0.06, +0.27]} & {\scriptsize [+0.01, +0.21]} & {\scriptsize [$-$0.11, +0.15]} & {\scriptsize [+0.09, +0.32]} & {\scriptsize [+0.05, +0.27]} \\
Energy & 0.86 & 0.72 & 0.76 & 0.81 & 0.67 & 0.72 \\
 & {\scriptsize [$-$0.18, +0.09]} & {\scriptsize [+0.05, +0.25]} & {\scriptsize [0.00, +0.21]} & {\scriptsize [$-$0.12, +0.17]} & {\scriptsize [+0.08, +0.31]} & {\scriptsize [+0.04, +0.27]} \\
Mahalanobis & 0.75 & 0.66 & 0.71 & 0.74 & 0.67 & 0.71 \\
 & {\scriptsize [0.00, +0.15]} & {\scriptsize [+0.16, +0.26]} & {\scriptsize [+0.12, +0.21]} & {\scriptsize [+0.02, +0.17]} & {\scriptsize [+0.14, +0.24]} & {\scriptsize [+0.11, +0.19]} \\
kNN & 0.55 & 0.53 & 0.53 & 0.56 & 0.52 & 0.53 \\
 & {\scriptsize [+0.05, +0.51]} & {\scriptsize [+0.22, +0.49]} & {\scriptsize [+0.23, +0.46]} & {\scriptsize [+0.06, +0.54]} & {\scriptsize [+0.19, +0.53]} & {\scriptsize [+0.21, +0.50]} \\
Nearest centroid & 0.59 & 0.63 & 0.64 & 0.59 & 0.64 & 0.65 \\
 & {\scriptsize [+0.02, +0.40]} & {\scriptsize [+0.16, +0.29]} & {\scriptsize [+0.15, +0.30]} & {\scriptsize [$-$0.03, +0.43]} & {\scriptsize [+0.14, +0.29]} & {\scriptsize [+0.13, +0.30]} \\
\midrule
\multicolumn{7}{@{}l}{\textit{Set-mean scores, centered}} \\
MSP & 0.82 & 0.82 & 0.82 & 0.77 & 0.76 & 0.77 \\
 & {\scriptsize [$-$0.15, +0.12]} & {\scriptsize [$-$0.05, +0.12]} & {\scriptsize [$-$0.05, +0.13]} & {\scriptsize [$-$0.07, +0.21]} & {\scriptsize [$-$0.01, +0.21]} & {\scriptsize [$-$0.01, +0.21]} \\
Energy & 0.81 & 0.81 & 0.81 & 0.73 & 0.74 & 0.74 \\
 & {\scriptsize [$-$0.13, +0.15]} & {\scriptsize [$-$0.03, +0.13]} & {\scriptsize [$-$0.03, +0.15]} & {\scriptsize [$-$0.05, +0.27]} & {\scriptsize [+0.01, +0.25]} & {\scriptsize [+0.01, +0.26]} \\
Mahalanobis & 0.77 & 0.75 & 0.76 & 0.76 & 0.74 & 0.76 \\
 & {\scriptsize [$-$0.01, +0.13]} & {\scriptsize [+0.06, +0.19]} & {\scriptsize [+0.06, +0.16]} & {\scriptsize [+0.01, +0.14]} & {\scriptsize [+0.05, +0.17]} & {\scriptsize [+0.06, +0.15]} \\
kNN & 0.51 & 0.48 & 0.49 & 0.47 & 0.48 & 0.48 \\
 & {\scriptsize [+0.14, +0.50]} & {\scriptsize [+0.26, +0.51]} & {\scriptsize [+0.26, +0.50]} & {\scriptsize [+0.20, +0.55]} & {\scriptsize [+0.26, +0.55]} & {\scriptsize [+0.27, +0.54]} \\
Nearest centroid & 0.77 & 0.75 & 0.76 & 0.78 & 0.73 & 0.75 \\
 & {\scriptsize [$-$0.03, +0.17]} & {\scriptsize [+0.06, +0.18]} & {\scriptsize [+0.06, +0.17]} & {\scriptsize [$-$0.03, +0.12]} & {\scriptsize [+0.07, +0.18]} & {\scriptsize [+0.07, +0.17]} \\
\midrule
\multicolumn{7}{@{}l}{\textit{LineagePrint variants}} \\
LineagePrint (LDA), uncentered & 0.80 & 0.67 & 0.75 & 0.81 & 0.69 & 0.76 \\
 & {\scriptsize [0.00, +0.02]} & {\scriptsize [+0.15, +0.22]} & {\scriptsize [+0.09, +0.13]} & {\scriptsize [$-$0.02, +0.03]} & {\scriptsize [+0.13, +0.20]} & {\scriptsize [+0.08, +0.12]} \\
LineagePrint (PLDA), uncentered & 0.89 & 0.74 & 0.81 & 0.86 & 0.72 & 0.79 \\
 & {\scriptsize [$-$0.16, 0.00]} & {\scriptsize [+0.07, +0.18]} & {\scriptsize [0.00, +0.11]} & {\scriptsize [$-$0.11, +0.04]} & {\scriptsize [+0.07, +0.20]} & {\scriptsize [+0.02, +0.14]} \\
LineagePrint (PLDA) & 0.90 & 0.88 & 0.89 & 0.87 & 0.85 & 0.86 \\
 & {\scriptsize [$-$0.18, 0.00]} & {\scriptsize [$-$0.07, +0.04]} & {\scriptsize [$-$0.08, +0.03]} & {\scriptsize [$-$0.13, +0.06]} & {\scriptsize [$-$0.05, +0.09]} & {\scriptsize [$-$0.06, +0.08]} \\
\bottomrule
\end{tabular}
\end{table*}
\begin{table*}[t]
\centering
\caption{\textbf{Discovery per unseen model on fresh captions.} The confirmation run of \cref{sec:supp_fresh}: 36 disjoint sets of $k=8$ per model against 36 per registered generator, CF features, one run, chance 0.50. AUROC of each model's sets against the registered sets on clean images, on the six new pipelines (New) and on all ten (All). Family: the family used by the bootstrap over families. Relative: the registered generator the model derives from or is a sibling of. Kandinsky 3, OmniGen v1, Stable Cascade and W\"urstchen v2 are the new families. OmniGen v1 is its own base model and decodes with the autoencoder weights of the registered SDXL. SD2.1 is the 512-pixel base and SD2.1-768 its 768-pixel relative. FLUX.1-dev$\times$Krea-dev is our weight merge.}
\label{tab:supp_fresh_models}
\footnotesize
\setlength{\tabcolsep}{3.5pt}
\begin{tabular}{@{}lll ccc@{\hspace{8pt}}ccc@{\hspace{8pt}}ccc@{}}
\toprule
& & & \multicolumn{3}{c}{LineagePrint (LDA)} & \multicolumn{3}{c}{LineagePrint (PLDA)} & \multicolumn{3}{c}{Energy, centered} \\
\cmidrule(lr){4-6}\cmidrule(lr){7-9}\cmidrule(l){10-12}
Model & Family & Relative & Clean & New & All & Clean & New & All & Clean & New & All \\
\midrule
\multicolumn{12}{@{}l}{\textit{Unseen, derived from or a sibling of a registered generator}} \\
Anything-v5 & SD1 & SD1.5 & 0.96 & 0.97 & 0.97 & 0.99 & 0.94 & 0.96 & 0.97 & 0.91 & 0.90 \\
DreamShaper-8 & SD1 & SD1.5 & 0.97 & 0.97 & 0.98 & 1.00 & 0.99 & 1.00 & 1.00 & 1.00 & 1.00 \\
Openjourney & SD1 & SD1.5 & 0.40 & 0.65 & 0.64 & 0.56 & 0.54 & 0.58 & 0.42 & 0.49 & 0.47 \\
RealisticVision-5.1 & SD1 & SD1.5 & 0.90 & 0.98 & 0.97 & 0.96 & 0.97 & 0.97 & 0.88 & 0.98 & 0.96 \\
SD1.4 & SD1 & SD1.5 & 0.71 & 0.54 & 0.61 & 0.22 & 0.25 & 0.28 & 0.07 & 0.14 & 0.14 \\
Animagine-XL-3.1 & SDXL & SDXL & 0.94 & 0.97 & 0.97 & 1.00 & 0.99 & 0.99 & 0.96 & 0.90 & 0.92 \\
DreamShaper-XL & SDXL & SDXL & 1.00 & 0.99 & 0.99 & 0.99 & 0.97 & 0.98 & 0.25 & 0.54 & 0.50 \\
SDXL-Turbo & SDXL & SDXL & 0.99 & 0.98 & 0.99 & 0.99 & 0.98 & 0.99 & 0.50 & 0.82 & 0.76 \\
Playground v2 & Playground & Playground v2.5 & 0.97 & 0.96 & 0.96 & 1.00 & 0.99 & 0.99 & 1.00 & 0.97 & 0.98 \\
AWPortrait-FL & FLUX & FLUX.1-dev & 0.96 & 0.94 & 0.95 & 1.00 & 0.97 & 0.98 & 0.89 & 0.90 & 0.90 \\
Flex.1-alpha & FLUX & FLUX.1-dev & 0.98 & 0.97 & 0.98 & 1.00 & 0.99 & 0.99 & 0.98 & 0.91 & 0.92 \\
FLUX.1-dev + Turbo-Alpha & FLUX & FLUX.1-dev & 0.52 & 0.58 & 0.58 & 0.75 & 0.70 & 0.71 & 0.71 & 0.63 & 0.65 \\
FLUX.1-dev$\times$Krea-dev ($w$=0.50) & FLUX & FLUX.1-dev & 0.63 & 0.77 & 0.76 & 0.90 & 0.90 & 0.90 & 0.84 & 0.81 & 0.81 \\
FLUX.1-Krea-dev & FLUX & FLUX.1-dev & 0.94 & 0.96 & 0.97 & 1.00 & 0.99 & 0.99 & 0.93 & 0.96 & 0.96 \\
FLUX.1-lite-8B & FLUX & FLUX.1-dev & 0.51 & 0.63 & 0.61 & 0.66 & 0.70 & 0.68 & 0.56 & 0.51 & 0.52 \\
FLUX.1-schnell & FLUX & FLUX.1-dev & 0.47 & 0.78 & 0.73 & 0.90 & 0.92 & 0.93 & 0.93 & 0.89 & 0.90 \\
SANA 1.5 & SANA & SANA & 0.37 & 0.65 & 0.64 & 0.83 & 0.88 & 0.89 & 0.90 & 0.86 & 0.87 \\
\midrule
\multicolumn{12}{@{}l}{\textit{Unseen, no registered relative}} \\
SD2.1 & SD2 & -- & 0.32 & 0.59 & 0.59 & 0.38 & 0.47 & 0.50 & 0.34 & 0.41 & 0.40 \\
SD2.1-768 & SD2 & -- & 0.81 & 0.91 & 0.91 & 0.92 & 0.84 & 0.88 & 0.64 & 0.67 & 0.66 \\
SD3.5-large & SD3.5 & -- & 0.97 & 0.98 & 0.98 & 1.00 & 0.99 & 0.99 & 0.99 & 0.99 & 0.99 \\
SD3.5-large-turbo & SD3.5 & -- & 0.94 & 0.97 & 0.97 & 1.00 & 0.99 & 0.99 & 0.98 & 0.96 & 0.96 \\
SD3.5-large-TurboX & SD3.5 & -- & 0.97 & 0.98 & 0.98 & 1.00 & 0.99 & 0.99 & 0.93 & 0.96 & 0.97 \\
SD3.5-medium & SD3.5 & -- & 0.98 & 0.98 & 0.98 & 1.00 & 0.99 & 0.99 & 1.00 & 0.99 & 0.99 \\
SD3.5-medium-turbo & SD3.5 & -- & 0.99 & 0.96 & 0.97 & 1.00 & 0.92 & 0.93 & 0.84 & 0.57 & 0.58 \\
Z-Image-Turbo & Z-Image & -- & 0.97 & 0.95 & 0.96 & 1.00 & 0.98 & 0.99 & 0.92 & 0.90 & 0.91 \\
Kandinsky 3 & Kandinsky & -- & 0.87 & 0.94 & 0.93 & 1.00 & 0.98 & 0.99 & 1.00 & 1.00 & 1.00 \\
OmniGen v1 & OmniGen & -- & 0.57 & 0.64 & 0.67 & 0.96 & 0.81 & 0.86 & 0.96 & 0.82 & 0.85 \\
Stable Cascade & W\"urstchen & -- & 0.96 & 0.92 & 0.93 & 1.00 & 0.96 & 0.97 & 1.00 & 0.98 & 0.98 \\
W\"urstchen v2 & W\"urstchen & -- & 0.92 & 0.88 & 0.90 & 1.00 & 0.96 & 0.97 & 1.00 & 0.99 & 0.99 \\
\midrule
\multicolumn{3}{@{}l}{Mean over the 29 models} & 0.81 & 0.86 & 0.86 & 0.90 & 0.88 & 0.89 & 0.81 & 0.81 & 0.81 \\
\bottomrule
\end{tabular}
\end{table*}
\begin{table}[t]
\centering
\caption{\textbf{False alarms on fresh captions.} Split-conformal threshold for $\alpha=0.05$ from 216 same-pipeline anchor sets of the 12 registered generators (half of their sets of $k=8$), CF features, one run. PLDA and LDA are the two LineagePrint scores, and MSP and Energy read the centered set mean. FPR: false-alarm rate on the other 216 registered sets. TPR: fraction of the sets of the 29 unseen models flagged, mean over the models.}
\label{tab:supp_fresh_calibration}
\small
\setlength{\tabcolsep}{3pt}
\begin{tabular}{@{}lcccccccc@{}}
\toprule
& \multicolumn{2}{c}{PLDA} & \multicolumn{2}{c}{LDA} & \multicolumn{2}{c}{MSP} & \multicolumn{2}{c}{Energy} \\
\cmidrule(lr){2-3}\cmidrule(lr){4-5}\cmidrule(lr){6-7}\cmidrule(l){8-9}
Pipeline & FPR & TPR & FPR & TPR & FPR & TPR & FPR & TPR \\
\midrule
Clean & 0.02 & 0.73 & 0.03 & 0.47 & 0.04 & 0.52 & 0.04 & 0.52 \\
JPEG70 & 0.02 & 0.69 & 0.04 & 0.67 & 0.05 & 0.56 & 0.07 & 0.57 \\
JPEG80 & 0.01 & 0.69 & 0.02 & 0.65 & 0.05 & 0.54 & 0.04 & 0.54 \\
Blur & 0.02 & 0.76 & 0.08 & 0.55 & 0.03 & 0.50 & 0.04 & 0.50 \\
WebP & 0.05 & 0.58 & 0.06 & 0.61 & 0.06 & 0.50 & 0.08 & 0.58 \\
Sharp. & 0.01 & 0.43 & 0.07 & 0.51 & 0.02 & 0.49 & 0.05 & 0.44 \\
Color & 0.02 & 0.72 & 0.02 & 0.62 & 0.04 & 0.53 & 0.05 & 0.58 \\
Noise & 0.03 & 0.60 & 0.04 & 0.48 & 0.03 & 0.44 & 0.05 & 0.47 \\
Median & 0.04 & 0.71 & 0.03 & 0.54 & 0.02 & 0.48 & 0.06 & 0.52 \\
Crop & 0.01 & 0.71 & 0.04 & 0.68 & 0.01 & 0.52 & 0.05 & 0.52 \\
\midrule
Mean & 0.02 & 0.66 & 0.04 & 0.58 & 0.04 & 0.51 & 0.05 & 0.52 \\
\bottomrule
\end{tabular}
\end{table}
\begin{table}[t]
\centering
\caption{\textbf{Unseen models with a registered relative.} The fresh-caption run of \cref{sec:supp_fresh}, CF features, sets of $k=8$, mean over the ten pipelines, 36 sets per model and pipeline. PG: Playground. Flagged: share of sets above the same-pipeline threshold of LineagePrint (LDA) for $\alpha=0.05$. To relative: share of sets whose nearest registered generator is the relative.}
\label{tab:supp_relatives}
\footnotesize
\setlength{\tabcolsep}{2.5pt}
\begin{tabular}{@{}llcc@{}}
\toprule
Unseen model & Relative & Flagged & \shortstack{To\\relative} \\
\midrule
FLUX.1-dev + Turbo-Alpha & FLUX.1-dev & 0.04 & 1.00 \\
FLUX.1-lite-8B & FLUX.1-dev & 0.07 & 0.97 \\
SANA 1.5 1.6B & SANA 1.0 & 0.08 & 0.64 \\
SD1.4 & SD1.5 & 0.12 & 1.00 \\
FLUX.1-schnell & FLUX.1-dev & 0.15 & 0.75 \\
Openjourney & SD1.5 & 0.18 & 0.99 \\
FLUX.1-dev$\times$Krea-dev ($w$=0.50) & FLUX.1-dev & 0.19 & 0.89 \\
AWPortrait-FL & FLUX.1-dev & 0.69 & 0.58 \\
Playground v2 & PG v2.5 & 0.73 & 0.19 \\
FLUX.1-Krea-dev & FLUX.1-dev & 0.78 & 0.17 \\
Animagine-XL-3.1 & SDXL & 0.80 & 0.39 \\
RealisticVision-5.1 & SD1.5 & 0.84 & 0.38 \\
DreamShaper-8 & SD1.5 & 0.86 & 0.04 \\
Anything v5 & SD1.5 & 0.88 & 0.36 \\
Flex.1-alpha & FLUX.1-dev & 0.89 & 0.43 \\
SDXL-Turbo & SDXL & 0.89 & 0.76 \\
DreamShaper-XL & SDXL & 0.94 & 0.95 \\
\bottomrule
\end{tabular}
\end{table}
\begin{table*}[t]
\centering
\caption{\textbf{Open-set recognition methods on the same features and sets.} CF features, the fit, pool, pipelines and disjoint sets of $k=8$ of \cref{sec:supp_fresh}. OpenMax~\cite{bendale2016openmax} on the logits of the linear probe, a Gaussian-mixture density per registered generator (the open-set rule of \citet{nguyen2025forensic} on these features) and a prototype classifier trained on the fit features~\cite{yang2018robust}, each scored per image and averaged over the set, on the set mean, and on the pool-centered set mean. Last column: LineagePrint (PLDA) minus the method over all ten pipelines on the fresh captions.}
\label{tab:supp_openset}
\footnotesize
\setlength{\tabcolsep}{4pt}
\begin{tabular}{@{}ll ccc@{\hspace{10pt}}ccc@{\hspace{10pt}}l@{}}
\toprule
& & \multicolumn{3}{c}{Fresh captions, 29 models} & \multicolumn{3}{c}{Family-Tree captions, 25 models} & \\
\cmidrule(lr){3-5}\cmidrule(lr){6-8}
Method & Input & Clean & New & All & Clean & New & All & PLDA minus method \\
\midrule
OpenMax & per image & 0.68 & 0.57 & 0.60 & 0.66 & 0.57 & 0.60 & +0.287 \\
 & set mean & 0.78 & 0.68 & 0.71 & 0.73 & 0.62 & 0.65 & +0.185 \\
 & set mean, centered & 0.72 & 0.74 & 0.73 & 0.63 & 0.67 & 0.66 & +0.158 \\
\addlinespace[2pt]
Per-generator density & per image & 0.55 & 0.58 & 0.59 & 0.55 & 0.57 & 0.58 & +0.300 \\
 & set mean & 0.83 & 0.61 & 0.68 & 0.84 & 0.63 & 0.69 & +0.215 \\
 & set mean, centered & 0.73 & 0.70 & 0.71 & 0.73 & 0.70 & 0.71 & +0.181 \\
\addlinespace[2pt]
Prototype classifier (trained) & per image & 0.55 & 0.53 & 0.55 & 0.57 & 0.58 & 0.60 & +0.338 \\
 & set mean & 0.86 & 0.69 & 0.77 & 0.85 & 0.70 & 0.77 & +0.123 \\
 & set mean, centered & 0.88 & 0.89 & 0.89 & 0.86 & 0.86 & 0.87 & $-$0.002 \\
\bottomrule
\end{tabular}
\end{table*}
\begin{table}[t]
\centering
\caption{\textbf{Pool, pipeline and set size on the 29 unseen models.} The fresh-caption run of \cref{sec:supp_fresh}, CF features, discovery AUROC over all ten pipelines, sets of $k=8$ unless stated. MSP and Energy score the pool-centered set mean, and the last column is the per-image Energy averaged over the set. Random pools are means over 20 draws from the pool of the test pipeline. Skewed: proportions of the 12 registered generators drawn from a Dirichlet distribution with parameter 0.3. Half unseen: half of the pool replaced by outputs of the unseen models outside the test sets. One generator: mean over the 12 registered generators. Pipeline rows: pool and anchor sets from the named pipeline, mean over the ten test pipelines, with the false-alarm rate on registered sets of the test pipeline (FPR, target 0.05) and the share of unseen sets flagged (TPR).}
\label{tab:supp_poolchannel}
\footnotesize
\setlength{\tabcolsep}{2.2pt}
\begin{tabular}{@{}llccccc@{}}
\toprule
 & & \multicolumn{2}{c}{LineagePrint} & \multicolumn{2}{c}{Set mean} & Image \\
\cmidrule(lr){3-4}\cmidrule(lr){5-6}\cmidrule(l){7-7}
Condition & & LDA & PLDA & MSP & Energy & Energy \\
\midrule
\multicolumn{7}{@{}l}{\textit{Pool (images)}} \\
Full pool &  & 0.86 & 0.89 & 0.82 & 0.81 & 0.71 \\
200 &  & 0.86 & 0.89 & 0.82 & 0.81 & 0.71 \\
100 &  & 0.85 & 0.88 & 0.82 & 0.80 & 0.70 \\
50 &  & 0.84 & 0.88 & 0.81 & 0.80 & 0.70 \\
20 &  & 0.80 & 0.86 & 0.79 & 0.78 & 0.69 \\
10 &  & 0.75 & 0.83 & 0.76 & 0.75 & 0.66 \\
Skewed &  & 0.78 & 0.84 & 0.77 & 0.74 & 0.66 \\
Half unseen &  & 0.84 & 0.89 & 0.85 & 0.85 & 0.76 \\
One generator &  & 0.58 & 0.73 & 0.64 & 0.64 & 0.60 \\
No pool &  & 0.75 & 0.81 & 0.75 & 0.76 & 0.69 \\
\addlinespace[2pt]
\multicolumn{7}{@{}l}{\textit{Pipeline of the pool and anchors}} \\
Test pipeline & AUROC & 0.86 & 0.89 & 0.82 & 0.81 & 0.71 \\
 & FPR & 0.04 & 0.02 & 0.04 & 0.05 & 0.05 \\
 & TPR & 0.58 & 0.66 & 0.51 & 0.52 & 0.27 \\
Clean images & AUROC & 0.75 & 0.81 & 0.72 & 0.71 & 0.63 \\
 & FPR & 0.28 & 0.35 & 0.20 & 0.23 & 0.12 \\
 & TPR & 0.55 & 0.84 & 0.55 & 0.56 & 0.29 \\
Nearest other & AUROC & 0.74 & 0.80 & 0.73 & 0.73 & 0.66 \\
 & FPR & 0.35 & 0.31 & 0.15 & 0.15 & 0.09 \\
 & TPR & 0.72 & 0.79 & 0.51 & 0.50 & 0.26 \\
\addlinespace[2pt]
\multicolumn{7}{@{}l}{\textit{Set size $k$}} \\
1 & & 0.53 & 0.63 & 0.60 & 0.62 & 0.62 \\
2 & & 0.62 & 0.74 & 0.69 & 0.70 & 0.65 \\
4 & & 0.77 & 0.83 & 0.77 & 0.77 & 0.68 \\
8 & & 0.86 & 0.89 & 0.82 & 0.81 & 0.71 \\
16 & & 0.91 & 0.92 & 0.85 & 0.83 & 0.73 \\
\bottomrule
\end{tabular}
\end{table}
\begin{table}[t]
\centering
\caption{\textbf{Mixed sets.} The 29 unseen models on fresh captions (\cref{sec:supp_fresh}), CF features, sets of $k=8$, mean over all ten pipelines, one run. In every test set $m$ images are replaced by images of one other generator from the same pipeline, and a set keeps the label of its original generator. Anchor sets stay pure. AUROC: discovery AUROC. FPR: share of registered-label sets flagged at the threshold for $\alpha=0.05$. Top-1: share of registered-label sets attributed to their label. MSP and Energy score the pool-centered set mean.}
\label{tab:supp_mixed}
\small
\setlength{\tabcolsep}{3pt}
\begin{tabular}{@{}lcccccccc@{}}
\toprule
 & \multicolumn{4}{c}{AUROC} & \multicolumn{2}{c}{FPR} & \multicolumn{2}{c}{Top-1} \\
\cmidrule(lr){2-5}\cmidrule(lr){6-7}\cmidrule(l){8-9}
$m$ of 8 & LDA & PLDA & MSP & Energy & LDA & PLDA & LDA & PLDA \\
\midrule
0 (pure sets) & 0.86 & 0.89 & 0.82 & 0.81 & 0.04 & 0.02 & 1.00 & 1.00 \\
\addlinespace[2pt]
\multicolumn{9}{@{}l}{\textit{Foreign images from one other registered generator}} \\
1 & 0.87 & 0.88 & 0.81 & 0.79 & 0.04 & 0.08 & 0.99 & 0.99 \\
2 & 0.81 & 0.82 & 0.75 & 0.74 & 0.13 & 0.27 & 0.95 & 0.96 \\
4 & 0.49 & 0.50 & 0.44 & 0.53 & 0.50 & 0.81 & 0.47 & 0.47 \\
\addlinespace[2pt]
\multicolumn{9}{@{}l}{\textit{Foreign images from one other unseen model}} \\
1 & 0.87 & 0.88 & 0.81 & 0.79 & 0.05 & 0.07 & 0.99 & 0.99 \\
2 & 0.84 & 0.85 & 0.78 & 0.76 & 0.11 & 0.25 & 0.97 & 0.97 \\
4 & 0.63 & 0.63 & 0.59 & 0.61 & 0.45 & 0.74 & 0.69 & 0.67 \\
\bottomrule
\end{tabular}
\end{table}
 
\noindent\textbf{Setting.}
The fresh block holds the first 400 COCO-dataset captions (88 COCO images) whose COCO image is not behind a Family-Tree caption, a fitting image or the images used to develop the method.
The 25 models of the Family-Tree test render them with their earlier settings and new seeds, joined by Kandinsky 3, OmniGen v1, Stable Cascade and W\"urstchen v2 from three families that are neither registered nor among the eight families of the 25 models, and the registered generators keep their COCO-dataset images of these captions.
Fit, pool, pipelines and set rules are those of the Family-Tree test (\cref{sec:supp_splits}), with 36 disjoint sets of 8 per generator. Before the run we fixed one comparison, LineagePrint (LDA) against Energy on the centered set mean over all ten pipelines.

\noindent\textbf{Results.}
The fixed comparison is level: LineagePrint (LDA) reaches 0.86 as on the Family-Tree captions, and Energy on the centered set mean reaches 0.81 (\cref{tab:supp_fresh_scores}).
LineagePrint (PLDA) reaches 0.89 on the fresh captions and 0.86 on the Family-Tree captions, against 0.81 and 0.74 for Energy and 0.82 and 0.77 for MSP on the centered set mean.
LineagePrint (LDA) leads the best per-image score on both blocks and is level with MSP on the centered set mean.
For the 17 models with a registered relative LineagePrint (PLDA) reaches 0.87 against 0.77 for Energy on the centered set mean, and for the 12 models without one 0.92 against 0.86.
On the four models of the three new families, Energy on the centered set mean (0.96) leads LineagePrint (LDA) (0.86) and is level with LineagePrint (PLDA) (0.95).
Energy gives 0.50 or less on SD1.4 (0.14), SD2.1 (0.40), Openjourney (0.48) and DreamShaper-XL (0.50), where LineagePrint (LDA) gives 0.61, 0.59, 0.64 and 0.99 (\cref{tab:supp_fresh_models}).
At the same-pipeline threshold for $\alpha=0.05$, the FPR of LineagePrint (LDA) ranges from 0.02 to 0.08 over the ten pipelines (mean 0.04) and exceeds 0.05 under WebP, sharpening and blur, and it flags 0.58 of the unseen-model sets on average (\cref{tab:supp_fresh_calibration}).
LineagePrint (PLDA) keeps the FPR between 0.01 and 0.05 (mean 0.02) and flags 0.66 of the unseen-model sets, against 0.51 for MSP and 0.52 for Energy on the centered set mean.
\Cref{tab:supp_openset} adds three open-set recognition methods on the same features and sets. LineagePrint (PLDA) leads OpenMax and the per-generator density in every form (0.73 and 0.71 on the centered set mean). The prototype classifier, which trains an embedding and one prototype per registered generator, reaches 0.89 on the centered set mean and is level with LineagePrint (PLDA), and it reaches 0.77 without centering. Like the learned heads of \cref{tab:supp_learned_head}, a trained head with the set fingerprint and its centering matches the closed form.
\Cref{tab:supp_relatives} lists, for the 17 unseen models with a registered relative, the share of sets flagged as new and the share whose nearest registered generator is the relative. Seven models are flagged in at most 0.19 of their sets and attributed to the relative in 0.64 to 1.00, and the other ten are flagged in 0.69 to 0.94.

\noindent\textbf{Pool, pipeline and set size.}
\Cref{tab:supp_poolchannel} changes the pool, the pipeline of the pool and the anchor sets, and the set size on this test, with the scores recomputed from the stored features.
LineagePrint (PLDA) reaches 0.88 with 50 pool images and 0.83 with 10, against 0.89 with the full pool, and it leads MSP and Energy on the centered set mean at every pool size.
With half of the pool replaced by outputs of the unseen models it stays at 0.89, and MSP and Energy on the centered set mean rise to 0.85.
Without a pool it reaches 0.81 against 0.76 for the best baseline.
A pool with skewed proportions of the registered generators gives 0.84, and a pool from one registered generator gives 0.73 on average (0.62 to 0.84 over the 12 generators), against 0.64 for the baselines.
LineagePrint (LDA) falls below the best baseline without a pool (0.75 against 0.76), with a pool from one generator (0.58) and with half of the pool unseen (0.84).
With pool and anchor sets from clean images or from the nearest other pipeline, LineagePrint (PLDA) keeps the highest AUROC (0.81 and 0.80), and the false-alarm rate rises to 0.35 and 0.31 for the 0.05 target, so the threshold needs anchors from the test pipeline.
The pipeline can be read from the queries: for each of the ten pipelines, the pool whose mean is nearest to the mean of all query sets of that pipeline is the pool of that pipeline.
LineagePrint (PLDA) leads every baseline at every set size, with 0.63 for single images, 0.83 for sets of 4 and 0.92 for sets of 16, and LineagePrint (LDA) leads from $k=8$ on.
\noindent\textbf{Mixed sets.}
\Cref{tab:supp_mixed} replaces $m$ of the 8 images of every test set by images of one other generator, a registered one or an unseen one, and keeps the anchor sets pure.
With one foreign image LineagePrint (PLDA) reaches a discovery AUROC of 0.88, against 0.89 for pure sets, and registered sets are attributed to their own generator in 0.99 of the cases.
With two foreign images it reaches 0.82 to 0.85 and stays ahead of MSP and Energy on the centered set mean (0.74 to 0.78).
The false-alarm rate rises with the share of foreign images, since a mixed set matches no single registered generator: for PLDA from 0.02 to 0.07 and 0.08 with one foreign image and to 0.25 and 0.27 with two, and for LDA from 0.04 to 0.04 and 0.05 and to 0.11 and 0.13.
With half of the images foreign every score is close to chance.

\noindent\textbf{Second block of captions, comparison fixed in advance.}
\label{sec:supp_confirm}
After the test above we fixed one rule before any further image existed. On a second block of 400 COCO-dataset captions (83 COCO images) whose COCO images lie behind no caption rendered before, LineagePrint (PLDA) leads only if the 95\% family-level intervals of its paired AUROC difference to MSP and to Energy on the centered set mean both exclude zero.
The 29 unseen models render the block with their earlier settings. The registered generators are rendered again as well: an admission rule that uses no part of LineagePrint admits nine of the ten open ones (Kolors is not admitted), and the two API generators contribute no sets.
Fit, gallery, pipelines and set rules are unchanged.
LineagePrint (PLDA) reaches 0.86 over the ten pipelines, against 0.76 for MSP (difference $+0.09$, interval $[+0.04, +0.14]$, ahead on 26 of the 29 models) and 0.73 for Energy ($+0.13$, $[+0.06, +0.22]$, ahead on 24).
It also leads on clean images (0.87 against 0.76 and 0.71) and on the six new pipelines (0.85 against 0.76 and 0.73).
The best other score on the centered set mean is Mahalanobis with 0.78.
At the same-pipeline threshold for $\alpha=0.05$ its FPR is 0.03 on average and it flags 0.58 of the unseen-model sets, against 0.38 for MSP and 0.34 for Energy.
LineagePrint (LDA) reaches 0.83 and is level with both baselines, and on the four models of the three new families Energy on the centered set mean (0.96) leads LineagePrint (PLDA) (0.93).

All numbers of this section come from one run per caption block.

\clearpage  %
\section{Speech}
\label{sec:supp_speech}

\begin{table*}[t]
\centering
\caption{\textbf{Discovery of unseen speech generators across codecs.} ASVspoof 2021 LA: sets from 11 unseen speech generators (text-to-speech and voice-conversion systems) against sets from the 2 registered systems that reappear in 2021, per channel (uncompressed and six telephony channels) and averaged. The subspace is fit on the six ASVspoof 2019 LA training systems with AASIST features. Sets of $k=16$ utterances, mean over 3 speaker-disjoint splits, chance 0.50. Attr.: closed-set attribution accuracy of registered-system sets among the six registered systems (chance 0.17). Best in bold, second best underlined. LineagePrint (PLDA) has the best mean AUROC and attributes every registered set correctly.}
\label{tab:speech}
\small
\begin{tabular}{@{}lcccccccc@{\hspace{10pt}}c@{}}
\toprule
& \multicolumn{8}{c}{Discovery AUROC $\uparrow$} & \\
\cmidrule(lr){2-9}
Method & None & A-law & $\mu$-law & G.722 & GSM & Opus & PSTN & Mean & Attr. $\uparrow$ \\
\midrule
MSP & 0.43 & 0.52 & 0.56 & 0.51 & 0.82 & 0.51 & 0.73 & 0.58 & 0.69 \\
Energy & 0.30 & 0.42 & 0.44 & 0.37 & 0.53 & 0.42 & 0.46 & 0.42 & 0.69 \\
Mahalanobis & 0.80 & \underline{0.78} & 0.79 & 0.69 & \textbf{0.86} & 0.74 & \underline{0.82} & 0.78 & \underline{0.94} \\
kNN & 0.47 & 0.42 & 0.51 & 0.42 & 0.54 & 0.48 & 0.48 & 0.47 & 0.75 \\
\midrule
LineagePrint (LDA) & \underline{0.83} & \textbf{0.79} & \textbf{0.83} & \underline{0.75} & \underline{0.83} & \underline{0.80} & \textbf{0.87} & \underline{0.81} & \underline{0.94} \\
LineagePrint (PLDA) & \textbf{0.88} & \underline{0.78} & \underline{0.82} & \textbf{0.84} & 0.78 & \textbf{0.87} & 0.81 & \textbf{0.83} & \textbf{1.00} \\
\bottomrule
\end{tabular}
\end{table*}
\begin{table}[t]
\centering
\caption{\textbf{False alarms per speech channel.} ASVspoof 2021 LA, AASIST features, sets of $k=8$ utterances, $\alpha=0.05$, mean over 3 splits. Source: threshold calibrated on ASVspoof 2019 LA. Same channel: threshold from 20 anchor sets of registered systems in the test channel. Same-channel anchors keep the mean false-alarm rate close to $\alpha$, while source thresholds exceed $\alpha$ in every channel. At the source thresholds the mean TPR is 0.71 (LDA) and 0.69 (PLDA).}
\label{tab:supp_codec_calibration}
\small
\setlength{\tabcolsep}{3.5pt}
\begin{tabular}{@{}lcccccc@{}}
\toprule
& \multicolumn{4}{c}{FPR ($\alpha=0.05$)} & \multicolumn{2}{c}{TPR $\uparrow$} \\
\cmidrule(lr){2-5}\cmidrule(l){6-7}
& \multicolumn{2}{c}{Source} & \multicolumn{2}{c}{Same channel} & \multicolumn{2}{c}{Same channel} \\
\cmidrule(lr){2-3}\cmidrule(lr){4-5}\cmidrule(l){6-7}
Channel & LDA & PLDA & LDA & PLDA & LDA & PLDA \\
\midrule
None & 0.11 & 0.16 & 0.05 & 0.07 & 0.06 & 0.48 \\
A-law & 0.29 & 0.30 & 0.12 & 0.02 & 0.28 & 0.24 \\
$\mu$-law & 0.15 & 0.29 & 0.03 & 0.12 & 0.04 & 0.47 \\
G.722 & 0.35 & 0.29 & 0.11 & 0.00 & 0.20 & 0.16 \\
GSM & 0.64 & 0.90 & 0.08 & 0.01 & 0.28 & 0.29 \\
Opus & 0.23 & 0.11 & 0.08 & 0.00 & 0.25 & 0.21 \\
PSTN & 0.45 & 0.53 & 0.02 & 0.01 & 0.19 & 0.38 \\
\midrule
Mean & 0.32 & 0.37 & 0.07 & 0.03 & 0.18 & 0.32 \\
\bottomrule
\end{tabular}
\end{table}
 
The same closed-form fingerprint applies to speech with a speech encoder.

\noindent\textbf{Related work.}
Source tracing names the text-to-speech or voice-conversion system behind an utterance.
Prior work classifies generators and their components~\cite{klein2024sourcetracing,falez2025multiattribute,yi2023add}, scores unseen generators~\cite{klein2025openset,xie2024generalizedsourcetracing,firc2025stopa}, learns metric embeddings~\cite{neamtu2026anchoring} or splits a generator into architecture and training data~\cite{rubio2026compositional}.
\citet{chhibber2026zeroshot} average the utterance embeddings of one generator into an enrolled fingerprint, and source verification compares an utterance with references of one generator~\cite{negroni2025source}.
Source tracing also covers new languages~\cite{xuan2025multilingualsourcetracing,muller2024mlaad} and codec-based generators~\cite{chen2025codecsourcetracing,xie2025neuralcodecsourcetracing}. One codec pass can lower closed-set attribution by up to 70 Macro-F1 points~\cite{shi2026codecstress}.
These methods train speech embeddings, while our closed-form fingerprint on frozen features serves images and speech alike.

\noindent\textbf{Setup.}
ASVspoof 2019 LA~\cite{wang2020asvspoof2019} holds 19 speech generators, 6 for training and 13 for evaluation (two repeat training generators), and ASVspoof 2021 LA~\cite{liu2023asvspoof2021} sends the 13 through six telephony channels and the uncompressed one, the seven pipelines of this test (\cref{sec:supp_asvspoof}).
AASIST features come from a frozen anti-spoofing network~\cite{tak2022wav2vec} (XLS-R~\cite{babu2022xlsr} wav2vec 2.0~\cite{baevski2020wav2vec2} front end, AASIST back end~\cite{jung2022aasist}) trained on ASVspoof 2019 LA.
We fit on the six ASVspoof 2019 LA training generators and test on 2021 LA, where 11 generators are unseen and 2 registered, with speaker-disjoint splits and sets of $k=16$ utterances.
The registered generators give the shift that replaces the pool mean, for every method (\cref{sec:supp_splits}).
Calibration uses $N=20$ anchor sets per channel.

\noindent\textbf{Results.}
With AASIST features LineagePrint (PLDA) flags unseen generators with a mean AUROC of 0.83 over the seven channels (0.78 to 0.88 per channel), against 0.78 for the best baseline (Mahalanobis, \cref{tab:speech}), and attributes every registered set correctly (top-1 1.00).
It leads at every set size (\cref{fig:discovery}c), with 0.78 against 0.72 at $k=8$.
The Energy score reaches 0.42.
With wav2vec 2.0 base features every score reaches 0.92 to 0.98 on this task (LineagePrint (PLDA) 0.97, Mahalanobis 0.98, Energy 0.97 at $k=16$). With the AASIST features of the anti-spoofing network, the scores separate, and LineagePrint leads.

\noindent\textbf{False alarms per speech channel.}
\Cref{tab:supp_codec_calibration} gives the false-alarm rate in each channel at $\alpha=0.05$.
Thresholds calibrated on ASVspoof 2019 LA exceed $\alpha$ in every channel, give a mean FPR of 0.32 (LDA) and 0.37 (PLDA) and reach 0.64 (LDA) and 0.90 (PLDA) under GSM.
Same-channel anchors bring the mean false-alarm rate to 0.07 (LDA) and 0.03 (PLDA).
At the same-channel threshold LineagePrint (PLDA) flags 0.32 of unseen-system sets of 8 utterances at an FPR of 0.03.
The speech evidence covers one corpus (ASVspoof 2019 and 2021 LA).

\section{What moves a fingerprint}
\label{sec:supp_moves}

The fingerprint records how a model was trained.
This section measures how far training moves it, in a controlled fine-tuning experiment (\cref{sec:supp_finetune}), in public style fine-tunes (\cref{sec:supp_style}), and how well the fingerprint tells apart community fine-tunes of the same base (\cref{sec:supp_gallery}).

\subsection{Controlled fine-tuning}
\label{sec:supp_finetune}

\begin{table*}[t]
\centering
\caption{\textbf{Fingerprint distance of LoRA fine-tunes.} One LoRA run per base model and fine-tuning set, CF features, label-free fingerprint distance (\cref{sec:supp_tree_metric}). In this distance the Family-Tree set has mean 2.78 within a family and 6.06 across families, and SD1.5 and SDXL lie 6.76 apart. The fingerprint of each checkpoint is the mean over its $n=400$ images, one per Family-Tree caption, rendered with the seeds and sampler settings of its base. True base: distance to the base model the run starts from, after 250, 1,000 and 4,000 steps. Other base: distance to the other base model (SDXL for SD1.5 runs, SD1.5 for SDXL runs) after 4,000 steps. Nearest base: the closest of the four registered bases (SD1.5, SD2.1, SDXL, Playground v2) after 4,000 steps. Three of the four fine-tunes stay nearest to their base after 4,000 steps, and SDXL fine-tuned on photographs moves as far as a new base model.}
\label{tab:supp_finetune}
\small
\setlength{\tabcolsep}{6pt}
\begin{tabular}{@{}llccccl@{}}
\toprule
& & \multicolumn{3}{c}{True base} & Other base & Nearest base \\
\cmidrule(lr){3-5}\cmidrule(lr){6-6}\cmidrule(l){7-7}
Base model & Fine-tuning set & 250 & 1,000 & 4,000 & 4,000 & 4,000 \\
\midrule
SD1.5 & Photographs & 1.99 & 1.92 & 2.15 & 7.49 & SD1.5 \\
SD1.5 & Anime & 0.70 & 0.97 & 1.45 & 7.00 & SD1.5 \\
SDXL & Photographs & 5.81 & 5.48 & 6.82 & 6.77 & SD2.1 \\
SDXL & Anime & 4.82 & 4.99 & 5.27 & 7.40 & SDXL \\
\bottomrule
\end{tabular}
\end{table*}
 
Fine-tuning an open text-to-image model on a small targeted set is a common way to change what it renders, for example object states~\cite{chen2025objectstate}.
Does fine-tuning move a fingerprint away from its base model, and does the base model or the training style decide where it goes?

\noindent\textbf{Setup.}
We fine-tune SD1.5 (512 pixels) and SDXL (1,024 pixels) with low-rank adaptation (LoRA)~\cite{hu2022lora} of rank 64 and scale 64 on the attention projections of the denoising U-Net. LoRA is one of several ways to fine-tune a pretrained model by changing few parameters, and tuning only the bias terms and normalization layers is another~\cite{li2024clipfit}.
Training uses AdamW with learning rate $10^{-4}$ after a 100-step linear warm-up, half-precision base weights and full-precision LoRA weights.
SD1.5 uses batch size 4, and SDXL batch size 2 with two-step gradient accumulation.
Each image is resized so that its short side matches the training resolution, randomly cropped and randomly flipped.
Each base is fine-tuned on two sets: the 1,200 Flickr30k photographs of the Flickr dataset with their captions (photographs), and 1,221 illustrations from the Naruto series with BLIP~\cite{li2022blip} captions (anime, \cite{cervenka2022naruto}).
Neither set shares a caption with the evaluation captions.
Each base and set gives one training run.
Checkpoints after 250, 1,000 and 4,000 steps render the 400 Family-Tree captions with the seeds and sampler settings of their base (50 steps and guidance 7.5 for SD1.5, 30 steps and guidance 7.0 for SDXL).
The fingerprint of a checkpoint is the mean projected feature (CF or CLIP ViT-L/14) of its $n=400$ images, and distances use the label-free distance of \cref{sec:supp_tree_metric}.
CF features come from a forensic detector, and we call them forensic features. CLIP features come from image-text training, and we call them semantic features.
In this distance SD1.5 and SDXL lie 6.76 apart, and the Family-Tree set has a mean distance of 2.78 within a family and 6.06 across families.
The registered bases are SD1.5, SD2.1, SDXL and Playground v2.

\noindent\textbf{The base model stays nearest.}
The nearest registered base is the true base at 10 of 12 checkpoints (\cref{tab:supp_finetune}).
All three checkpoints of SD1.5 on photographs, SD1.5 on anime and SDXL on anime stay nearest to their base, and one of three for SDXL on photographs.
In all four runs the distance to the base is larger after 4,000 steps than after 250 steps, for example from 0.70 to 1.45 for SD1.5 on anime and from 4.82 to 5.27 for SDXL on anime.

\noindent\textbf{Forensic features follow the base, semantic features follow the style.}
We split the spread of the four fingerprints after 4,000 steps into a base part, along the axis from the mean of the SD1.5 runs to the mean of the SDXL runs, a style part, along the axis from the mean of the photograph runs to the mean of the anime runs, and the remainder.
With CF features the base model explains 0.61 of this variance and the style 0.24, with 0.16 left over.
With CLIP ViT-L/14 features the style explains 0.57 and the base 0.24, with 0.19 left over.
With CF features, two runs with the same style and different bases lie 7.34 apart (anime) and 6.55 apart (photographs), while the two SD1.5 runs with different styles lie 2.21 apart.
The two SDXL runs lie 6.70 apart, and both move far from SDXL after 4,000 steps (5.27 on anime, 6.82 on photographs).

\noindent\textbf{A fine-tune can move as far as a new base model.}
SDXL fine-tuned on photographs moves furthest.
After 4,000 steps it lies 6.82 from SDXL, 6.77 from SD1.5, 6.74 from SD2.1 and 8.00 from Playground v2.
It is as far from its own base as SD1.5 is from SDXL (6.76), so in fingerprint space it behaves like a new base model.

\noindent\textbf{Sampler settings keep the base.}
The same SDXL weights, captions and seeds sampled with 50 steps and guidance 5.0, instead of 30 steps and guidance 7.0, lie 2.61 from the original SDXL fingerprint with CF features (0.31 with CLIP ViT-L/14).
This set stays nearest to SDXL, with Playground v2 at 4.47 and SD1.5 at 6.89.
A change of sampler settings keeps the set attributed to its base.

The fingerprint follows the weights and the training data.
Forensic features keep the base model through LoRA fine-tuning in three of four runs, and a fine-tune can move as far as a new base model, so such a descendant is discovered as new.

\subsection{Heavy style fine-tunes}
\label{sec:supp_style}

\begin{table*}[t]
\centering
\caption{\textbf{Where the Family-Tree descendants go.} Family-Tree set, setting of \cref{tab:tree}: sets of $k=16$ images, 72 sets per model, a gallery of the four base models SD1.5, SD2.1, SDXL and Playground v2 (PG v2). DS-8 is DreamShaper-8, RV-5.1 is RealisticVision-5.1 and Anim-XL is Animagine-XL-3.1. Each count is the number of sets whose nearest gallery fingerprint is this base, with CF or CLIP ViT-L/14 features. The most frequent base is in bold. Base: the registered base of the model's family. Nearest model: the closest of the other 17 models by fingerprint distance (CF). With CF features every SD1/SD2 descendant goes to an SD1 or SD2 base (720 of 720 sets), the heavy style fine-tunes of SD1.5 and their merge go to SD2.1 (359 of 360 sets), and Animagine-XL-3.1 goes to Playground v2 (72 of 72 sets).}
\label{tab:supp_style}
\small
\setlength{\tabcolsep}{4pt}
\begin{tabular}{@{}llcccc@{\hspace{10pt}}cccc@{\hspace{10pt}}l@{}}
\toprule
& & \multicolumn{4}{c}{CF ViT-S/16} & \multicolumn{4}{c}{CLIP ViT-L/14} & \\
\cmidrule(lr){3-6}\cmidrule(lr){7-10}
Model & Base & SD1.5 & SD2.1 & SDXL & PG v2 & SD1.5 & SD2.1 & SDXL & PG v2 & Nearest model \\
\midrule
\multicolumn{11}{@{}l}{\textit{Sibling version}} \\
SD1.4 & SD1.5 & \textbf{72} & 0 & 0 & 0 & \textbf{71} & 1 & 0 & 0 & SD1.5 \\
\addlinespace[2pt]
\multicolumn{11}{@{}l}{\textit{768-pixel relative}} \\
SD2.1-768 & SD2.1 & 0 & \textbf{72} & 0 & 0 & 10 & \textbf{62} & 0 & 0 & SD2.1 \\
\addlinespace[2pt]
\multicolumn{11}{@{}l}{\textit{Distillation}} \\
SDXL-Turbo & SDXL & 0 & 0 & \textbf{72} & 0 & 0 & 0 & \textbf{72} & 0 & SDXL \\
\addlinespace[2pt]
\multicolumn{11}{@{}l}{\textit{Light fine-tune}} \\
DreamShaper-XL & SDXL & 0 & 0 & \textbf{72} & 0 & 0 & 0 & \textbf{72} & 0 & SDXL \\
\addlinespace[2pt]
\multicolumn{11}{@{}l}{\textit{Weight merge with a base}} \\
SD1.5$\times$DS-8 (0.25) & SD1.5 & \textbf{72} & 0 & 0 & 0 & \textbf{65} & 7 & 0 & 0 & SD1.5$\times$DS-8 (0.50) \\
SD1.5$\times$DS-8 (0.50) & SD1.5 & \textbf{70} & 2 & 0 & 0 & \textbf{50} & 22 & 0 & 0 & SD1.5$\times$DS-8 (0.25) \\
SD1.5$\times$DS-8 (0.75) & SD1.5 & \textbf{50} & 22 & 0 & 0 & 32 & \textbf{39} & 1 & 0 & SD1.5$\times$DS-8 (0.50) \\
SDXL$\times$Anim-XL (0.50) & SDXL & 0 & 0 & \textbf{71} & 1 & 0 & 0 & \textbf{72} & 0 & SDXL \\
\addlinespace[2pt]
\multicolumn{11}{@{}l}{\textit{Heavy style fine-tune}} \\
Anything-v5 & SD1.5 & 1 & \textbf{71} & 0 & 0 & 1 & 1 & \textbf{69} & 1 & SD2.1-768 \\
DreamShaper-8 & SD1.5 & 0 & \textbf{72} & 0 & 0 & 0 & \textbf{39} & 32 & 1 & DS-8$\times$RV-5.1 (0.50) \\
Openjourney & SD1.5 & 0 & \textbf{72} & 0 & 0 & 17 & \textbf{55} & 0 & 0 & SD2.1 \\
RealisticVision-5.1 & SD1.5 & 0 & \textbf{72} & 0 & 0 & 6 & 28 & \textbf{38} & 0 & DS-8$\times$RV-5.1 (0.50) \\
Animagine-XL-3.1 & SDXL & 0 & 0 & 0 & \textbf{72} & 0 & 0 & \textbf{58} & 14 & Playground v2 \\
\addlinespace[2pt]
\multicolumn{11}{@{}l}{\textit{Weight merge of two style fine-tunes}} \\
DS-8$\times$RV-5.1 (0.50) & SD1.5 & 0 & \textbf{72} & 0 & 0 & 2 & \textbf{60} & 10 & 0 & DreamShaper-8 \\
\bottomrule
\end{tabular}
\end{table*}
 
The five heavy style fine-tunes of the Family-Tree set move away from their base (\cref{tab:supp_style}).
Their sets go to their true base in 1 of 360 cases (CF, $k=16$).
Every set of every SD1 or SD2 descendant goes to an SD1 or SD2 base (720 of 720).
The four SD1.5 style fine-tunes and their merge go to SD2.1 in 359 of 360 sets, and DreamShaper-8 and RealisticVision-5.1 each lie nearest to their merge.
Animagine-XL-3.1 sends all 72 of its sets to Playground v2, which is also its nearest model.
With CLIP ViT-L/14 features 568 of 720 SD1 and SD2 descendant sets go to an SD1 or SD2 base, and Animagine-XL-3.1 sends 58 of 72 sets to SDXL.

Style also groups models across families when the features are semantic.
In the label-free distance with CLIP ViT-L/14 features, the anime fine-tunes Anything-v5 (on SD1.5) and Animagine-XL-3.1 (on SDXL) lie 2.00 apart, closer than to their own bases (2.97 and 2.92).
With CF features Anything-v5 stays near SD1.5 (3.24, against 6.42 to Animagine-XL-3.1), and Animagine-XL-3.1 lies 6.50 from its base SDXL, farther than the mean distance across families in this metric (6.06).
Semantic features group the two anime models by style, and forensic features do not.

\subsection{Generator-level discovery among community fine-tunes}
\label{sec:supp_gallery}

\begin{table*}[t]
\centering
\caption{\textbf{Generator-level discovery among community fine-tunes.} CF-Small (1,951 community generators): $M$ registered generators, every other generator is unseen. Discovery AUROC of single unseen generators, CF features, sets of $k=8$, mean over 5 draws, chance 0.50. Best in bold, second best underlined. Telling one community fine-tune from another becomes harder as the gallery grows, for every method. LineagePrint (PLDA) scores highest from $M=16$ on. Family-level discovery (\cref{tab:lineage}) is the task the fingerprint solves at this scale.}
\label{tab:supp_gallery}
\small
\setlength{\tabcolsep}{8pt}
\begin{tabular}{@{}lcccccccc@{}}
\toprule
& \multicolumn{8}{c}{Registered generators $M$} \\
\cmidrule(l){2-9}
Method & 8 & 16 & 32 & 64 & 128 & 256 & 512 & 1,024 \\
\midrule
MSP & 0.66 & 0.60 & 0.57 & 0.55 & 0.54 & \underline{0.54} & 0.52 & 0.51 \\
Energy & 0.65 & 0.58 & 0.57 & 0.54 & 0.54 & 0.53 & 0.51 & 0.50 \\
Mahalanobis & \textbf{0.72} & 0.62 & 0.57 & 0.55 & 0.53 & 0.51 & 0.52 & 0.51 \\
kNN & 0.59 & 0.54 & 0.55 & 0.54 & 0.52 & 0.51 & 0.51 & 0.51 \\
\midrule
LineagePrint (LDA) & \underline{0.71} & \underline{0.67} & \underline{0.61} & \underline{0.59} & \underline{0.56} & 0.53 & \underline{0.53} & \underline{0.53} \\
LineagePrint (PLDA) & 0.69 & \textbf{0.70} & \textbf{0.68} & \textbf{0.65} & \textbf{0.63} & \textbf{0.59} & \textbf{0.57} & \textbf{0.57} \\
\bottomrule
\end{tabular}
\end{table*}
 
At community scale the family is the unit that the fingerprint records.
CF-Small registers $M$ generators and treats every other generator as unseen, so queries and registered generators are latent diffusion models, many of them fine-tunes of the same base models.
\Cref{tab:supp_gallery} reports discovery of single unseen generators for $M=8$ to 1,024 (CF, $k=8$, 5 draws).
Fine-tunes of one base lie close together, and telling one from another becomes harder as the gallery grows.
LineagePrint (PLDA) scores highest from $M=16$ on, with 0.70 at $M=16$ and 0.57 at $M=1{,}024$, where the baselines score 0.50 to 0.51.
Across families the set mean separates generators well: a gallery of SD1/SD2 fine-tunes flags unseen SDXL fine-tunes with AUROC 0.87 and 0.90 for LineagePrint (LDA) and 0.91 and 0.95 for the raw set-mean centroid at $k=8$ and 16 (\cref{tab:lineage}).

\section{Stable Diffusion family tree}
\label{sec:supp_tree}

This section adds to \cref{sec:exp_tree}.
The heavy style fine-tunes of SD1.5 and their merge have SD2.1 as their nearest base (359 of 360 sets), which the decoder explains for Openjourney (\cref{sec:exp_controls}), and Animagine-XL-3.1 moves to Playground v2 (72 of 72, \cref{tab:supp_style}).
Playground v2~\cite{playground2023v2}, not the Playground v2.5~\cite{li2024playground} of the two datasets, decodes with the SDXL autoencoder and samples at 1,024 pixels like SDXL. For this model the fingerprint follows training rather than architecture or decoder, unlike the architecture traces that survive retraining in GANs~\cite{yang2022dnadet}.
The permutation test of the correlation between fingerprint distance and tree distance uses 20,000 label permutations.

\clearpage  %
\section{FLUX.1 and SD3.5 families}
\label{sec:supp_modern}
\label{sec:exp_modern}

\begin{table}[t]
\centering
\caption{\textbf{FLUX.1 and SD3.5 families.} Models sampled at 1,024 pixels, CF features, sets of $k=16$, 72 sets per model, 17 registered bases. To base: share of sets whose nearest registered fingerprint is the base of the model's family. $d_\mathrm{base}$, $d_\mathrm{other}$: fingerprint distance to this base and to the nearest other base, named. Family rows: mean distance between members and from members to models of other families. $^\dagger$Rule fixed before the run: share at least 0.80. L: large, M: medium, PG: Playground, dist.: distilled, $w$: merge weight. Relations from the model cards. Variant: the card lists FLUX.1-dev as base model. All 45 models in \cref{tab:supp_modern_all,tab:supp_modern_models}.}
\label{tab:modern}
\footnotesize
\setlength{\tabcolsep}{1.3pt}
\begin{tabular}{@{}llcccl@{}}
\toprule
Model & Relation & \shortstack{To\\base} & $d_\mathrm{base}$ & $d_\mathrm{other}$ & Other base \\
\midrule
\multicolumn{6}{@{}l}{\textit{FLUX.1 family, base FLUX.1-dev}} \\
FLUX.1-lite-8B$^\dagger$ & distilled & 0.96 & 1.47 & 3.38 & CogView4 \\
dev + Turbo-Alpha$^\dagger$ & LoRA & 0.90 & 2.01 & 3.23 & Z-Image \\
FLUX.1-schnell & sibling & 0.96 & 1.76 & 3.48 & SD3.5-L \\
dev$\times$Krea ($w$=0.50) & merge & 0.99 & 1.91 & 3.97 & Lumina 2.0 \\
FLUX.1-Krea-dev & variant & 0.00 & 4.73 & 3.07 & Gemini 2.5 \\
AWPortrait-FL & fine-tune & 0.04 & 4.11 & 3.28 & Gemini 2.5 \\
Flex.1-alpha & from schnell & 0.00 & 5.27 & 2.78 & Z-Image \\
FLUX.1-dev, 20 steps & sampler & 0.97 & 1.02 & 3.47 & CogView4 \\
\textit{Mean, family} & & & 3.61 & 7.93 & \\
\addlinespace[2pt]
\multicolumn{6}{@{}l}{\textit{SD3.5 family, base SD3.5-L}} \\
SD3.5-L-TurboX$^\dagger$ & few-step & 0.99 & 2.63 & 4.09 & Gemini 2.5 \\
SD3.5-M & sibling & 0.96 & 2.62 & 4.25 & Z-Image \\
SD3.5-M-turbo & sibling, dist. & 0.43 & 4.94 & 4.82 & PG v2.5 \\
SD3.5-L-turbo & distilled & 0.01 & 5.19 & 4.02 & Lumina 2.0 \\
\textit{Mean, family} & & & 4.60 & 7.42 & \\
\bottomrule
\end{tabular}
\end{table}
\begin{table}[t]
\centering
\caption{\textbf{Family verification.} Every set of the 25 relatives (sets of $k=16$, 72 per model) is scored against each of the 17 registered bases by the fingerprint distance of \cref{sec:supp_tree_metric}, with the base fingerprint taken from the other caption half. A trial is a target trial when the base is the documented base of the model's family. Per family: target trials of its relatives against the trials of every other model with that base. The three re-renderings of registered checkpoints are left out. $n$: relatives. PG: Playground. CLIP-L: CLIP ViT-L/14 features. AUROC, and equal error rate (EER).}
\label{tab:supp_verification}
\footnotesize
\setlength{\tabcolsep}{2.5pt}
\begin{tabular}{@{}llccccc@{}}
\toprule
& & & \multicolumn{2}{c}{CF} & \multicolumn{2}{c}{CLIP-L} \\
\cmidrule(lr){4-5}\cmidrule(l){6-7}
Family & Base & $n$ & AUROC & EER & AUROC & EER \\
\midrule
SD1 & SD1.5 & 9 & 0.98 & 0.06 & 0.78 & 0.26 \\
SD2 & SD2.1 & 1 & 0.74 & 0.29 & 0.92 & 0.15 \\
SDXL & SDXL & 4 & 0.88 & 0.21 & 0.61 & 0.38 \\
FLUX.1 & FLUX.1-dev & 7 & 0.97 & 0.10 & 0.94 & 0.13 \\
SD3.5 & SD3.5-large & 4 & 0.84 & 0.30 & 0.82 & 0.25 \\
\midrule
All trials & & 25 & 0.92 & 0.17 & 0.82 & 0.24 \\
\bottomrule
\end{tabular}
\end{table}
\begin{table*}[t]
\centering
\caption{\textbf{Family tree with newer open model families, all models.} 45 models, the same 400 COCO captions, sets of $k=16$ images, 72 sets per model from 3 caption splits $\times$ 2 halves, 17 registered bases (marked base). To base: share of a model\textquoteright s sets whose nearest registered fingerprint is the base of its family, with CF or CLIP ViT-L/14 features and the fingerprint distance (subspace fit on CF-Small without the Family-Tree generator names) or the label-free distance (whitened PCA basis, no generator labels). Top base and nearest model: CF, fingerprint distance. Rows marked other seeds or steps are the same checkpoint as their family\textquoteright s base, rendered with the COCO-dataset seeds and sampler. With 17 registered bases, Animagine-XL-3.1 goes to CogView4, not to Playground v2 as with the four Family-Tree bases (\cref{tab:supp_style}). Flex.1-alpha ran with its shipped scheduler, unlike the other FLUX.1 models.}
\label{tab:supp_modern_all}
\fontsize{8}{9.5}\selectfont
\setlength{\tabcolsep}{3.5pt}
\begin{tabular}{@{}llcccc@{\hspace{8pt}}ll@{}}
\toprule
& & \multicolumn{4}{c}{To base $\uparrow$} & & \\
\cmidrule(lr){3-6}
& & \multicolumn{2}{c}{CF} & \multicolumn{2}{c}{CLIP ViT-L/14} & & \\
\cmidrule(lr){3-4}\cmidrule(lr){5-6}
Model & Base & Fingerprint & Label-free & Fingerprint & Label-free & Top base & Nearest model \\
\midrule
\multicolumn{8}{@{}l}{\textit{SD1}} \\
SD1.5 & base & - & - & - & - & - & SD1.4 \\
SD1.4 & SD1.5 & 1.00 & 1.00 & 0.99 & 0.88 & SD1.5 & SD1.5 \\
DreamShaper-8 & SD1.5 & 0.00 & 0.00 & 0.00 & 0.03 & SD2.1 & DS-8$\times$RV-5.1 (0.50) \\
RealisticVision-5.1 & SD1.5 & 0.00 & 0.04 & 0.07 & 0.29 & SD2.1 & DS-8$\times$RV-5.1 (0.50) \\
DS-8$\times$RV-5.1 (0.50) & SD1.5 & 0.00 & 0.00 & 0.00 & 0.13 & SD2.1 & DreamShaper-8 \\
Openjourney & SD1.5 & 0.00 & 0.00 & 0.24 & 0.31 & SD2.1 & SD2.1 \\
Anything-v5 & SD1.5 & 0.01 & 0.08 & 0.00 & 0.19 & SD2.1 & SD2.1-768 \\
SD1.5$\times$DS-8 (0.25) & SD1.5 & 1.00 & 1.00 & 0.90 & 0.68 & SD1.5 & SD1.5$\times$DS-8 (0.50) \\
SD1.5$\times$DS-8 (0.50) & SD1.5 & 0.97 & 0.99 & 0.64 & 0.51 & SD1.5 & SD1.5$\times$DS-8 (0.25) \\
SD1.5$\times$DS-8 (0.75) & SD1.5 & 0.69 & 0.86 & 0.03 & 0.18 & SD1.5 & SD1.5$\times$DS-8 (0.50) \\
SD1.5, 30 steps & SD1.5 & 1.00 & 0.99 & 0.96 & 0.94 & SD1.5 & SD1.5 \\
\addlinespace[1pt]
\multicolumn{8}{@{}l}{\textit{SD2}} \\
SD2.1 & base & - & - & - & - & - & Openjourney \\
SD2.1-768 & SD2.1 & 0.99 & 0.97 & 0.86 & 0.71 & SD2.1 & SD2.1 \\
\addlinespace[1pt]
\multicolumn{8}{@{}l}{\textit{SDXL}} \\
SDXL & base & - & - & - & - & - & SDXL, other seeds \\
DreamShaper-XL & SDXL & 1.00 & 1.00 & 1.00 & 0.99 & SDXL & SDXL, other seeds \\
Animagine-XL-3.1 & SDXL & 0.00 & 0.00 & 0.00 & 0.06 & CogView4 & CogView4 \\
SDXL$\times$Anim-XL (0.50) & SDXL & 0.99 & 0.93 & 0.99 & 0.89 & SDXL & SDXL, other seeds \\
SDXL-Turbo & SDXL & 1.00 & 1.00 & 0.93 & 0.67 & SDXL & SDXL, other seeds \\
SDXL, other seeds & SDXL & 1.00 & 1.00 & 1.00 & 0.97 & SDXL & SDXL \\
\addlinespace[1pt]
\multicolumn{8}{@{}l}{\textit{Playground v2}} \\
Playground v2 & base & - & - & - & - & - & SANA \\
\addlinespace[1pt]
\multicolumn{8}{@{}l}{\textit{FLUX.1}} \\
FLUX.1-dev & base & - & - & - & - & - & FLUX.1-dev, 20 steps \\
FLUX.1-schnell & FLUX.1-dev & 0.96 & 0.93 & 0.14 & 0.22 & FLUX.1-dev & FLUX.1-dev \\
FLUX.1-lite-8B & FLUX.1-dev & 0.96 & 0.94 & 1.00 & 0.93 & FLUX.1-dev & FLUX.1-dev, 20 steps \\
FLUX.1-dev + Turbo-Alpha & FLUX.1-dev & 0.90 & 0.94 & 0.96 & 0.93 & FLUX.1-dev & FLUX.1-dev \\
FLUX.1-Krea-dev & FLUX.1-dev & 0.00 & 0.01 & 0.14 & 0.17 & Gemini 2.5 Flash Image & AWPortrait-FL \\
FLUX.1-dev$\times$Krea (0.50) & FLUX.1-dev & 0.99 & 0.96 & 0.96 & 0.90 & FLUX.1-dev & FLUX.1-dev \\
AWPortrait-FL & FLUX.1-dev & 0.04 & 0.08 & 0.31 & 0.36 & Gemini 2.5 Flash Image & FLUX.1-Krea-dev \\
Flex.1-alpha & FLUX.1-dev & 0.00 & 0.00 & 0.01 & 0.01 & Z-Image-Turbo & Z-Image-Turbo \\
FLUX.1-dev, 20 steps & FLUX.1-dev & 0.97 & 0.97 & 0.97 & 0.82 & FLUX.1-dev & FLUX.1-dev \\
\addlinespace[1pt]
\multicolumn{8}{@{}l}{\textit{SD3.5}} \\
SD3.5-large & base & - & - & - & - & - & SD3.5-medium \\
SD3.5-large-TurboX & SD3.5-large & 0.99 & 0.99 & 1.00 & 1.00 & SD3.5-large & SD3.5-large \\
SD3.5-medium-turbo & SD3.5-large & 0.43 & 0.49 & 0.07 & 0.26 & SD3.5-large & SD3.5-medium \\
SD3.5-large-turbo & SD3.5-large & 0.01 & 0.04 & 0.94 & 0.94 & Lumina-Image 2.0 & Lumina-Image 2.0 \\
SD3.5-medium & SD3.5-large & 0.96 & 0.97 & 0.74 & 0.60 & SD3.5-large & SD3.5-large \\
\addlinespace[1pt]
\multicolumn{8}{@{}l}{\textit{Other registered bases, one model per family}} \\
Z-Image-Turbo & base & - & - & - & - & - & Flex.1-alpha \\
SANA 1.5 & base & - & - & - & - & - & SANA \\
SANA & base & - & - & - & - & - & PixArt-$\Sigma$ \\
Lumina-Image 2.0 & base & - & - & - & - & - & CogView4 \\
PixArt-$\Sigma$ & base & - & - & - & - & - & SANA \\
Kolors & base & - & - & - & - & - & PixArt-$\Sigma$ \\
CogView4 & base & - & - & - & - & - & Lumina-Image 2.0 \\
Hunyuan-DiT & base & - & - & - & - & - & PixArt-$\Sigma$ \\
Playground v2.5 & base & - & - & - & - & - & SD3.5-medium \\
Gemini 2.5 Flash Image & base & - & - & - & - & - & FLUX.1-Krea-dev \\
Gemini 3.1 Flash Image & base & - & - & - & - & - & Z-Image-Turbo \\
\bottomrule
\end{tabular}
\end{table*}
\begin{table*}[t]
\centering
\caption{\textbf{Family distances with newer open model families.} Setting of \cref{tab:supp_modern_all}. Within: mean distance between the members of a family. Across: mean distance from its members to every model of another family. Same-model controls are left out of the family means. Controls: distance between two renderings of one checkpoint with different seeds (and steps for FLUX.1-dev and SD1.5), next to the median distance across families. The SANA row compares SANA 1.0 with SANA 1.5, a newer version of the same line. Rank: rank of the partner among the nearest models.}
\label{tab:supp_modern_families}
\footnotesize
\setlength{\tabcolsep}{4pt}
\begin{tabular}{@{}lc cc cc cc cc@{}}
\toprule
& & \multicolumn{4}{c}{CF} & \multicolumn{4}{c}{CLIP ViT-L/14} \\
\cmidrule(lr){3-6}\cmidrule(l){7-10}
& & \multicolumn{2}{c}{Fingerprint} & \multicolumn{2}{c}{Label-free} & \multicolumn{2}{c}{Fingerprint} & \multicolumn{2}{c}{Label-free} \\
\cmidrule(lr){3-4}\cmidrule(lr){5-6}\cmidrule(lr){7-8}\cmidrule(l){9-10}
Family & Models & Within & Across & Within & Across & Within & Across & Within & Across \\
\midrule
SD1 & 10 & 2.74 & 8.60 & 2.22 & 7.98 & 2.02 & 2.73 & 1.40 & 2.08 \\
SD2 & 2 & 2.79 & 7.02 & 2.53 & 6.48 & 0.82 & 2.42 & 0.55 & 1.80 \\
SDXL & 5 & 5.87 & 7.54 & 5.37 & 7.01 & 2.59 & 2.80 & 1.81 & 2.02 \\
FLUX.1 & 8 & 3.61 & 7.93 & 3.31 & 7.47 & 1.60 & 2.51 & 1.25 & 1.90 \\
SD3.5 & 5 & 4.60 & 7.42 & 4.22 & 6.92 & 1.78 & 2.65 & 1.44 & 2.07 \\
\midrule
Control & Partner & Dist. & Median & Dist. & Median & Dist. & Median & Dist. & Median \\
\midrule
FLUX.1-dev, 20 steps & FLUX.1-dev & 1.02 & 7.40 & 1.01 & 6.90 & 0.46 & 2.47 & 0.42 & 1.92 \\
SD1.5, 30 steps & SD1.5 & 0.45 & 7.40 & 0.41 & 6.90 & 0.30 & 2.47 & 0.29 & 1.92 \\
SDXL, other seeds & SDXL & 0.43 & 7.40 & 0.39 & 6.90 & 0.26 & 2.47 & 0.24 & 1.92 \\
SANA & SANA 1.5 & 2.27$^{(2)}$ & 7.40 & 1.88 & 6.90 & 1.50$^{(7)}$ & 2.47 & 1.16$^{(4)}$ & 1.92 \\
\bottomrule
\multicolumn{10}{@{}l}{\scriptsize Superscript: rank of the partner among the nearest models when it is not the nearest.} \\
\end{tabular}
\end{table*}
\begin{table*}[t]
\centering
\caption{\textbf{Newer open models added to the Family-Tree set.} Fifteen models rendered for this experiment and 12 generators whose COCO-dataset images of the same 400 captions already existed (COCO-dataset seeds, not the Family-Tree seeds). Relation: as documented by the model card or by construction. Sampler: steps / guidance / resolution in pixels. Rendered models use the Family-Tree seed of each caption. FLUX.1 models share the FLUX.1-dev text encoders and autoencoder, and SD3.5 models share the SD3.5-large text encoders and autoencoder.}
\label{tab:supp_modern_models}
\footnotesize
\setlength{\tabcolsep}{4pt}
\begin{tabular}{@{}lp{6.6cm}p{4.6cm}c@{}}
\toprule
Model & Checkpoint & Relation & Sampler \\
\midrule
\multicolumn{4}{@{}l}{\textit{Rendered for this experiment}} \\
FLUX.1-dev & \raggedright \nolinkurl{black-forest-labs/FLUX.1-dev} & \raggedright registered base (guidance-distilled from FLUX.1 pro) & 28 / 3.5 / 1,024 \\
FLUX.1-schnell & \raggedright \nolinkurl{black-forest-labs/FLUX.1-schnell} & \raggedright sibling of FLUX.1-dev (step-distilled from the FLUX.1 line) & 4 / 0.0 / 1,024 \\
FLUX.1-lite-8B & \raggedright \nolinkurl{Freepik/flux.1-lite-8B} & \raggedright distilled from FLUX.1-dev (8B) & 28 / 3.5 / 1,024 \\
FLUX.1-dev + Turbo-Alpha & \raggedright \nolinkurl{black-forest-labs/FLUX.1-dev} + \nolinkurl{alimama-creative/FLUX.1-Turbo-Alpha} & \raggedright 8-step distillation LoRA on FLUX.1-dev, fused at scale 1 & 8 / 3.5 / 1,024 \\
FLUX.1-Krea-dev & \raggedright \nolinkurl{black-forest-labs/FLUX.1-Krea-dev} & \raggedright card lists FLUX.1-dev as base model & 28 / 3.5 / 1,024 \\
FLUX.1-dev$\times$Krea (0.50) & \raggedright FLUX.1-dev and FLUX.1-Krea-dev & \raggedright transformer merge, 0.5 FLUX.1-dev + 0.5 FLUX.1-Krea-dev & 28 / 3.5 / 1,024 \\
AWPortrait-FL & \raggedright \nolinkurl{Shakker-Labs/AWPortrait-FL} & \raggedright full fine-tune of FLUX.1-dev & 28 / 3.5 / 1,024 \\
Flex.1-alpha & \raggedright \nolinkurl{ostris/Flex.1-alpha} & \raggedright fine-tune of FLUX.1-schnell, pruned to 8B and retrained & 28 / 3.5 / 1,024 \\
SD3.5-large & \raggedright \nolinkurl{stabilityai/stable-diffusion-3.5-large} & \raggedright registered base & 28 / 3.5 / 1,024 \\
SD3.5-large-TurboX & \raggedright \nolinkurl{tensorart/stable-diffusion-3.5-large-TurboX} & \raggedright third-party few-step variant of SD3.5-large & 8 / 1.5 / 1,024 \\
SD3.5-large-turbo & \raggedright \nolinkurl{stabilityai/stable-diffusion-3.5-large-turbo} & \raggedright official distillation of SD3.5-large & 4 / 0.0 / 1,024 \\
SD3.5-medium & \raggedright \nolinkurl{stabilityai/stable-diffusion-3.5-medium} & \raggedright separately trained smaller SD3.5 model & 40 / 4.5 / 1,024 \\
SD3.5-medium-turbo & \raggedright \nolinkurl{tensorart/stable-diffusion-3.5-medium-turbo} & \raggedright distilled from SD3.5-medium & 8 / 1.5 / 1,024 \\
Z-Image-Turbo & \raggedright \nolinkurl{Tongyi-MAI/Z-Image-Turbo} & \raggedright registered base (distilled Z-Image) & 9 / 0.0 / 1,024 \\
SANA 1.5 & \raggedright \nolinkurl{Efficient-Large-Model/SANA1.5_1.6B_1024px_diffusers} & \raggedright registered base & 20 / 4.5 / 1,024 \\
\midrule
\multicolumn{4}{@{}l}{\textit{Existing COCO-dataset images}} \\
FLUX.1-dev, 20 steps & \raggedright \nolinkurl{black-forest-labs/FLUX.1-dev} & \raggedright same model as FLUX.1-dev, other seeds and sampler & 20 / 3.5 / 1,024 \\
Lumina-Image 2.0 & \raggedright \nolinkurl{Alpha-VLLM/Lumina-Image-2.0} & \raggedright registered base & 30 / 4.0 / 1,024 \\
SANA & \raggedright \nolinkurl{Efficient-Large-Model/Sana_1600M_1024px_diffusers} & \raggedright registered base & 20 / 5.0 / 1,024 \\
PixArt-$\Sigma$ & \raggedright \nolinkurl{PixArt-alpha/PixArt-Sigma-XL-2-1024-MS} & \raggedright registered base & 20 / 4.5 / 1,024 \\
Kolors & \raggedright \nolinkurl{Kwai-Kolors/Kolors-diffusers} & \raggedright registered base & 30 / 5.0 / 1,024 \\
CogView4 & \raggedright \nolinkurl{THUDM/CogView4-6B} & \raggedright registered base & 30 / 3.5 / 1,024 \\
Hunyuan-DiT & \raggedright \nolinkurl{Tencent-Hunyuan/HunyuanDiT-Diffusers} & \raggedright registered base & 20 / 5.0 / 1,024 \\
Playground v2.5 & \raggedright \nolinkurl{playgroundai/playground-v2.5-1024px-aesthetic} & \raggedright registered base & 30 / 3.0 / 1,024 \\
Gemini 2.5 Flash Image & \raggedright \nolinkurl{gemini-2.5-flash-image (API)} & \raggedright registered base (API) & - / - / 1,024 \\
Gemini 3.1 Flash Image & \raggedright \nolinkurl{gemini-3.1-flash-image (API)} & \raggedright registered base (API) & - / - / 1,024 \\
SD1.5, 30 steps & \raggedright \nolinkurl{runwayml/stable-diffusion-v1-5} & \raggedright same model as SD1.5, other seeds and sampler & 30 / 7.5 / 512 \\
SDXL, other seeds & \raggedright \nolinkurl{stabilityai/stable-diffusion-xl-base-1.0} & \raggedright same model as SDXL, other seeds & 30 / 7.0 / 1,024 \\
\bottomrule
\end{tabular}
\end{table*}
 
This section gives the full setup and every model of the FLUX.1 and SD3.5 experiment of \cref{sec:exp_tree}.

\noindent\textbf{Setup.}
We add 27 models to the 18 Family-Tree models, and all 45 render the same 400 COCO captions (\cref{tab:supp_modern_models}).
Fifteen newer open models are rendered at 1,024 pixels: FLUX.1-dev~\cite{bfl2024flux} with seven relatives, SD3.5-large~\cite{esser2024sd3} with four relatives, Z-Image-Turbo and SANA 1.5.
The other 12 are COCO-dataset generators whose images of these captions already existed, and three of them repeat FLUX.1-dev, SD1.5 and SDXL with other seeds.
Sets hold $k=16$ images of 16 different captions, and 3 caption splits $\times$ 2 halves give 72 sets per model.
We use CF features and the fingerprint distance of \cref{sec:supp_tree_metric}. Its subspace is fit on CF-Small, which holds no FLUX or SD3 generator.
Each set is attributed to the nearest of 17 registered bases: SD1.5, SD2.1, SDXL, Playground v2, FLUX.1-dev, SD3.5-large, Z-Image-Turbo, SANA 1.5 and the nine COCO-dataset generators that are not repeats.
In this experiment SD3.5-large is a registered base, while in the discovery tests of \cref{sec:exp_discovery} it is unseen.
Each model is rendered once, and all numbers of this section come from one run.

Three tests were fixed before the run.
(1) FLUX.1-lite-8B and FLUX.1-dev with the fused Turbo-Alpha LoRA are attributed to FLUX.1-dev, and SD3.5-large-TurboX to SD3.5-large, in at least 0.80 of their sets.
(2) In the FLUX.1 and SD3.5 families the mean distance within the family is below the mean distance to other families.
(3) The merge of FLUX.1-dev and FLUX.1-Krea-dev with weight 0.50 lies between its parents.

\noindent\textbf{Results.}
All three tests pass (\cref{tab:modern}).
The parents of the merge lie 4.73 apart, and the merge lies 1.66 from the line between them.
All six merges of the paper lie nearer their first parent than their mixing weight ($t=0.20$ for $w=0.50$ here), so $t$ orders merges between their parents and does not estimate $w$.
The five models that are not attributed to the base of their family lie farther from it than from the nearest other base (\cref{tab:modern}).
\Cref{tab:supp_modern_all} lists all 45 models with CF and CLIP ViT-L/14 features and with both distances.
Twelve of its 14 FLUX.1 and SD3.5 models have their nearest model in their own family. Flex.1-alpha lies nearest to Z-Image-Turbo and SD3.5-large-turbo to Lumina-Image 2.0.
With CLIP ViT-L/14 features FLUX.1-schnell is attributed to FLUX.1-dev in 0.14 of its sets and SD3.5-medium-turbo to SD3.5-large in 0.07.
\Cref{tab:supp_modern_families} gives the family means and the distance between two renderings of one checkpoint.
\Cref{tab:supp_verification} scores every set of the 25 relatives against every registered base. Over all trials the fingerprint distance verifies the documented family with AUROC 0.92 with CF features and 0.82 with CLIP ViT-L/14 features.
The relations are taken from the model cards.

\clearpage  %
\section{Autoencoder and resolution controls}
\label{sec:supp_controls}

\begin{table}[t]
\centering
\caption{\textbf{Autoencoder and sampling-resolution controls.} Family-Tree captions, 24 disjoint sets of $k=16$ images per condition, one run, CF features and the fingerprint distance of \cref{sec:supp_tree_metric}. Each set goes to the nearest base fingerprint (SD1.5, SD2.1, SDXL and PG v2, plus FLUX.1-dev and SD3.5-large for those two models). AE: autoencoder. A round trip encodes a finished image and decodes it again. In family: sets out of 24 attributed to the model's own family after the manipulation (24 before, and $a{\to}b$ means $a$ before). Moved: fingerprint distance to the unmanipulated model. Among the 15 models of this experiment the mean distance is 4.13 within a family and 7.50 across families, and the own AE or decoder moves a model by 0.48 on average. Toward: position between the own base (0) and the base the borrowed AE or decoder comes from (1), and for 512-pixel sampling toward the 512-pixel bases SD1.5 and SD2.1. It is the larger value when there are two such bases and -- where none is borrowed. Ranges: minimum--maximum over the listed models and AEs. 3 AEs: SDXL, SD3.5 and FLUX.1. 4 AEs: those of the other four families. 2 decoders: FP16-Fix and TAESDXL. DS-XL: DreamShaper-XL, DS-8: DreamShaper-8, PG v2: Playground v2. $^\dagger$SD1.5 fine-tunes that ship their own decoder and go to SD2.1 before any manipulation. Every SD1.5 set that leaves its family goes to SD2.1. The rule fixed before the run (at least 20 of 24 sets stay in every tested condition) holds for 27 of 32 round trips, 6 of 10 decoder swaps and 1 of 4 resolution changes, and with SD1 and SD2 as one family for 31 of 32, 10 of 10 and 1 of 4. Every condition is in \cref{tab:supp_controls_rt,tab:supp_controls_swap}.}
\label{tab:controls}
\small
\setlength{\tabcolsep}{2.4pt}
\begin{tabular}{@{}lccc@{}}
\toprule
Manipulation & In family & Moved & Toward \\
\midrule
\multicolumn{4}{@{}l}{\textit{Round trip through another family's AE}} \\
SD1.5 via SD2.1 AE & 3 & 1.68 & 0.70 \\
SD2.1 via SD1.5 AE & 1 & 1.91 & 0.94 \\
SD2.1-768 via SD1.5 AE & 22 & 1.42 & 0.57 \\
SD1.5 via SDXL AE & 0 & 1.99 & 0.12 \\
SD1.5 via FLUX.1 AE & 13 & 1.52 & 0.03 \\
SD1.5 via SD3.5 AE & 24 & 1.35 & 0.02 \\
SD2.1, SD2.1-768, 3 AEs & 21--24 & 0.91--1.26 & 0.02--0.08 \\
SDXL, DS-XL, PG v2, 4 AEs & 24 & 0.43--2.98 & $-$0.01--0.17 \\
FLUX.1-dev, 4 AEs & 19--24 & 0.53--3.47 & 0.01--0.28 \\
SD3.5-large, 4 AEs & 22--24 & 0.50--4.00 & 0.02--0.31 \\
\addlinespace[2pt]
\multicolumn{4}{@{}l}{\textit{Same latents, other decoder}} \\
SD1.5 with SD2.1 decoder & 0 & 1.81 & 0.90 \\
SD2.1 with SD1.5 decoder & 1 & 1.83 & 0.92 \\
Openjourney, SD1.5 dec.$^\dagger$ & 0${\to}$24 & 1.61 & -- \\
DS-8, SD1.5 dec.$^\dagger$ & 0${\to}$11 & 2.23 & -- \\
SDXL, PG v2 with 2 decoders & 24 & 0.21--4.66 & 0.00 \\
\addlinespace[2pt]
\multicolumn{4}{@{}l}{\textit{Sampling resolution}} \\
SDXL at 512 px & 0 & 8.07 & 0.97 \\
DreamShaper-XL at 512 px & 0 & 8.39 & 0.82 \\
PG v2 at 512 px & 0 & 6.84 & 0.78 \\
SDXL-Turbo at 1,024 px & 24 & 6.83 & -- \\
\bottomrule
\end{tabular}
\end{table}
\begin{table*}[t]
\centering
\caption{\textbf{Autoencoder round trips, every condition.} Setting of \cref{tab:controls}: Family-Tree captions, 24 disjoint sets of $k=16$ images per condition, base fingerprints from the other caption half, one run. Each finished image is encoded to the posterior mode and decoded at its native resolution, in float32. Before / After: sets out of 24 attributed to the model's own family without / with the manipulation (nearest of SD1.5, SD2.1, SDXL and PG v2, plus FLUX.1-dev and SD3.5-large for those two models). SD1+2: After with SD1 and SD2 counted as one family. Majority base: the base most sets go to (count). Moved: fingerprint distance to the unmanipulated model. Own AE: the same distance for the model's own AE or own decoder on the same latents (-- where none exists). Toward: position between the own base (0) and the base the borrowed AE belongs to (1), the larger value if it serves two. Net: the same measured from the own-AE condition instead of the unmanipulated model (-- where no other base is borrowed). CLIP L/14 and CF label-free: Before / After with CLIP ViT-L/14 features in the fingerprint distance and with CF features in the label-free distance. Reference distances (CF, fingerprint distance): 4.13 within a family and 7.50 across families. $^\ast$Own AE (reference, not tested). $^\dagger$Model already outside its family before the round trip (DreamShaper-8 goes to SD2.1), not tested. At least 20 of 24 sets stay for 27 of 32 tested conditions.}
\label{tab:supp_controls_rt}
\scriptsize\renewcommand{\arraystretch}{0.96}
\setlength{\tabcolsep}{2.4pt}
\begin{tabular}{@{}llcccl@{\hspace{6pt}}cccc@{\hspace{8pt}}cc@{\hspace{8pt}}cc@{}}
\toprule
& & \multicolumn{4}{c}{CF, fingerprint distance} & \multicolumn{4}{c}{CF, distances} & \multicolumn{2}{c}{CLIP L/14} & \multicolumn{2}{c}{CF label-free} \\
\cmidrule(lr){3-6}\cmidrule(lr){7-10}\cmidrule(lr){11-12}\cmidrule(l){13-14}
Model & Manipulation & Before & After & SD1+2 & Majority base & Moved & Own AE & Toward & Net & Before & After & Before & After \\
\midrule
\multicolumn{14}{@{}l}{\textit{Round trip of finished images through an autoencoder}} \\
SD1.5$^\ast$ & SD1.5 AE (own) & 24 & 24 & 24 & SD1.5 (24) & 1.02 & -- & -- & -- & 22 & 24 & 24 & 24 \\
SD1.5 & SD2.1 AE & 24 & 3 & 24 & SD2.1 (21) & 1.68 & 1.02 & 0.70 & 0.90 & 22 & 5 & 24 & 4 \\
SD1.5 & SDXL AE & 24 & 0 & 24 & SD2.1 (24) & 1.99 & 1.02 & 0.12 & 0.12 & 22 & 7 & 24 & 0 \\
SD1.5 & SD3.5 AE & 24 & 24 & 24 & SD1.5 (24) & 1.35 & 1.02 & 0.02 & 0.02 & 22 & 19 & 24 & 24 \\
SD1.5 & FLUX.1 AE & 24 & 13 & 24 & SD1.5 (13) & 1.52 & 1.02 & 0.03 & 0.02 & 22 & 15 & 24 & 17 \\
\addlinespace[1pt]
DS-8$^\ast$ & DS-8 AE (own) & 0 & 0 & 24 & SD2.1 (24) & 0.71 & -- & -- & -- & 0 & 0 & 0 & 0 \\
DS-8$^\dagger$ & SD1.5 AE & 0 & 12 & 24 & SD1.5/SD2.1 (12 each) & 2.17 & 0.71 & -- & -- & 0 & 6 & 0 & 23 \\
DS-8$^\dagger$ & SD2.1 AE & 0 & 0 & 24 & SD2.1 (24) & 0.77 & 0.71 & $-$0.07 & 0.00 & 0 & 0 & 0 & 0 \\
DS-8$^\dagger$ & SDXL AE & 0 & 0 & 24 & SD2.1 (24) & 1.25 & 0.71 & 0.09 & 0.12 & 0 & 0 & 0 & 0 \\
DS-8$^\dagger$ & SD3.5 AE & 0 & 0 & 24 & SD2.1 (24) & 0.89 & 0.71 & 0.00 & 0.02 & 0 & 0 & 0 & 0 \\
DS-8$^\dagger$ & FLUX.1 AE & 0 & 0 & 24 & SD2.1 (24) & 0.64 & 0.71 & 0.02 & 0.04 & 0 & 0 & 0 & 0 \\
\addlinespace[1pt]
SD2.1$^\ast$ & SD2.1 AE (own) & 24 & 24 & 24 & SD2.1 (24) & 0.76 & -- & -- & -- & 22 & 21 & 24 & 24 \\
SD2.1 & SD1.5 AE & 24 & 1 & 24 & SD1.5 (23) & 1.91 & 0.76 & 0.94 & 0.86 & 22 & 8 & 24 & 1 \\
SD2.1 & SDXL AE & 24 & 24 & 24 & SD2.1 (24) & 1.26 & 0.76 & 0.08 & 0.08 & 22 & 24 & 24 & 24 \\
SD2.1 & SD3.5 AE & 24 & 24 & 24 & SD2.1 (24) & 0.94 & 0.76 & 0.02 & 0.02 & 22 & 22 & 24 & 23 \\
SD2.1 & FLUX.1 AE & 24 & 24 & 24 & SD2.1 (24) & 1.04 & 0.76 & 0.03 & 0.03 & 22 & 24 & 24 & 24 \\
\addlinespace[1pt]
SD2.1-768$^\ast$ & SD2.1 AE (own) & 24 & 24 & 24 & SD2.1 (24) & 0.80 & -- & -- & -- & 20 & 20 & 24 & 24 \\
SD2.1-768 & SD1.5 AE & 24 & 22 & 24 & SD2.1 (22) & 1.42 & 0.80 & 0.57 & 0.54 & 20 & 12 & 24 & 22 \\
SD2.1-768 & SDXL AE & 24 & 24 & 24 & SD2.1 (24) & 1.00 & 0.80 & 0.07 & 0.09 & 20 & 20 & 24 & 24 \\
SD2.1-768 & SD3.5 AE & 24 & 24 & 24 & SD2.1 (24) & 0.91 & 0.80 & 0.04 & 0.05 & 20 & 20 & 24 & 24 \\
SD2.1-768 & FLUX.1 AE & 24 & 21 & 21 & SD2.1 (21) & 1.16 & 0.80 & 0.07 & 0.08 & 20 & 21 & 24 & 22 \\
\addlinespace[1pt]
SDXL$^\ast$ & SDXL AE (own) & 24 & 24 & 24 & SDXL (24) & 0.85 & -- & -- & -- & 24 & 24 & 24 & 24 \\
SDXL & SD1.5 AE & 24 & 24 & 24 & SDXL (24) & 1.84 & 0.85 & 0.14 & 0.11 & 24 & 24 & 24 & 24 \\
SDXL & SD2.1 AE & 24 & 24 & 24 & SDXL (24) & 1.03 & 0.85 & 0.10 & 0.06 & 24 & 24 & 24 & 24 \\
SDXL & SD3.5 AE & 24 & 24 & 24 & SDXL (24) & 0.66 & 0.85 & $-$0.01 & 0.05 & 24 & 24 & 24 & 24 \\
SDXL & FLUX.1 AE & 24 & 24 & 24 & SDXL (24) & 0.48 & 0.85 & 0.01 & 0.07 & 24 & 24 & 24 & 24 \\
\addlinespace[1pt]
DreamShaper-XL$^\ast$ & SDXL AE (own) & 24 & 24 & 24 & SDXL (24) & 1.09 & -- & -- & -- & 24 & 22 & 24 & 24 \\
DreamShaper-XL & SD1.5 AE & 24 & 24 & 24 & SDXL (24) & 2.98 & 1.09 & 0.17 & 0.15 & 24 & 24 & 24 & 24 \\
DreamShaper-XL & SD2.1 AE & 24 & 24 & 24 & SDXL (24) & 1.73 & 1.09 & 0.11 & 0.08 & 24 & 24 & 24 & 24 \\
DreamShaper-XL & SD3.5 AE & 24 & 24 & 24 & SDXL (24) & 0.60 & 1.09 & 0.01 & 0.10 & 24 & 24 & 24 & 24 \\
DreamShaper-XL & FLUX.1 AE & 24 & 24 & 24 & SDXL (24) & 0.54 & 1.09 & 0.01 & 0.08 & 24 & 24 & 24 & 24 \\
\addlinespace[1pt]
PG v2$^\ast$ & SDXL AE (own) & 24 & 24 & 24 & PG v2 (24) & 0.65 & -- & -- & -- & 24 & 24 & 24 & 24 \\
PG v2 & SD1.5 AE & 24 & 24 & 24 & PG v2 (24) & 1.83 & 0.65 & 0.16 & 0.13 & 24 & 24 & 24 & 24 \\
PG v2 & SD2.1 AE & 24 & 24 & 24 & PG v2 (24) & 0.89 & 0.65 & 0.08 & 0.04 & 24 & 24 & 24 & 24 \\
PG v2 & SD3.5 AE & 24 & 24 & 24 & PG v2 (24) & 0.78 & 0.65 & 0.00 & 0.00 & 24 & 24 & 24 & 24 \\
PG v2 & FLUX.1 AE & 24 & 24 & 24 & PG v2 (24) & 0.43 & 0.65 & 0.00 & 0.01 & 24 & 24 & 24 & 24 \\
\addlinespace[1pt]
FLUX.1-dev$^\ast$ & FLUX.1 AE (own) & 24 & 24 & 24 & FLUX.1-dev (24) & 0.44 & -- & -- & -- & 24 & 24 & 24 & 24 \\
FLUX.1-dev & SD1.5 AE & 24 & 19 & 19 & FLUX.1-dev (19) & 3.47 & 0.44 & 0.28 & 0.26 & 24 & 24 & 24 & 18 \\
FLUX.1-dev & SD2.1 AE & 24 & 24 & 24 & FLUX.1-dev (24) & 2.32 & 0.44 & 0.18 & 0.16 & 24 & 24 & 24 & 24 \\
FLUX.1-dev & SDXL AE & 24 & 24 & 24 & FLUX.1-dev (24) & 1.59 & 0.44 & 0.17 & 0.14 & 24 & 24 & 24 & 24 \\
FLUX.1-dev & SD3.5 AE & 24 & 24 & 24 & FLUX.1-dev (24) & 0.53 & 0.44 & 0.01 & 0.01 & 24 & 24 & 24 & 24 \\
\addlinespace[1pt]
SD3.5-large$^\ast$ & SD3.5 AE (own) & 24 & 24 & 24 & SD3.5-large (24) & 0.53 & -- & -- & -- & 24 & 22 & 24 & 24 \\
SD3.5-large & SD1.5 AE & 24 & 22 & 22 & SD3.5-large (22) & 4.00 & 0.53 & 0.31 & 0.29 & 24 & 22 & 24 & 21 \\
SD3.5-large & SD2.1 AE & 24 & 24 & 24 & SD3.5-large (24) & 2.77 & 0.53 & 0.22 & 0.20 & 24 & 24 & 24 & 24 \\
SD3.5-large & SDXL AE & 24 & 24 & 24 & SD3.5-large (24) & 2.02 & 0.53 & 0.21 & 0.19 & 24 & 24 & 24 & 24 \\
SD3.5-large & FLUX.1 AE & 24 & 24 & 24 & SD3.5-large (24) & 0.50 & 0.53 & 0.02 & 0.02 & 24 & 24 & 24 & 24 \\
\bottomrule
\end{tabular}
\end{table*}
\begin{table*}[t]
\centering
\caption{\textbf{Decoder swaps and resolution changes, every condition.} Setting of \cref{tab:controls}. Decoder swap: one set of latents per model, sampled with the Family-Tree settings, is decoded by its own decoder and by each other decoder in float32. ft-EMA is sd-vae-ft-ema, FP16-Fix is sdxl-vae-fp16-fix, and TAESD and TAESDXL are tiny decoders of a different architecture. Resolution: the SDXL-family models sampled at 512 pixels, and SDXL-Turbo, which the Family-Tree set samples at 512 pixels, sampled at 1,024. Resize: bicubic resize of finished images before the standard resizing to 256 pixels, with the median PSNR in dB between the two 256-pixel images in brackets (52.2 to 56.3 dB). Before / After: sets out of 24 attributed to the model's own family without / with the manipulation (nearest of SD1.5, SD2.1, SDXL and PG v2, plus FLUX.1-dev and SD3.5-large for those two models). SD1+2: After with SD1 and SD2 counted as one family. Majority base: the base most sets go to (count). Moved: fingerprint distance to the unmanipulated model. Own AE: the same distance for the model's own AE or own decoder on the same latents (-- where none exists). Toward: position between the own base (0) and the base the borrowed AE belongs to (1), the larger value if it serves two. Net: the same measured from the own-AE condition instead of the unmanipulated model (-- where no other base is borrowed). CLIP L/14 and CF label-free: Before / After with CLIP ViT-L/14 features in the fingerprint distance and with CF features in the label-free distance. $^\ast$Own decoder (reference, not tested). $^\dagger$Model outside its family before any manipulation (DreamShaper-8, Openjourney, RealisticVision-5.1 and Anything-v5 go to SD2.1, and Animagine-XL-3.1 to PG v2), not tested. At least 20 of 24 sets stay for 6 of 10 tested decoder swaps and 1 of 4 tested resolution changes.}
\label{tab:supp_controls_swap}
\scriptsize
\setlength{\tabcolsep}{2.4pt}
\begin{tabular}{@{}llcccl@{\hspace{6pt}}cccc@{\hspace{8pt}}cc@{\hspace{8pt}}cc@{}}
\toprule
& & \multicolumn{4}{c}{CF, fingerprint distance} & \multicolumn{4}{c}{CF, distances} & \multicolumn{2}{c}{CLIP L/14} & \multicolumn{2}{c}{CF label-free} \\
\cmidrule(lr){3-6}\cmidrule(lr){7-10}\cmidrule(lr){11-12}\cmidrule(l){13-14}
Model & Manipulation & Before & After & SD1+2 & Majority base & Moved & Own AE & Toward & Net & Before & After & Before & After \\
\midrule
\multicolumn{14}{@{}l}{\textit{Same latents, other decoder}} \\
SD1.5$^\ast$ & SD1.5 decoder (own) & 24 & 24 & 24 & SD1.5 (24) & 0.04 & -- & -- & -- & 22 & 22 & 24 & 24 \\
SD1.5 & SD2.1 decoder & 24 & 0 & 24 & SD2.1 (24) & 1.81 & 0.04 & 0.90 & 0.89 & 22 & 6 & 24 & 0 \\
SD1.5 & ft-EMA decoder & 24 & 0 & 24 & SD2.1 (24) & 2.06 & 0.04 & -- & -- & 22 & 4 & 24 & 0 \\
SD1.5 & TAESD decoder & 24 & 24 & 24 & SD1.5 (24) & 3.69 & 0.04 & -- & -- & 22 & 19 & 24 & 24 \\
\addlinespace[1pt]
SD2.1$^\ast$ & SD2.1 decoder (own) & 24 & 24 & 24 & SD2.1 (24) & 0.04 & -- & -- & -- & 22 & 22 & 24 & 24 \\
SD2.1 & SD1.5 decoder & 24 & 1 & 24 & SD1.5 (23) & 1.83 & 0.04 & 0.92 & 0.92 & 22 & 11 & 24 & 1 \\
SD2.1 & ft-EMA decoder & 24 & 24 & 24 & SD2.1 (24) & 0.58 & 0.04 & -- & -- & 22 & 22 & 24 & 24 \\
SD2.1 & TAESD decoder & 24 & 1 & 24 & SD1.5 (23) & 3.88 & 0.04 & -- & -- & 22 & 19 & 24 & 1 \\
\addlinespace[1pt]
SDXL$^\ast$ & SDXL decoder (own) & 24 & 24 & 24 & SDXL (24) & 0.13 & -- & -- & -- & 24 & 24 & 24 & 24 \\
SDXL & FP16-Fix decoder & 24 & 24 & 24 & SDXL (24) & 0.21 & 0.13 & -- & -- & 24 & 24 & 24 & 24 \\
SDXL & TAESDXL decoder & 24 & 24 & 24 & SDXL (24) & 4.66 & 0.13 & -- & -- & 24 & 24 & 24 & 24 \\
\addlinespace[1pt]
PG v2$^\ast$ & SDXL decoder (own) & 24 & 24 & 24 & PG v2 (24) & 0.17 & -- & -- & -- & 24 & 24 & 24 & 24 \\
PG v2 & FP16-Fix decoder & 24 & 24 & 24 & PG v2 (24) & 0.25 & 0.17 & 0.00 & 0.00 & 24 & 24 & 24 & 24 \\
PG v2 & TAESDXL decoder & 24 & 24 & 24 & PG v2 (24) & 4.21 & 0.17 & -- & -- & 24 & 24 & 24 & 24 \\
\addlinespace[1pt]
DS-8$^\ast$ & DS-8 decoder (own) & 0 & 0 & 24 & SD2.1 (24) & 0.12 & -- & -- & -- & 0 & 0 & 0 & 0 \\
DS-8$^\dagger$ & SD1.5 decoder & 0 & 11 & 24 & SD2.1 (13) & 2.23 & 0.12 & -- & -- & 0 & 6 & 0 & 22 \\
\addlinespace[1pt]
Openjourney$^\ast$ & Openjourney decoder (own) & 0 & 0 & 24 & SD2.1 (24) & 0.05 & -- & -- & -- & 2 & 3 & 0 & 0 \\
Openjourney$^\dagger$ & SD1.5 decoder & 0 & 24 & 24 & SD1.5 (24) & 1.61 & 0.05 & -- & -- & 2 & 19 & 0 & 24 \\
\addlinespace[1pt]
Anim-XL$^\ast$ & Anim-XL decoder (own) & 0 & 0 & 0 & PG v2 (24) & 0.26 & -- & -- & -- & 17 & 16 & 0 & 0 \\
Anim-XL$^\dagger$ & SDXL decoder & 0 & 0 & 0 & PG v2 (24) & 0.26 & 0.26 & $-$0.01 & $-$0.01 & 17 & 17 & 0 & 0 \\
\addlinespace[2pt]
\multicolumn{14}{@{}l}{\textit{Sampled at another resolution}} \\
SDXL & sampled at 512 px & 24 & 0 & 0 & SD2.1 (23) & 8.07 & -- & 0.97 & -- & 24 & 0 & 24 & 0 \\
DreamShaper-XL & sampled at 512 px & 24 & 0 & 0 & SD1.5/SD2.1 (12 each) & 8.39 & -- & 0.82 & -- & 24 & 0 & 24 & 0 \\
Anim-XL$^\dagger$ & sampled at 512 px & 0 & 0 & 0 & SD2.1 (17) & 9.49 & -- & 0.69 & -- & 17 & 0 & 0 & 0 \\
PG v2 & sampled at 512 px & 24 & 0 & 0 & SD2.1 (24) & 6.84 & -- & 0.78 & -- & 24 & 17 & 24 & 0 \\
SDXL-Turbo & sampled at 1,024 px & 24 & 24 & 24 & SDXL (24) & 6.83 & -- & -- & -- & 23 & 24 & 24 & 24 \\
\addlinespace[2pt]
\multicolumn{14}{@{}l}{\textit{Finished images resized before the standard resizing to 256 pixels (check only, not tested)}} \\
SD1.4 & up to 1,024 px (54.7 dB) & 24 & 24 & 24 & SD1.5 (24) & 0.12 & -- & 0.00 & -- & 23 & 23 & 24 & 24 \\
SD1.5 & up to 1,024 px (54.7 dB) & 24 & 24 & 24 & SD1.5 (24) & 0.12 & -- & 0.00 & -- & 22 & 22 & 24 & 24 \\
DS-8 & up to 1,024 px (55.4 dB) & 0 & 0 & 24 & SD2.1 (24) & 0.16 & -- & 0.00 & -- & 0 & 0 & 0 & 0 \\
RV-5.1 & up to 1,024 px (55.5 dB) & 0 & 0 & 24 & SD2.1 (24) & 0.16 & -- & 0.00 & -- & 2 & 3 & 2 & 2 \\
Openjourney & up to 1,024 px (54.9 dB) & 0 & 0 & 24 & SD2.1 (24) & 0.12 & -- & 0.00 & -- & 2 & 4 & 0 & 0 \\
Anything-v5 & up to 1,024 px (54.8 dB) & 0 & 0 & 24 & SD2.1 (24) & 0.14 & -- & 0.00 & -- & 0 & 0 & 1 & 1 \\
SD2.1 & up to 1,024 px (54.9 dB) & 24 & 24 & 24 & SD2.1 (24) & 0.14 & -- & 0.00 & -- & 22 & 22 & 24 & 24 \\
SDXL-Turbo & up to 1,024 px (56.3 dB) & 24 & 24 & 24 & SDXL (24) & 0.18 & -- & -- & -- & 23 & 23 & 24 & 24 \\
SDXL & down to 512 px (53.6 dB) & 24 & 24 & 24 & SDXL (24) & 0.48 & -- & $-$0.01 & -- & 24 & 24 & 24 & 24 \\
DreamShaper-XL & down to 512 px (53.4 dB) & 24 & 24 & 24 & SDXL (24) & 0.45 & -- & $-$0.01 & -- & 24 & 23 & 24 & 24 \\
Anim-XL & down to 512 px (52.2 dB) & 0 & 0 & 0 & PG v2 (24) & 0.42 & -- & $-$0.01 & -- & 17 & 18 & 0 & 0 \\
PG v2 & down to 512 px (53.4 dB) & 24 & 24 & 24 & PG v2 (24) & 0.61 & -- & 0.01 & -- & 24 & 24 & 24 & 24 \\
\bottomrule
\end{tabular}
\end{table*}
 
This section gives the full setup and every condition of \cref{sec:exp_controls}.

\noindent\textbf{Setup.}
We take the Family-Tree images of the 13 public Family-Tree models and of FLUX.1-dev and SD3.5-large and build 83 conditions, 67 manipulations and 16 references with the model's own autoencoder or decoder.
(a) Finished images make a round trip through the autoencoder of another family (SD1.5, SD2.1, SDXL, SD3.5 or FLUX.1).
(b) The latents of a model are decoded by another decoder.
(c) SDXL, DreamShaper-XL, Animagine-XL-3.1 and Playground v2 are sampled at 512 pixels, and SDXL-Turbo at 1,024 pixels.
A further check resizes finished images before the standard resizing to 256 pixels.
Each condition has 24 disjoint sets of $k=16$ images from one caption split, and each set is attributed to the nearest base fingerprint built from the other caption half.
Before the run we fixed the rule that at least 20 of the 24 sets stay with their family in every tested condition. A condition is tested when the unmanipulated model passes the same rule.
The rule holds for 27 of 32 round trips, 6 of 10 decoder swaps and 1 of 4 resolution changes, and with SD1 and SD2 counted as one family for 31 of 32, 10 of 10 and 1 of 4 (\cref{tab:controls}).
The family signal therefore does not survive every manipulation, and the conditions where it fails show what it is made of.

\noindent\textbf{Decoder.}
SD1.5 latents decoded with the SD2.1 decoder keep 0 of 24 sets at SD1.5, and SD2.1 latents decoded with the SD1.5 decoder keep 1 of 24 at SD2.1 (\cref{tab:supp_controls_swap}).
Round trips of finished images agree: SD1.5 keeps 3 of 24 sets through the SD2.1 autoencoder, and SD2.1 keeps 1 of 24 through the SD1.5 autoencoder (\cref{tab:supp_controls_rt}).
Openjourney goes from 0 to 24 of 24 sets at SD1.5 with the SD1.5 decoder, and DreamShaper-8 from 0 to 11.
Through the SDXL and FLUX.1 autoencoders SD1.5 keeps 0 and 13 of 24 sets, and every set that leaves goes to SD2.1.

\noindent\textbf{Sampling resolution.}
Sampled at 512 pixels, SDXL, DreamShaper-XL and Playground v2 move by 6.84 to 8.39, an across-family distance, toward SD1.5 and SD2.1.
SDXL-Turbo sampled at 1,024 pixels moves by 6.83 and keeps all 24 sets at SDXL.
Resizing finished images before the standard resizing to 256 pixels keeps 24 of 24 sets in the 7 conditions of models that start in their family.

\begin{table}[t]
\centering
\caption{\textbf{Sampling resolution in the newer families.} Models sampled at 512 and 768 pixels on the Family-Tree captions with the seeds and settings of their 1,024-pixel images, 24 disjoint sets of $k=16$, CF features and the fingerprint distance of \cref{sec:supp_tree_metric}, one run. Number of the 24 sets attributed to the base of the model's own family. 17 bases: the 17 registered bases at their native resolution. 6 bases: the gallery of \cref{tab:controls}. Matched: the 17 bases plus every base sampled at 512 or 768 pixels, for models that are not a base themselves. Moved: distance from the fingerprint of the 1,024-pixel images.}
\label{tab:supp_resolution}
\footnotesize
\setlength{\tabcolsep}{3pt}
\begin{tabular}{@{}lccccc@{}}
\toprule
Model & Pixels & 17 bases & 6 bases & Matched & Moved \\
\midrule
FLUX.1-dev & 512 & 1 & 15 & -- & 4.43 \\
 & 768 & 18 & 24 & -- & 2.26 \\
\addlinespace[1pt]
FLUX.1-schnell & 512 & 0 & 2 & 24 & 5.33 \\
 & 768 & 13 & 24 & 24 & 2.35 \\
\addlinespace[1pt]
FLUX.1-lite-8B & 512 & 0 & 6 & 24 & 4.57 \\
 & 768 & 14 & 24 & 19 & 1.84 \\
\addlinespace[1pt]
SD3.5-large & 512 & 8 & 22 & -- & 4.47 \\
 & 768 & 24 & 24 & -- & 1.10 \\
\addlinespace[1pt]
SD3.5-large-TurboX & 512 & 13 & 23 & 24 & 5.70 \\
 & 768 & 24 & 24 & 24 & 2.06 \\
\addlinespace[1pt]
SD3.5-medium & 512 & 0 & 1 & 23 & 3.97 \\
 & 768 & 17 & 22 & 20 & 1.43 \\
\addlinespace[1pt]
Z-Image-Turbo & 512 & 18 & -- & -- & 4.59 \\
 & 768 & 24 & -- & -- & 1.64 \\
\addlinespace[1pt]
SANA 1.5 & 512 & 11 & -- & -- & 4.15 \\
 & 768 & 24 & -- & -- & 1.94 \\
\bottomrule
\end{tabular}
\end{table}
 
\noindent\textbf{Sampling resolution in the newer families.}
A second run samples FLUX.1-dev, FLUX.1-schnell, FLUX.1-lite-8B, SD3.5-large, SD3.5-large-TurboX, SD3.5-medium, Z-Image-Turbo and SANA 1.5 at 512 and at 768 pixels on 384 of the Family-Tree captions, with the seeds and settings of their 1,024-pixel images (\cref{tab:supp_resolution}).
Each set is attributed among the 17 registered bases of \cref{sec:supp_modern}.
Before the run we fixed the rule that at least 20 of the 24 sets of a model reach the base of its family at both resolutions, and the rule does not hold for either family.
At 512 pixels the six FLUX.1 and SD3.5 models keep 0 to 13 of 24 sets, and at 768 pixels SD3.5-large and SD3.5-large-TurboX keep all 24 and the other four 13 to 18.
The sets that leave go most often to CogView4, and the fingerprints move by 3.97 to 5.70 at 512 pixels and by 1.10 to 2.35 at 768 pixels.
With the 6-base gallery of \cref{tab:controls}, all six models keep at least 22 of 24 sets at 768 pixels.
We then add every base sampled at 512 or 768 pixels to the gallery, and a model's own images never build a fingerprint it is compared with.
FLUX.1-schnell, FLUX.1-lite-8B, SD3.5-large-TurboX and SD3.5-medium reach the base of their family in 95 of 96 sets at 512 pixels and in 87 of 96 at 768 pixels, almost always at the fingerprint of their own resolution.
The rule fixed for this gallery holds for 7 of the 9 tested conditions: FLUX.1-lite-8B at 768 pixels keeps 19 of 24 sets, and DreamShaper-XL at 512 pixels keeps none and goes to SD1.5 and SD2.1 with SDXL registered at 512 pixels.

\noindent\textbf{Family.}
SDXL, DreamShaper-XL and Playground v2 keep 24 of 24 sets through the SD1.5, SD2.1, SD3.5 and FLUX.1 autoencoders while their fingerprint moves by 0.43 to 2.98.
Decoded with TAESDXL, a decoder of another architecture, SDXL and Playground v2 keep all 24 sets while they move by 4.66 and 4.21, more than the mean distance within a family.
SD3.5-large keeps 22 of 24 sets through the SD1.5 autoencoder and 24 through the other three.
FLUX.1-dev keeps 19 of 24 through the SD1.5 autoencoder and 24 through the other three.
SD2.1 and SD2.1-768 keep 21 to 24 sets through the SDXL, SD3.5 and FLUX.1 autoencoders.
With CLIP ViT-L/14 features the families separate less (mean distance 2.59 within a family against 2.83 across).

\clearpage  %
\section{Discovery and enrollment of unseen generators}
\label{sec:supp_enroll}

A discovered generator can join the gallery as one more fingerprint, as in the discover-cluster-enroll loop of open-world GAN attribution~\cite{girish2021discovery}.
As in open-world GAN attribution~\cite{girish2021discovery} and generalized category discovery~\cite{vaze2022gcd}, sets flagged as unseen (\cref{sec:method_calibration}) can be clustered, and each cluster joins the gallery as a new fingerprint and tree branch.
The gallery grows by one fingerprint per cluster without refitting, as classes are added in class-incremental learning~\cite{xiang2022coarse,zhou2024continual}.
We test this loop on sets that LineagePrint flags as unseen.

\noindent\textbf{Setup.}
The gallery holds the 10 generators shared by the two datasets, with the subspace fit on clean COCO-dataset images.
The unseen sets come from 19 CF-Eval generators (all except GALIP and FLUX.1 schnell), or from AuraFlow and SD3.5 in the Flickr dataset.
Their CF-Eval images are treated as unseen, including FLUX.1 dev, whose COCO-dataset images are in the gallery.
LineagePrint (PLDA) flags sets of $k=8$ at $\alpha=0.05$.
For CF-Eval the threshold comes from anchor sets of the registered generators in the source dataset, and in the Flickr dataset from same-pipeline anchor sets.
The flagged sets are split at random into discovery sets (70\%) and later sets (30\%).
Ward clustering~\cite{ward1963hierarchical}, which merges the two clusters whose union least increases the within-cluster variance, groups the discovery sets.
The number of clusters, searched from 2 to 12, minimizes the squared distance of the later sets to their nearest cluster mean, so it uses no generator labels.
Each cluster is enrolled as a new gallery fingerprint, the mean of its discovery sets, and each later set is attributed to the nearest enrolled fingerprint.

\noindent\textbf{Metrics.}
Purity is the fraction of discovery sets that belong to the most common generator of their cluster.
The adjusted Rand index (ARI)~\cite{hubert1985comparing} measures the agreement between clusters and true generators (1 for a perfect match, 0 for a random assignment).
Later-set accuracy is the fraction of later sets attributed to a cluster whose most common discovery generator is their own generator.
All values are means over 3 splits.

\noindent\textbf{Results.}
On the 19 CF-Eval generators with CF features, the flagged sets cluster by generator with purity 0.77 and ARI 0.69.
The number of clusters reaches the top of the search range (12).
Later sets, which can share images with the discovery sets, reach the cluster of their own generator with accuracy 0.76.
With CLIP ViT-B/32 features the same loop gives purity 0.62, ARI 0.49 and later-set accuracy 0.62.
In the Flickr dataset, every cluster of flagged AuraFlow and SD3.5 sets holds a single generator (purity 1.00, CF).
Each of the two generators spans several clusters, so the ARI is 0.16, and every later set reaches a cluster of its own generator (accuracy 1.00).
Flagged sets therefore form clusters that the gallery can enroll as new fingerprints.

\section{Extended related work}
\label{sec:supp_related}

This section gives the related work of the main paper in full.

\noindent\textbf{AI-generated image detection and its generalization.}
Detectors train on images of one generative adversarial network (GAN)~\cite{wang2020cnn} or on frozen CLIP~\cite{radford2021clip} features~\cite{ojha2023universal}. Later detectors use spectral, up-sampling, autoencoder-reconstruction and semantic cues of diffusion images~\cite{corvi2023diffusion,corvi2023intriguing,bammey2024synthbuster,tan2024npr,ricker2024aeroblade,chen2024drct,yan2025aide}.
Detection generalizes better with more training generators~\cite{park2025community}. Detectors also learn format biases such as JPEG compression and image size~\cite{grommelt2024fakejpeg}, and aligned training images reduce these biases~\cite{rajan2025aligned,guillaro2025bfree}.
Detector thresholds shift on unseen generators and can be corrected on a small target set~\cite{yang2026calibrated}.
Proactive authentication instead checks a signal that real capture produces, such as the Moir\'e fringes of a grating filmed by a camera~\cite{qing2026moire}.
These detectors answer real versus fake, whereas we ask which generator or family made a set of outputs.

\noindent\textbf{Model attribution and fingerprints.}
Camera forensics averages sensor noise over many images into a device fingerprint~\cite{lukas2006digital,goljan2009large} and clusters images by device~\cite{marra2017blind}, as \citet{marra2019gans} do for GANs.
\citet{yu2019attributing} learn fingerprints that separate GANs differing only in their training seed, RepMix~\cite{bui2022repmix} makes GAN attribution robust to image transforms, and ManiFPT~\cite{song2024manifpt,song2025riemannian} defines fingerprints by the artifacts in a model's outputs. Other methods attribute single images to text-to-image generators~\cite{sha2023defake,xu2025origin}, to an architecture or family~\cite{asnani2023reverse,yang2022dnadet,guo2023hifinet,guarnera2024levelup}, or to one model by inversion~\cite{wang2023where,laszkiewicz2024single,wang2024latenttracer} or a few reference images~\cite{liu2024which}.
Open-set attribution rejects images of unseen generators~\cite{fang2023openset,abady2024siamese,yang2023pose,cioni2025clip,bongini2025wild}. \citet{girish2021discovery} and others~\cite{nguyen2025forensic,sun2023contrastive,zheng2026confidence,liu2025data} cluster the rejected images into new sources, as concurrent work does for faces~\cite{infantino2026facetrace}.
Incremental and few-shot methods add new generators~\cite{qin2026increfa,lee2025fewshot,sullivanpao2024lorax,wu2026fewshot,wang2026attribution}.
These methods score one test image at a time. Instead, we fingerprint the test set itself in a subspace that separates the registered generators and flag an unseen generator at a chosen false-alarm rate.

\noindent\textbf{Model lineage and provenance.}
From weights, \citet{horwitz2025unsupervised} recover the tree of fine-tunes, including a Stable Diffusion tree and the two parents of a merge, and propose an atlas of public models~\cite{horwitz2025atlas}.
Neural Lineage and Neural Phylogeny find the parent and the direction of a fine-tune~\cite{yu2024neural,yu2025phylogeny}, and \citet{shang2026attesting} attest lineage from parameter changes.
From outputs, DIF~\cite{sinitsa2024deep} extracts a fingerprint of one model from a few of its images and relates fine-tuned models to it. The closest work, \citet{ha2026identifying}, tests the output features of a prompted suspect model against a pool of base models.
\citet{lee2026csf} attribute fine-tunes with compositional prompts, TrainProVe~\cite{xie2025trainprove} tests training on a generator's outputs, and language-model provenance tests compare prompted outputs~\cite{nikolic2025provenance,yax2025phylolm}.
These methods need the weights, prompts sent to the suspect model, or a fingerprint extracted for each reference model.
We attribute community fine-tunes and descendants of the FLUX.1 and SD3.5 families from released outputs with one subspace, place weight merges in a family tree, and measure with decoder swaps what the family signal is made of.
A family and its models form a two-level label hierarchy. Hierarchical visual recognition is still hard for vision language models~\cite{tan2026llmbottleneck}, and hyperbolic embeddings capture hierarchical relations among images~\cite{li2025hypdae}.
Concurrent work recovers family structure from the features of an image attributor~\cite{livne2026scalable}.

\noindent\textbf{Out-of-distribution scores, set-level tests and conformal calibration.}
Out-of-distribution (OOD) scores include maximum softmax probability (MSP)~\cite{hendrycks2017msp,vaze2022openset}, logit energy~\cite{liu2020energy}, Mahalanobis distance~\cite{lee2018mahalanobis}, nearest-neighbor (kNN) distance~\cite{sun2022knn} and feature residuals~\cite{wang2022vim}, and open-set recognition rejects unknown classes~\cite{scheirer2013toward,bendale2016openmax}, also under a domain shift in universal domain adaptation~\cite{lai2024memspm}.
Set-level tests ask whether a sample differs from a reference, with maximum mean discrepancy (MMD)~\cite{gretton2012kernel,rabanser2019failing}, typicality~\cite{nalisnick2019typicality} or group scores~\cite{muandet2013oneclass}, and MMD on embeddings separates generative processes~\cite{mukherjee2026beyond}.
Conformal prediction turns a score into a threshold with a finite-sample guarantee~\cite{vovk2005conformal,lei2018conformal,angelopoulos2023conformal}, also under covariate shift~\cite{tibshirani2019conformal} and per class~\cite{ding2023classconditional}. Conformal outlier tests bound the false-alarm rate when calibration and test inliers are exchangeable~\cite{bates2023outliers}.
We obtain exchangeable calibration sets from registered generators that went through the same pipeline as the test sets.
 \fi
\end{document}